\documentclass[13pt]{article}
\usepackage{latexsym}
\usepackage{geometry}
\usepackage{graphicx}
\usepackage{amsmath, amssymb, amsthm}
\usepackage{booktabs}
\usepackage{algorithm}
\usepackage{algorithmic}

\usepackage{url}
\usepackage{natbib}
\usepackage{appendix}
\usepackage{amsthm}
\usepackage{amsmath}

\usepackage{amsfonts}
\usepackage{multirow}
\usepackage{multicol}

\usepackage{color}
\usepackage{algorithm}
\usepackage{algorithmic}

\usepackage{xcolor,colortbl}
\definecolor{LightCyan}{rgb}{0.88,1,1}

\usepackage{graphicx}
\usepackage{subfigure}
\usepackage{hyperref}
\usepackage{tcolorbox}

\newtheorem{theorem}{Theorem}
\newtheorem{lemma}{Lemma}

\newtheorem{assumption}{Assumption}
\newtheorem{remark}{Remark}

\begin{document}
\title{ Faster Adaptive Momentum-Based Federated Methods for Distributed Composition Optimization }
\author{Feihu Huang\thanks{ Feihu Huang is with College of Computer Science and Technology,
Nanjing University of Aeronautics and Astronautics, Nanjing, China;
and also with MIIT Key Laboratory of Pattern Analysis and Machine Intelligence, Nanjing, China.
E-mail: huangfeihu2018@gmail.com } }

\date{}
\maketitle

\begin{abstract}
Federated Learning is a popular distributed learning paradigm in machine learning.  
Meanwhile, composition optimization is an effective hierarchical learning model, which
 appears in many machine learning applications such as meta learning
and robust learning. More recently, although a few federated composition optimization algorithms have been proposed, they still suffer from high sample and communication complexities.
 In the paper, thus, we propose a class of faster federated compositional optimization algorithms (i.e., MFCGD and AdaMFCGD) to solve the nonconvex distributed composition problems, which builds on the momentum-based variance reduced and local-SGD techniques. In particular, our adaptive algorithm (i.e., AdaMFCGD) uses a unified adaptive matrix to flexibly incorporate various adaptive learning rates. Moreover, we provide a solid theoretical analysis for our algorithms under non-i.i.d. setting, and prove our algorithms obtain a lower sample and communication complexities simultaneously than the existing federated compositional algorithms. Specifically, our algorithms obtain lower sample complexity of $\tilde{O}(\epsilon^{-3})$
with lower communication complexity of $\tilde{O}(\epsilon^{-2})$ in finding an $\epsilon$-stationary solution.
We conduct the numerical experiments on robust federated learning and distributed meta learning tasks to demonstrate the efficiency of our algorithms.
\end{abstract}

\section{Introduction}
Composition optimization is an effective hierarchical model, which is widely used to many applications such as reinforcement learning~\citep{wang2017accelerating,huo2018accelerated},
meta learning~\citep{wang2021memory}, risk management~\citep{huo2018accelerated} and deep AUC maximization ~\citep{yuan2022compositional}.
In the paper, we study the following distributed composition optimization problem:
\begin{align} \label{eq:1}
 \min_{x \in \mathbb{R}^d} & \ F(x):=\frac{1}{M}\sum_{m=1}^M\mathbb{E}_{\xi^m \sim \mathcal{D}^m}\bigg[f^m\Big(\mathbb{E}_{\zeta^m \sim \mathcal{S}^m}\big[g^m(x;\zeta^m)\big];\xi^m\Big)\bigg],
\end{align}
where $y^m=g^m(x)=\mathbb{E}_{\zeta^m \sim \mathcal{S}^m}\big[g^m(x;\zeta^m)\big]$ and $f^m(y^m)=\mathbb{E}_{\xi^m \sim \mathcal{D}^m}\big[f^m(y^m;\xi^m)\big]$
for any $m\in[M]$ denote the inner
and outer objective functions respectively in
$m$-th client. Here $\xi^m$ and $\zeta^m$ for any $m\in[M]$ are independent random variables follow unknown distributions $\mathcal{D}^m$
and $\mathcal{S}^m$ respectively. For any $m,j\in [M]$ possibly $\mathcal{D}^m \neq \mathcal{D}^j$,
$\mathcal{S}^m \neq \mathcal{S}^j$  and $\mathcal{D}^m\neq \mathcal{S}^j$.
Applications of the problem \eqref{eq:1} involves many machine learning problems with a compositional structure, which include model-agnostic meta learning \citep{tutunov2020compositional,chen2020solving,wang2021memory},
reinforcement learning~\citep{wang2017accelerating,huo2018accelerated} and sparse additive models \citep{wang2017stochastic}.
In the following, we give two specific applications
that can be formulated as the distributed composition optimization problem \eqref{eq:1}.

\begin{table*}
  \centering
  \caption{ \textbf{Sample} and \textbf{Communication} complexities comparison of the representative \textbf{federated compositional optimization} algorithms in finding
  an $\epsilon$-stationary solution of the distributed composition optimization problem \eqref{eq:1}, i.e., $\mathbb{E}\|\nabla F(x)\|\leq \epsilon$
  or its equivalent variants. \textbf{ALR} denotes adaptive learning rate.  }
  \label{tab:1}
   \resizebox{\textwidth}{!}{
\begin{tabular}{c|c|c|c|c}
  \hline
   \textbf{Algorithm} & \textbf{Reference} & \textbf{Sample Complexity} & \textbf{Communication Complexity}& \textbf{ALR}   \\ \hline
  ComFedL  & \cite{huang2021compositional}  & $O(\epsilon^{-8})$ & $O(\epsilon^{-4})$ &\\  \hline
  LocalMOML  & \cite{wang2021memory} & $O(\epsilon^{-5})$ & $O(\epsilon^{-3})$ &  \\  \hline
  FEDNEST  & \cite{tarzanagh2022fednest} & $\tilde{O}(\epsilon^{-4})$ & $\tilde{O}(\epsilon^{-4})$ &  \\  \hline
  Local-SCGDM  & \cite{gao2022convergence} & $O(\epsilon^{-4})$ & $O(\epsilon^{-3})$ &  \\  \hline
  MFCGD  & Ours  & {\color{red}{$\tilde{O}(\epsilon^{-3})$}} & {\color{red}{$\tilde{O}(\epsilon^{-2})$}} &  \\  \hline
  AdaMFCGD  & Ours & {\color{red}{$\tilde{O}(\epsilon^{-3})$}} & {\color{red}{$\tilde{O}(\epsilon^{-2})$}} & $\checkmark$ \\  \hline
\end{tabular}
 }
\end{table*}

\textbf{1). Task-Distributed Meta Learning.}
Meta Learning is to learn some properties in the optimal model to improve model performances
with more experiences, i.e., learning to learn \citep{andrychowicz2016learning}.
Model-Agnostic Meta Learning (MAML) \citep{finn2017model} is a class of popular meta learning methods,
which is to find a common initialization that can adapt to a desired model for a set of new tasks after taking several gradient descent steps.
In the paper, we consider a class of task-distributed MAMLs, where a set of tasks $\{\mathcal{T}_m\}_{m=1}^M$ are drawn from a certain task distribution
and each task is assigned in each client.
Specifically, we solve the following task-distributed MAML problem:
\begin{align} \label{eq:2}
 \min_{x\in \mathbb{R}^d} \frac{1}{M}\sum_{m=1}^M f^m\big(x-\eta\nabla f^m(x)\big),
\end{align}
where $f^m(x)=\mathbb{E}_{\xi^m\sim \mathcal{D}^m}[f(x;\xi^m)]$, and random variable $\xi^m$ follows the unknown distribution $\mathcal{D}^m$, and $\eta >0$ is a learning rate.
Let $f^m(y^m)=f^m(g^m(x))$ and $y^m=g^m(x)=x-\eta\nabla f^m(x)$, the above problem (\ref{eq:2}) is a special case of the above composition problem \ref{eq:1}.

\textbf{2). Distributionally Robust Federated Learning.}
Federated learning (FL)~\citep{mcmahan2017communication,kairouz2019advances} is a distributed and privacy preserving machine learning method to learn a global model
collaboratively from decentralized data distributed over a network of devices.
To tackle the data heterogeneity from different devices,
some robust FL algorithms \citep{mohri2019agnostic,reisizadeh2020robust,deng2020distributionally} have been studied.
In the paper, as in \citep{huang2021compositional}, we consider solving the following distributed composition problem
to reach distributionally robust FL, defined as
\begin{align} \label{eq:3}
 \min_{x \in \mathbb{R}^d} \frac{1}{M}\sum_{m=1}^M f\Big(\mathbb{E}\big[g^m(x;\xi^m)\big]\Big),
\end{align}
where $g^m(x) = \mathbb{E}\big[g^m(x;\xi^m)\big]$ denotes the loss function in the $m$-th client,
and $f(\cdot)$ is a monotonically increasing function.
Clearly, the problem \eqref{eq:3} is a special case of the above problem \eqref{eq:1}.

Although recently many compositional gradient algorithms have been proposed to solve the composition problems, few distributed algorithms focus on
 solving the distributed composition optimization problems. More recently, \cite{huang2021compositional,wang2021memory,gao2022convergence,tarzanagh2022fednest} proposed some federated compositional gradient algorithms for the distributed stochastic composition problems.
However, few adaptive algorithm focuses on
the composition optimization problems under the distributed setting.
Meanwhile, these existing federated composition optimization
methods suffer from large sample complexity and communication complexity (Please see Table \ref{tab:1}).
Then there exists a natural question:
\begin{center}
\begin{tcolorbox}
\textbf{ Could we develop faster and adaptive federated learning methods to solve the distributed composition optimization problem \eqref{eq:1} ? }
\end{tcolorbox}
\end{center}

In the paper, we provide an affirmative answer to the above question and propose
a class of faster momentum-based federated compositional gradient descent algorithms (i.e., MFCGD and AdaMFCGD)
 to solve the problem \eqref{eq:1},
which builds on the local Stochastic Gradient Descent (SGD) and momentum-based variance reduced techniques
to obtain a lower sample and communication complexities simultaneously.
Our main contributions are as follows:
\begin{itemize}
\item[(1)] We propose a class of faster momentum-based federated compositional gradient descent algorithms (i.e., MFCGD and AdaMFCGD) to solve the nonconvex distributed composition problems, which builds on the momentum-based variance reduced and local-SGD techniques. In particular, our adaptive algorithm (i.e., AdaMFCGD) uses a unified adaptive matrix to flexibly incorporate various adaptive learning rates.
\item[(2)] We provide a solid convergence analysis framework for our algorithms under non-i.i.d. setting, and prove that our algorithms obtain simultaneously lower sample complexity of $\tilde{O}(\epsilon^{-3})$
and lower communication complexity of $\tilde{O}(\epsilon^{-2})$ than the existing federated composition methods for finding an $\epsilon$-stationary solution (Please see Table \ref{tab:1}).
\item[(3)] Experimental results demonstrate efficiency of our algorithms on the robust federated learning and  task-distributed meta learning.
\end{itemize}

\section{Related Works}
In this section, we overview some representative composition optimization, federated optimization and
adaptive optimization methods, respectively.

\subsection{ Composition Optimization }
Composition optimization has been widely applied to
many applications such as reinforcement learning \citep{wang2017accelerating},
model-agnostic meta Learning \citep{tutunov2020compositional} and risk management \citep{huo2018accelerated}.
Recently, many compositional gradient-based methods have recently been proposed
to solve these composition optimization problems.
For example, stochastic compositional gradient methods \citep{wang2017stochastic,wang2017accelerating,ghadimi2020single} have been proposed
to solve these problems.
Subsequently, some variance-reduced compositional algorithms \citep{huo2018accelerated,lin2018improved,zhang2019multi}
have been proposed for composition optimization.
\cite{tutunov2020compositional,chen2020solving} presented a class of momentum-based compositional
gradient methods for stochastic composition optimization. More recently, \cite{jiang2022optimal} proposed a class of efficient momentum-based variance reduced methods for non-convex
stochastic composition optimization. \cite{huang2022riemannian} studied the stochastic composition optimization on Riemannian manifolds.

For the distributed setting, \cite{huang2021compositional} firstly studied federated learning algorithm for the general distributed composition optimization. Meanwhile, \cite{wang2021memory} studied personalized federated learning algorithm based on the composition optimization. Subsequently, \cite{gao2022convergence,tarzanagh2022fednest} proposed some accelerated federated learning algorithms for the distributed composition optimization.

\subsection{ Federated Optimization }
Federated Learning (FL) is a popular distributed machine learning framework for collaboratively training the global model without sharing the local data, and is widely used in many applications such as healthcare informatics \citep{xu2021federated} and automatic diagnosis of COVID-19 \citep{yang2021flop}. \cite{mcmahan2017communication} first studied FL and
proposed the FedAvg algorithm for FL based on local-SGD algorithms~\citep{stich2019local},
where each client conducts multiple steps of SGD with its local data
and then sends the learned model to the server for averaging.
Subsequently, \citep{li2019convergence,karimireddy2019error,deng2021local} have studied the convergence properties of the local-SGD and FedAvg algorithms or their variations.
To accelerate the vanilla local-SGD and FedAvg algorithms, various accelerated FL algorithms~\citep{yuan2020federated,karimireddy2020scaffold,khanduri2021stem,chen2020fedcluster} have been developed and studied.
For example, \cite{karimireddy2020scaffold} proposed a stochastic controlled
averaging algorithm for FL by adopting the variance-reduced technique of SARAH~\citep{nguyen2017sarah}/SPIDER~\citep{fang2018spider}. Subsequently, \cite{khanduri2021stem} proposed a  faster federated algorithm based on momentum-based variance reduced technique of STORM \citep{cutkosky2019momentum} and ProxHSGD~\citep{tran2022hybrid}, which obtains lower sample and communication complexities simultaneously.

To solve the data heterogeneity in FL, \cite{mohri2019agnostic,deng2020distributionally}
proposed some effective robust FL algorithms by learning the
worst-case loss based on the minimax optimization problems. To further incorporate personalization in FL,
some personalized federated learning models \citep{fallah2020personalized,deng2020adaptive,li2021ditto} have been developed and studied. For example, \citep{li2021ditto} proposed an effective and efficient personalized FL
algorithm (i.e., Ditto) by learning a regularized local model for each client.

\subsection{ Adaptive Optimization Methods }
Adaptive optimization methods~\citep{duchi2011adaptive,kingma2014adam}
are a class of efficient optimization methods due to using adaptive learning rates in machine learning, and they have been widely studied in machine learning community.
For example, AdaGrad~\citep{duchi2011adaptive} is the first adaptive gradient method.
Adam~\citep{kingma2014adam} is a popular variation of AdaGrad algorithm based on the momentum technique, which is the default optimization algorithm
 for training large-scale machine learning models.
Meanwhile, some variants of Adam algorithm \citep{reddi2019convergence,chen2019convergence}
have been proposed to obtain a convergence guarantee under the nonconvex setting.
To further improve the performance of Adam algorithm, recently some new its variants such as AdamW \citep{loshchilov2018decoupled} have been developed.
More recently, some accelerated adaptive gradient methods \citep{cutkosky2019momentum,huang2021super} have been proposed based on the momentum-based variance reduced techniques.
In parallel, some adaptive gradient methods \citep{reddi2020adaptive,chen2020toward} are proposed for distributed optimization. For example, \cite{reddi2020adaptive} proposed a class of adaptive federated algorithms for FL
by using adaptive learning rates at the server side.

\section{Preliminaries}

\subsection{Notations}
Let $[M]$ denote the set $\{1,2,\cdots,M\}$.
$\|\cdot\|$ denotes the $\ell_2$ norm for vectors and Frobenius norm for matrices.
$\langle x,y\rangle$ denotes the inner product of two vectors $x$ and $y$. For vectors $x$ and $y$, $x^r \ (r>0)$ denotes the element-wise
power operation, $x/y$ denotes the element-wise division and $\max(x,y)$ denotes the element-wise maximum. $I_{d}$ denotes a $d$-dimensional identity matrix. $A\succ 0$ denotes that $A$ is a positive definite matrix.
$a_t=O(b_t)$ denotes that $a_t \leq c b_t$ for some constant $c>0$. The notation $\tilde{O}(\cdot)$ hides logarithmic terms. $\Pi_{C}\big[x\big] = \arg\min_{||w||\leq C} ||x-w||^2$ denote a projection onto the
ball with radius $C>0$.

\subsection{Classic Federated Learning}
The classic Federated Learning (FL) solves the following distributed optimization problem:
\begin{align}
 \min_{x \in \mathbb{R}^d} \frac{1}{M}\sum_{m=1}^M \mathbb{E}_{\xi^m \sim \mathcal{D}^m}[\ell_m(x;\xi^m)],
\end{align}
where $\ell_m(x;\xi^m)$ is the loss function on $m$-th device,
and $\mathcal{D}^m$ denotes the data distribution on $m$-th device.
In FL, the data distributions $\{\mathcal{D}^m\}_{m=1}^M$ generally are different, i.e., for any $m,j\in [M]$ possibly $\mathcal{D}^m \neq \mathcal{D}^j$. The goal of FL is to
learn a global variable $x$ based on these heterogeneous data from different data distributions.

\subsection{Federated Composition Optimization}
In the paper, we studied Federated Composition Optimization (FCO) defined as:
\begin{align}
 \min_{x \in \mathbb{R}^d} & \ F(x):=\frac{1}{M}\sum_{m=1}^M\mathbb{E}_{\xi^m \sim \mathcal{D}^m}\bigg[f^m\Big(\mathbb{E}_{\zeta^m \sim \mathcal{S}^m}\big[g^m(x;\zeta^m)\big];\xi^m\Big)\bigg],
\end{align}
where $y^m=g^m(x)=\mathbb{E}_{\zeta^m \sim \mathcal{S}^m}\big[g^m(x;\zeta^m)\big]$ and $f^m(y^m)=\mathbb{E}_{\xi^m \sim \mathcal{D}^m}\big[f^m(y^m;\xi^m)\big]$
for any $m\in[M]$ denote the inner
and outer objective functions respectively in
$m$-th client. Here $\xi^m$ and $\zeta^m$ for any $m\in[M]$ are independent random variables follow unknown distributions $\mathcal{D}^m$
and $\mathcal{S}^m$ respectively. For any $m,j\in [M]$ possibly $\mathcal{D}^m \neq \mathcal{D}^j$,
$\mathcal{S}^m \neq \mathcal{S}^j$  and $\mathcal{D}^m\neq \mathcal{S}^j$. Due to the existence of composition objective function and double unknown distributions, FCO has more challenges than classic FL.

\section{ Federated Compositional Gradient Descent Algorithms }
In this section, we propose a class of faster momentum-based federated compositional gradient descent algorithms (i.e., MFCGD and AdaMFCGD) to
solve the problem \eqref{eq:1}, which builds on the local-SGD and
momentum-based variance reduced techniques. Specifically, the local-SGD technique reduce the communication complexity and
the momentum-based variance reduced technique reduce the sample complexity without relying on large batches.
Meanwhile, our AdaMFCGD algorithm uses the unified adaptive matrix to flexibly
incorporate various adaptive learning rates in updating variables.
Specifically, Algorithm \ref{alg:1} provides a procedure framework of our MFCGD and AdaMFCGD algorithms.

\begin{algorithm}[t]
\caption{ \textbf{MFCGD} and \textbf{AdaMFCGD} Algorithms }
\label{alg:1}
\begin{algorithmic}[1]
\STATE {\bfseries Input:} $T, q$, tuning parameters $\{\gamma, \eta_t, \alpha_t, \beta_t, \varrho_t\}$ and initial input $x_1\in \mathbb{R}^d$; \\
\STATE {\bfseries initialize:} Set $x^m_1=x_1$ for $m \in [M]$, and draw $2q$ independent samples $\{\xi^m_{1,j}\}_{j=1}^q$ and $\{\zeta^m_{1,j}\}_{j=1}^q$,
and then compute $h^m_1 = \frac{1}{q}\sum_{j=1}^q g^m(x^m_1;\zeta^m_{1,j})$, $u^m_1 = \frac{1}{q}\sum_{j=1}^q \nabla g^m(x^m_1;\zeta^m_{1,j})$ and $v^m_1 = \frac{1}{q}\sum_{j=1}^q \nabla f(h^m_1;\xi^m_{1,j})$ for all $m \in [M]$;
Generate adaptive matrix $A_1 \in \mathbb{R}^{d \times d}$. \\
\FOR{$t=1$ \textbf{to} $T$}
\IF {$\mod(t,q)=0$}
\STATE $\bar{w}_t = \frac{1}{M} \sum_{m=1}^{M} w^m_t$ and $\bar{x}_t = \frac{1}{M} \sum_{m=1}^{M} x^m_t$; \\
\STATE Generate the adaptive matrix $A_t \in \mathbb{R}^{d \times d}$;\\
\textcolor{blue}{One example of $A_t$ by using update rule ($a_0 = 0$, $ 0 < \vartheta_t < 1$, $\rho>0$.) } \\
\textcolor{blue}{ Compute $a_t = \vartheta_t a_{t-1} + (1 - \vartheta_{t})\bar{w}_t^2$, $A_t = \mbox{diag}(\sqrt{a_t} + \rho)$}; \\
\STATE $x^m_{t+1}=\bar{x}_{t+1} = \arg\min_{x\in \mathbb{R}^d }\Big\{\langle x,\bar{w}_t\rangle + \frac{1}{2\eta_t\gamma}\big(x-\bar{x}_t\big)^TA_{t}\big(x-\bar{x}_t\big)\Big\}$; (Sent them to Clients) \\
\ELSE
\FOR{each client $m\in [M]$ \ (\textbf{in parallel})}
\STATE $w^m_t = (u^m_t)^Tv^m_t$; \\
\STATE $x^m_{t+1} = \arg\min_{x\in \mathbb{R}^d }\Big\{\langle x,w^m_t\rangle + \frac{1}{2\eta_t\gamma}\big(x-x^m_t\big)^TA_{t}\big(x-x^m_t\big)\Big\}$; \\
\STATE $A_{t+1} = A_t$; \\
\ENDFOR
\ENDIF
\FOR{each client $m\in [M]$ \ (\textbf{in parallel})}
\STATE Draw two independent samples $\xi^m_{t+1}$ and $\zeta^m_{t+1}$;
\STATE{} $h^m_{t+1} = g^m(x^m_{t+1};\zeta^m_{t+1}) + (1-\alpha_{t+1})\big(h^m_t - g^m(x^m_t;\zeta^m_{t+1})\big)$;
\STATE{} $u^m_{t+1} = \Pi_{C_g}\Big[ \nabla g^m(x^m_{t+1};\zeta^m_{t+1}) + (1-\beta_{t+1})\big(u^m_t - \nabla g^m(x^m_t;\zeta^m_{t+1})\big)\Big]$;
\STATE{} $v^m_{t+1} = \Pi_{C_f}\Big[ \nabla f^m(h^m_{t+1};\xi^m_{t+1}) + (1-\varrho_{t+1})\big(v^m_t - \nabla f(h^m_t;\xi^m_{t+1})\big) \Big]$;
\STATE{} $w^m_{t+1} = (u^m_{t+1})^Tv^m_{t+1}$;
\ENDFOR
\ENDFOR
\STATE {\bfseries Output:} Chosen uniformly random from $\{\bar{x}_t\}_{t=1}^{T}$.
\end{algorithmic}
\end{algorithm}

In Algorithm \ref{alg:1}, when $\mbox{mod}(t,q)= 0$ (i.e., \textbf{synchronization} step), the server receives the local variables $\{x^m_t\}_{m=1}^M$ and
local gradients $\{w_t^m\}_{m=1}^M$ from the clients, and then averages them to obtain
the averaged variables $\{\bar{x}_t\}$ and averaged gradients $\{\bar{w}_t\}$.
Based on
these averaged gradients $\{\bar{w}_t\}$, we can generate some adaptive matrices $\{A_t\}_{t\geq 1}$ (i.e., adaptive learning rates).
\textbf{Note that} for our non-adaptive MFCGD algorithm, we only set $A_t=I_d$ for all $t\geq 1$ in Algorithm \ref{alg:1}.
Besides one example given
at the line 6 of Algorithm \ref{alg:1}, we can also generate many other adaptive matrices. For example, we can generate
adaptive matrix $A_t$ as the norm-type of Adam, defined as
\begin{align}
 & a_t = \vartheta_t a_{t-1} + (1 - \vartheta_{t})\|\bar{w}_t\|, \quad A_t = \mbox{diag}(a_t + \rho),
\end{align}
where $0<\vartheta_{t}\leq 1$. Note that we can directly choose $\alpha_t$, $\beta_t$ or $\varrho_t$ instead of $\vartheta_t$ to reduce the number of tuning parameters in
our algorithm. Next, based on these adaptive matrices, we can update the variable $x$ in the server,
then sent it to each client.

When $\mbox{mod}(t,q) \neq 0$ (i.e., \textbf{asynchronization} step), the clients receive the updated variables $\{\bar{x}_{t+1}\}$
and the generated adaptive matrices $\{A_t\}$ from the server. Then the clients use the momentum-based variance reduced technique of STORM \citep{cutkosky2019momentum} and ProxHSGD~\citep{tran2022hybrid} 
to update the stochastic gradients based on local data: for $m\in [M]$
\begin{align}
& h^m_{t+1} = g^m(x^m_{t+1};\zeta^m_{t+1}) + (1-\alpha_{t+1})\big(h^m_t - g^m(x^m_t;\zeta^m_{t+1})\big)  \\
& u^m_{t+1} = \Pi_{C_g}\Big[ \nabla g^m(x^m_{t+1};\zeta^m_{t+1}) + (1-\beta_{t+1})\big(u^m_t - \nabla g^m(x^m_t;\zeta^m_{t+1})\big)\Big] \\
& v^m_{t+1} = \Pi_{C_f}\Big[ \nabla f^m(h^m_{t+1};\xi^m_{t+1}) + (1-\varrho_{t+1})\big(v^m_t - \nabla f(h^m_t;\xi^m_{t+1})\big) \Big],
\end{align}
where $\alpha_{t+1}\in (0,1)$, $\beta_{t+1}\in (0,1)$ and $\varrho_{t+1} \in (0,1)$. Here the projection functions $\Pi_{C_g}\big[\cdot\big]$ and
$\Pi_{C_f}\big[\cdot\big]$ ensure that the estimated stochastic gradients $u^m_{t+1}$ and $v^m_{t+1}$ are bounded, i.e., $\|u^m_{t+1}\|\leq C_g $
and $\|v^m_{t+1}\|\leq C_f$ for any $t\geq 1$. Based on the estimated stochastic gradients and adaptive matrices, the clients
update the variables $\{x^m_t\}_{m=1}^M$, defined as
\begin{align}
 x^m_{t+1} = x^m_t - \gamma \eta_t A_t^{-1} w^m_t = \arg\min_{x\in \mathbb{R}^d }\Big\{\langle x,w^m_t\rangle + \frac{1}{2\eta_t\gamma}\big(x-x^m_t\big)^TA_{t}\big(x-x^m_t\big)\Big\},
\end{align}
where $\gamma>0$ and $\eta_t>0$.
In our algorithms, all clients use the same adaptive matrix generated from the server as in~\citep{chen2020toward}. \textbf{Note that} the existing adaptive FL algorithms such as
local-AMSGrad~\citep{chen2020toward} only builds on some specific adaptive learning rates such as AMSGrad~\citep{reddi2019convergence}. However, our algorithms can use the unified adaptive matrix to flexibly incorporate various adaptive learning rates.

\section{Convergence Analysis}
In this section, we study the convergence properties of our MFCGD and AdaMFCGD algorithms under some mild assumptions. All related proofs are provided in the Appendix.
 We first review some useful lemmas and assumptions.

\begin{assumption} \label{ass:1}
\textbf{(Lipschitz Gradients)}
For any $m\in [M]$, there exist constants $L_f$ and $L_g$ for $\nabla f^m(y;\xi^m)$, $\nabla g^m(x;\zeta^m)$ respectively satisfying
\begin{align}
& \|\nabla g^m(x_1,\zeta^m) - \nabla g^m(x_2,\zeta^m)\| \leq L_g \|x_1-x_2\|, \ \forall x_1,x_2 \in \mathbb{R}^d, \nonumber\\
& \|\nabla f^m(y_1;\xi^m) - \nabla f^m(y_2;\xi^m)\| \leq L_f \|y_1-y_2\|, \ \forall y_1,y_2 \in \mathbb{R}^p.  \nonumber
\end{align}
\end{assumption}

\begin{assumption} \label{ass:2}
\textbf{(Bounded Gradients)}
For any $m\in [M]$, gradient $\nabla g^m(x;\zeta^m)$ and Jacobian matrix $\nabla f^m(y;\xi^m)$ have the upper bounds $C_g$ and $C_f$ respectively, i.e.,
\begin{align}
 \|\nabla g^m(x;\zeta^m)\| \leq C_g, \ \|\nabla f^m(y;\xi^m)\| \leq C_f, \ \forall x \in \mathbb{R}^d,\ y \in \mathbb{R}^p. \nonumber
\end{align}
\end{assumption}

\begin{assumption}
\textbf{(Bounded Variances)} \label{ass:3}
For any $m\in [M]$, functions $f^m(y;\xi^m)$ and
$g^m(x;\zeta^m)$ and its gradients are unbiased and the bounded variances, i.e., we have
$\mathbb{E}[g^m(x;\zeta^m)] = g^m(x)$, $\mathbb{E}[\nabla g^m(x;\zeta^m)] = \nabla g^m(x)$,
$\mathbb{E}[\nabla f^m(y;\xi^m)] = \nabla f^m(y)$ and
\begin{align}
  & \mathbb{E}\|g^m(x;\zeta^m) - g^m(x)\|^2 \leq \sigma^2, \quad
    \mathbb{E}\|\nabla g^m(x;\zeta^m)- \nabla g^m(x)\|^2 \leq \sigma^2, \nonumber \\
  & \mathbb{E}\|\nabla f^m(y;\xi^m) - \nabla f^m(y)\|^2 \leq \sigma^2, \quad  \forall x \in \mathbb{R}^d, \ y \in \mathbb{R}^p \nonumber
\end{align}
where $\sigma>0$.
\end{assumption}

\begin{assumption} \label{ass:4}
$F(x)$ has a lower bound, i.e.,  $F^* = \inf_{x\in \mathbb{R}^d} F(x)$.
\end{assumption}

\begin{assumption} \label{ass:5}
In our algorithms, the adaptive matrix $A_t$ for all $t\geq 1$ satisfies
 $A_t \succeq \rho I_d $, where $\rho> 0$ is an appropriate positive number.
\end{assumption}

\begin{assumption} \label{ass:6}
 For any $m, j \in [M]$, $x\in \mathbb{R}^d$ and $y\in \mathbb{R}^p$, we have $ \|\nabla f^{m}(y) - \nabla f^{j}(y)\| \leq \delta_f$,
 $ \| \nabla g^{m}(x) - \nabla g^{j}(x)\| \leq \delta_{g}$ and $ \|g^{m}(x) - g^{j}(x)\| \leq \delta_{g}$, where $\delta_f>0$ and $\delta_{g}>0$ are constants.
\end{assumption}

Assumptions \ref{ass:1} ensures the smoothness of functions $f^m(y;\xi^m)$, $g^m(x;\zeta^m)$ for any $m\in [M]$,
Assumption \ref{ass:2} ensures the bounded gradients (or Jacobian matrix) of functions $f^m(y;\xi^m)$ and $g^m(x;\zeta^m)$ for any $m\in [M]$.
Assumption \ref{ass:3} ensures the bounded variances of stochastic gradient or value of functions $f^m(y;\xi^m)$ and
$g^m(x;\zeta^m)$ for any $m\in [M]$.
Assumption~\ref{ass:4} guarantees the feasibility of the problem~\eqref{eq:1}.
Assumptions~\ref{ass:1}-\ref{ass:4} have been commonly used in the convergence
analysis of the stochastic composition algorithms~\citep{wang2017stochastic,wang2017accelerating}.
Assumption \ref{ass:5} has been commonly used in the existing adaptive methods~\citep{huang2021super}.
Assumption~\ref{ass:6}  is the standard condition constrained the data heterogeneity
in non-i.i.d FL setting \citep{li2019convergence}. In fact, we can obtain the part results of Assumption~\ref{ass:6} based on
Assumptions~\ref{ass:1}-\ref{ass:2}. For example, we have
\begin{align}
 & \|\nabla f^{m}(y) - \nabla f^{j}(y)\| =  \|\nabla f^{m}(y) - \nabla f^{m}(y;\xi^m) + \nabla f^{m}(y;\xi^m) -\nabla f^{j}(y;\xi^j) + \nabla f^{j}(y;\xi^j) - \nabla f^{j}(y)\| \nonumber \\
 & \leq \|\nabla f^{m}(y) - \nabla f^{m}(y;\xi^m)\| + \|\nabla f^{m}(y;\xi^m)\| + \|\nabla f^{j}(y;\xi^j)\| + \|\nabla f^{j}(y;\xi^j) - \nabla f^{j}(y)\| \nonumber \\
 & \leq 2\sigma + 2C_{f},
\end{align}
where the last inequality holds by Assumptions~\ref{ass:1}-\ref{ass:2}. Similarly, we have $\|\nabla g^{m}(y) - \nabla g^{j}(y)\| \leq 2\sigma + 2C_{g}$ based on Assumptions~\ref{ass:1}-\ref{ass:2}.

\begin{lemma}
Given the above Assumptions \ref{ass:1}-\ref{ass:2},
the function $F(x)=\frac{1}{M}\sum_{m=1}^Mf^m(g^m(x))$ is $L$-smooth, i.e., for any $x_1,x_2\in \mathbb{R}^d$, we have
\begin{align}
 \|\nabla F(x_1) - \nabla F(x_2)\|^2 \leq L^2\|x_1-x_2\|^2,
\end{align}
where $L = \sqrt{2C^2_fL^2_g + 2C^4_gL^2_f}$.
\end{lemma}

\begin{lemma}
Assume the gradient estimator $\{\bar{w}_t\}_{t=1}^T$ generated from Algorithm \ref{alg:1}, where $w_t = \frac{1}{M}\sum_{m=1}^Mw^m_t$,
we have
\begin{align}
 \|\bar{w}_t - \nabla F(\bar{x}_t)\|^2
 & \leq \frac{1}{M}\sum_{m=1}^{M}\Big( 2C_f^2\|u^m_t - \nabla g^m(\bar{x}_t)\|^2 + 4C_g^2\|v^m_t - \nabla f^m(h^m_t)\|^2 \nonumber \\
 & \quad + 4C_g^2L^2_f\|h^m_t - g^m(\bar{x}_t)\|^2 \Big).
\end{align}
\end{lemma}

\subsection{ Convergence Properties of AdaMFCGD Algorithm }
In this subsection, we provide the convergence properties of our \textbf{AdaMFCGD} algorithm.

\begin{theorem} \label{th:1}
Assume the sequence $\{\bar{x}_t\}_{t=1}^T$ be generated from \textbf{AdaMFCGD} algorithm.
 Under the above Assumptions, and let $\eta_t=\frac{k}{(n+t)^{1/3}}$ for all $t\geq 0$, $\alpha_{t+1}=c_1\eta_t^2$, $\beta_{t+1}=c_2\eta_t^2$, $\varrho_{t+1}=c_3\eta_t^2$, $n \geq \max\big(2, k^3, (c_1k)^3, (c_2k)^3, (c_3k)^3, \frac{(24k\gamma qL_{fg}C_{fg})^3}{\rho^3}\big)$, $k>0$,
 $c_1 \geq \frac{2}{3k^3} + B$, $c_2 \geq \frac{2}{3k^3} + 5C^2_f$, $c^2_1 + c^2_2 \leq \frac{(24)^4q^2\gamma^4L^4_{fg}C^4_{fg}}{9\rho^4}$, $c_3 \geq \frac{2}{3k^3} + 5C^2_g$, $ \frac{\rho(c^2_1+c^2_3)^{1/4}}{12\sqrt{5q} L_{fg}C_{fg}} \leq \gamma \leq \min\Big(\frac{3\rho qL_{fg}C_{fg}}{4(C^2_g+L^2_g+ 2L^2_fC^2_g)}, \frac{n^{1/3}\rho}{2Lk} \Big)$, $B\geq 20C_g^2L^2_f + \frac{c^2_2C^2_gL^2_f}{216q^3\gamma^3L^3_{fg}C^3_{fg}} + \frac{\Theta\rho^2(c_1^2+c_3^2)}{30q^2\gamma^4C^2_{fg}L^2_{fg}C^2_g}$, $\Theta = \Big(5C^2_fL^2_g + \frac{c^2_2C^2_gL^2_f}{864q^3\gamma^3L^3_{fg}C^3_{fg}} \Big)\frac{\rho^2}{(24)^2L^2_{fg}C^2_{fg}}+\frac{\gamma\rho}{6qL_{fg}C_{fg}}\Big(C_g^2 + L_g^2 + 2L^2_fC^2_g\Big)$ and $\Theta+\frac{BC^2_g\rho^2}{(24)^2L^2_{fg}C^2_{fg}}\leq \frac{5\rho^2}{48}$, we have
\begin{align}
\frac{1}{T}\sum_{t=1}^T\mathbb{E}\|\nabla F(\bar{x}_t)\|  \leq \Big( \frac{\sqrt{2G}n^{1/6}}{T^{1/2}} + \frac{\sqrt{2G}}{T^{1/3}}\Big)\sqrt{\frac{1}{T}\sum_{t=1}^T\mathbb{E}\|A_t\|^2},
\end{align}
where $C^2_{fg}=\max(C^2_f,C^2_g)$, $L^2_{fg}=L^2_fC^2_g+L^2_g$, $G = \frac{4(F(\bar{x}_1) - F^*)}{k\rho\gamma} + \frac{12n^{1/3}\sigma^2}{qk^2\rho^2} + 4k^2\Big(\frac{\hat{\delta}^2}{4\gamma^2 L^2_{fg}} + \frac{\big( c^2_1 + c^2_2 + c^2_3 \big)\sigma^2}{3\rho\gamma qL_{fg}C_{fg}}\Big)\ln(n+T)$  and $\hat{\delta}^2 =  2c_1^2 L^2_f\sigma^2 + c_3^2\sigma^2 + 4c_3^2\delta^2_f + 4c_3^2 L^2_f\delta^2_g + c^2_2\sigma^2 + 3c^2_2\delta_{g}^2$.
\end{theorem}

\begin{remark}
Under the above Assumption \ref{ass:2}, we have $\big\|\frac{1}{M}\sum_{m=1}^M\big(\nabla g^m(\bar{x}_t)\big)^T\nabla f^m(g^m(\bar{x}_t))\big\|\leq C_fC_g$. When the adaptive matrix $A_t$ be generated from the line 6 of Algorithm \ref{alg:1}, we have $\sqrt{\frac{1}{T}\sum_{t=1}^T\mathbb{E}\|A_t\|^2} \leq 2(C^2_fC^2_g+\rho)$.
Without loss of generality, let $k=O(1)$, $\rho=O(1)$, $c_1=O(1)$, $c_2=O(1)$, $c_3=O(1)$ and $n=O(q^3)$,
we have and $G=\tilde{O}(1)$. Let $q=T^{1/3}$ and
\begin{align}
\frac{1}{T}\sum_{t=1}^T\mathbb{E}\|\nabla F(\bar{x}_t)\| \leq \tilde{O}\Big(\frac{\sqrt{q}}{\sqrt{T}}+\frac{1}{T^{1/3}}\Big) = \tilde{O}\Big(\frac{1}{T^{1/3}}\Big)\leq \epsilon,
\end{align}
then we have $T=\tilde{O}(\epsilon^{-3})$. Since our AdaMFCGD algorithm requires $2$ samples at each iteration expect for the first iteration requires $2q$ samples, it has a sample complexity of $2q+2T = \tilde{O}(\epsilon^{-3})$. Thus, our AdaMFCGD algorithm requires $\tilde{O}(\epsilon^{-3})$ sample (or gradient) complexity
and $\frac{T}{q} = T^{2/3}=\tilde{O}(\epsilon^{-2})$ communication complexity to find an $\epsilon$-stationary point of the distributed composition problem \eqref{eq:1}.
\end{remark}

\begin{remark}
From Theorem \ref{th:1}, our AdaMFCGD algorithm simultaneously have lower sample and communication complexities than the existing federated compositional optimization algorithms (Please see Table \ref{tab:1}).
Moreover, our AdaMFCGD algorithm simultaneously have lower sample and communication complexities than the existing adaptive single-level FL algorithms such as the local-AMSGrad~\citep{chen2020toward} algorithm that needs sample complexity of $O(\epsilon^{-4})$ and communication complexity of $O(\epsilon^{-3})$ for finding an $\epsilon$-stationary point of the distributed single-level optimization problem, i.e., the above problem \eqref{eq:1} with $g^m(x)=x$ for all $m\in [M]$.
\end{remark}

\subsection{ Convergence Properties of MFCGD Algorithm }
In this subsection, we provide the convergence properties of our non-adaptive \textbf{MFCGD} algorithm, i.e.,
set $A_t=I_d$ for all $t\geq 1$.

\begin{theorem} \label{th:2}
Assume the sequence $\{\bar{x}_t\}_{t=1}^T$ be generated from \textbf{MFCGD} algorithm, i.e., $A_t=I_d$ for all $t\geq 1$ in Algorithm \ref{alg:1}.
 Under the above Assumptions, and let $\eta_t=\frac{k}{(n+t)^{1/3}}$ for all $t\geq 0$, $\alpha_{t+1}=c_1\eta_t^2$, $\beta_{t+1}=c_2\eta_t^2$, $\varrho_{t+1}=c_3\eta_t^2$, $n \geq \max\big(2, k^3, (c_1k)^3, (c_2k)^3, (c_3k)^3, (24k\gamma qL_{fg}C_{fg})^3 \big)$, $k>0$,
 $c_1 \geq \frac{2}{3k^3} + B$, $c_2 \geq \frac{2}{3k^3} + 5C^2_f$, $c^2_1 + c^2_2 \leq \frac{(24)^4q^2\gamma^4L^4_{fg}C^4_{fg}}{9}$, $c_3 \geq \frac{2}{3k^3} + 5C^2_g$, $ \frac{(c^2_1+c^2_3)^{1/4}}{12\sqrt{5q} L_{fg}C_{fg}} \leq \gamma \leq \min\Big(\frac{3 qL_{fg}C_{fg}}{4(C^2_g+L^2_g+ 2L^2_fC^2_g)}, \frac{n^{1/3}}{2Lk} \Big)$, $B\geq 20C_g^2L^2_f + \frac{c^2_2C^2_gL^2_f}{216q^3\gamma^3L^3_{fg}C^3_{fg}} + \frac{\Theta(c_1^2+c_3^2)}{30q^2\gamma^4C^2_{fg}L^2_{fg}C^2_g}$, $\Theta = \Big(5C^2_fL^2_g + \frac{c^2_2C^2_gL^2_f}{864q^3\gamma^3L^3_{fg}C^3_{fg}} \Big)\frac{1}{(24)^2L^2_{fg}C^2_{fg}}+\frac{\gamma}{6qL_{fg}C_{fg}}\Big(C_g^2 + L_g^2 + 2L^2_fC^2_g\Big)$ and $\Theta+\frac{BC^2_g}{(24)^2L^2_{fg}C^2_{fg}}\leq \frac{5}{48}$, we have
\begin{align}
\frac{1}{T}\sum_{t=1}^T\mathbb{E}\|\nabla F(\bar{x}_t)\|  \leq \frac{\sqrt{2G}n^{1/6}}{T^{1/2}} + \frac{\sqrt{2G}}{T^{1/3}},
\end{align}
where $C^2_{fg}=\max(C^2_f,C^2_g)$, $L^2_{fg}=L^2_fC^2_g+L^2_g$, $G = \frac{4(F(\bar{x}_1) - F^*)}{k\gamma} + \frac{12n^{1/3}\sigma^2}{qk^2} + 4k^2\Big(\frac{\hat{\delta}^2}{4\gamma^2 L^2_{fg}} + \frac{\big( c^2_1 + c^2_2 + c^2_3 \big)\sigma^2}{3\gamma qL_{fg}C_{fg}}\Big)\ln(n+T)$  and $\hat{\delta}^2 =  2c_1^2 L^2_f\sigma^2 + c_3^2\sigma^2 + 4c_3^2\delta^2_f + 4c_3^2 L^2_f\delta^2_g + c^2_2\sigma^2 + 3c^2_2\delta_{g}^2$.
\end{theorem}

\begin{remark}
The proof of Theorem~\ref{th:2} can totally follow the proofs of the above Theorem~\ref{th:1}
with the parameter $\rho=1$.
Without loss of generality, let $k=O(1)$, $c_1=O(1)$, $c_2=O(1)$, $c_3=O(1)$ and $n=O(q^3)$,
we have and $G=\tilde{O}(1)$. Let $q=T^{1/3}$ and
\begin{align}
\frac{1}{T}\sum_{t=1}^T\mathbb{E}\|\nabla F(\bar{x}_t)\| \leq \tilde{O}\Big(\frac{\sqrt{q}}{\sqrt{T}}+\frac{1}{T^{1/3}}\Big) = \tilde{O}\Big(\frac{1}{T^{1/3}}\Big)\leq \epsilon,
\end{align}
then we have $T=\tilde{O}(\epsilon^{-3})$. Since our MFCGD algorithm requires $2$ samples at each iteration expect for the first iteration requires $2q$ samples, it has a sample complexity of $2q+2T = \tilde{O}(\epsilon^{-3})$.
As the above AdaMFCGD algorithm, our MFCGD algorithm also obtain lower sample  complexity of $\tilde{O}(\epsilon^{-3})$
and communication complexity of $\tilde{O}(\epsilon^{-2})$ in finding an $\epsilon$-stationary point of the problem \eqref{eq:1}.
\end{remark}

\section{Numerical Experiments}
In this section, we apply some numerical experiments to demonstrate the efficiency of our MFCGD and AdaMFCGD algorithms on robust federated learning and distributed meta learning tasks.
In the experiments, we compare our algorithms with the existing algorithms in Table \ref{tab:1} for solving distributed composition optimization problems.

\subsection{ Robust Federated Learning }

\subsection{ Task-Distributed Meta Learning }

\section{Conclusions}
In the paper, we proposed a class of faster momentum-based federated compositional gradient descent algorithms (i.e., MFCGD and AdaMFCGD) to solve the nonconvex distributed composition problems. Our adaptive algorithm (i.e., AdaMFCGD) uses a unified adaptive matrix to flexibly incorporate various adaptive learning rates. Moreover, we established a solid convergence analysis framework for our algorithms, and proved that our methods obtain lower sample and communication complexities simultaneously than the existing federated composition optimization methods.


\small

\bibliographystyle{plainnat}

\bibliography{AdaFCGD}

\begin{thebibliography}{50}
\providecommand{\natexlab}[1]{#1}
\providecommand{\url}[1]{\texttt{#1}}
\expandafter\ifx\csname urlstyle\endcsname\relax
  \providecommand{\doi}[1]{doi: #1}\else
  \providecommand{\doi}{doi: \begingroup \urlstyle{rm}\Url}\fi

\bibitem[Andrychowicz et~al.(2016)Andrychowicz, Denil, Colmenarejo, Hoffman,
  Pfau, Schaul, Shillingford, and de~Freitas]{andrychowicz2016learning}
M.~Andrychowicz, M.~Denil, S.~G. Colmenarejo, M.~W. Hoffman, D.~Pfau,
  T.~Schaul, B.~Shillingford, and N.~de~Freitas.
\newblock Learning to learn by gradient descent by gradient descent.
\newblock In \emph{Proceedings of the 30th International Conference on Neural
  Information Processing Systems}, pages 3988--3996, 2016.

\bibitem[Censor and Lent(1981)]{censor1981iterative}
Y.~Censor and A.~Lent.
\newblock An iterative row-action method for interval convex programming.
\newblock \emph{Journal of Optimization theory and Applications}, 34\penalty0
  (3):\penalty0 321--353, 1981.

\bibitem[Censor and Zenios(1992)]{censor1992proximal}
Y.~Censor and S.~A. Zenios.
\newblock Proximal minimization algorithm withd-functions.
\newblock \emph{Journal of Optimization Theory and Applications}, 73\penalty0
  (3):\penalty0 451--464, 1992.

\bibitem[Chen et~al.(2020{\natexlab{a}})Chen, Chen, Zhou, and
  Kailkhura]{chen2020fedcluster}
C.~Chen, Z.~Chen, Y.~Zhou, and B.~Kailkhura.
\newblock Fedcluster: Boosting the convergence of federated learning via
  cluster-cycling.
\newblock In \emph{2020 IEEE International Conference on Big Data (Big Data)},
  pages 5017--5026. IEEE, 2020{\natexlab{a}}.

\bibitem[Chen et~al.(2020{\natexlab{b}})Chen, Sun, and Yin]{chen2020solving}
T.~Chen, Y.~Sun, and W.~Yin.
\newblock Solving stochastic compositional optimization is nearly as easy as
  solving stochastic optimization.
\newblock \emph{arXiv preprint arXiv:2008.10847}, 2020{\natexlab{b}}.

\bibitem[Chen et~al.(2019)Chen, Liu, Sun, and Hong]{chen2019convergence}
X.~Chen, S.~Liu, R.~Sun, and M.~Hong.
\newblock On the convergence of a class of adam-type algorithms for non-convex
  optimization.
\newblock In \emph{7th International Conference on Learning Representations
  (ICLR)}, 2019.

\bibitem[Chen et~al.(2020{\natexlab{c}})Chen, Li, and Li]{chen2020toward}
X.~Chen, X.~Li, and P.~Li.
\newblock Toward communication efficient adaptive gradient method.
\newblock In \emph{Proceedings of the 2020 ACM-IMS on Foundations of Data
  Science Conference}, pages 119--128, 2020{\natexlab{c}}.

\bibitem[Cutkosky and Orabona(2019)]{cutkosky2019momentum}
A.~Cutkosky and F.~Orabona.
\newblock Momentum-based variance reduction in non-convex sgd.
\newblock \emph{Advances in neural information processing systems}, 32, 2019.

\bibitem[Deng and Mahdavi(2021)]{deng2021local}
Y.~Deng and M.~Mahdavi.
\newblock Local stochastic gradient descent ascent: Convergence analysis and
  communication efficiency.
\newblock In \emph{International Conference on Artificial Intelligence and
  Statistics}, pages 1387--1395. PMLR, 2021.

\bibitem[Deng et~al.(2020{\natexlab{a}})Deng, Kamani, and
  Mahdavi]{deng2020adaptive}
Y.~Deng, M.~M. Kamani, and M.~Mahdavi.
\newblock Adaptive personalized federated learning.
\newblock \emph{arXiv preprint arXiv:2003.13461}, 2020{\natexlab{a}}.

\bibitem[Deng et~al.(2020{\natexlab{b}})Deng, Kamani, and
  Mahdavi]{deng2020distributionally}
Y.~Deng, M.~M. Kamani, and M.~Mahdavi.
\newblock Distributionally robust federated averaging.
\newblock \emph{Advances in Neural Information Processing Systems},
  33:\penalty0 15111--15122, 2020{\natexlab{b}}.

\bibitem[Duchi et~al.(2011)Duchi, Hazan, and Singer]{duchi2011adaptive}
J.~Duchi, E.~Hazan, and Y.~Singer.
\newblock Adaptive subgradient methods for online learning and stochastic
  optimization.
\newblock \emph{Journal of machine learning research}, 12\penalty0 (7), 2011.

\bibitem[Fallah et~al.(2020)Fallah, Mokhtari, and
  Ozdaglar]{fallah2020personalized}
A.~Fallah, A.~Mokhtari, and A.~Ozdaglar.
\newblock Personalized federated learning: A meta-learning approach.
\newblock \emph{arXiv preprint arXiv:2002.07948}, 2020.

\bibitem[Fang et~al.(2018)Fang, Li, Lin, and Zhang]{fang2018spider}
C.~Fang, C.~J. Li, Z.~Lin, and T.~Zhang.
\newblock Spider: Near-optimal non-convex optimization via stochastic
  path-integrated differential estimator.
\newblock In \emph{Advances in Neural Information Processing Systems}, pages
  689--699, 2018.

\bibitem[Finn et~al.(2017)Finn, Abbeel, and Levine]{finn2017model}
C.~Finn, P.~Abbeel, and S.~Levine.
\newblock Model-agnostic meta-learning for fast adaptation of deep networks.
\newblock In \emph{International Conference on Machine Learning}, pages
  1126--1135. PMLR, 2017.

\bibitem[Gao et~al.(2022)Gao, Li, and Huang]{gao2022convergence}
H.~Gao, J.~Li, and H.~Huang.
\newblock On the convergence of local stochastic compositional gradient descent
  with momentum.
\newblock In \emph{International Conference on Machine Learning}, pages
  7017--7035. PMLR, 2022.

\bibitem[Ghadimi et~al.(2016)Ghadimi, Lan, and Zhang]{ghadimi2016mini}
S.~Ghadimi, G.~Lan, and H.~Zhang.
\newblock Mini-batch stochastic approximation methods for nonconvex stochastic
  composite optimization.
\newblock \emph{Mathematical Programming}, 155\penalty0 (1-2):\penalty0
  267--305, 2016.

\bibitem[Ghadimi et~al.(2020)Ghadimi, Ruszczynski, and Wang]{ghadimi2020single}
S.~Ghadimi, A.~Ruszczynski, and M.~Wang.
\newblock A single timescale stochastic approximation method for nested
  stochastic optimization.
\newblock \emph{SIAM Journal on Optimization}, 30\penalty0 (1):\penalty0
  960--979, 2020.

\bibitem[Huang and Gao(2022)]{huang2022riemannian}
F.~Huang and S.~Gao.
\newblock Riemannian gradient methods for stochastic composition problems.
\newblock \emph{Neural Networks}, 153:\penalty0 224--234, 2022.

\bibitem[Huang et~al.(2021{\natexlab{a}})Huang, Li, and
  Huang]{huang2021compositional}
F.~Huang, J.~Li, and H.~Huang.
\newblock Compositional federated learning: Applications in distributionally
  robust averaging and meta learning.
\newblock \emph{arXiv preprint arXiv:2106.11264}, 2021{\natexlab{a}}.

\bibitem[Huang et~al.(2021{\natexlab{b}})Huang, Li, and Huang]{huang2021super}
F.~Huang, J.~Li, and H.~Huang.
\newblock Super-adam: faster and universal framework of adaptive gradients.
\newblock \emph{Advances in Neural Information Processing Systems},
  34:\penalty0 9074--9085, 2021{\natexlab{b}}.

\bibitem[Huo et~al.(2018)Huo, Gu, Liu, and Huang]{huo2018accelerated}
Z.~Huo, B.~Gu, J.~Liu, and H.~Huang.
\newblock Accelerated method for stochastic composition optimization with
  nonsmooth regularization.
\newblock In \emph{Proceedings of the AAAI Conference on Artificial
  Intelligence}, volume~32, 2018.

\bibitem[Jiang et~al.(2022)Jiang, Wang, Wang, Zhang, and
  Yang]{jiang2022optimal}
W.~Jiang, B.~Wang, Y.~Wang, L.~Zhang, and T.~Yang.
\newblock Optimal algorithms for stochastic multi-level compositional
  optimization.
\newblock \emph{arXiv preprint arXiv:2202.07530}, 2022.

\bibitem[Kairouz et~al.(2019)Kairouz, McMahan, Avent, Bellet, Bennis, Bhagoji,
  Bonawitz, Charles, Cormode, Cummings, et~al.]{kairouz2019advances}
P.~Kairouz, H.~B. McMahan, B.~Avent, A.~Bellet, M.~Bennis, A.~N. Bhagoji,
  K.~Bonawitz, Z.~Charles, G.~Cormode, R.~Cummings, et~al.
\newblock Advances and open problems in federated learning.
\newblock \emph{arXiv preprint arXiv:1912.04977}, 2019.

\bibitem[Karimireddy et~al.(2019)Karimireddy, Rebjock, Stich, and
  Jaggi]{karimireddy2019error}
S.~P. Karimireddy, Q.~Rebjock, S.~Stich, and M.~Jaggi.
\newblock Error feedback fixes signsgd and other gradient compression schemes.
\newblock In \emph{International Conference on Machine Learning}, pages
  3252--3261. PMLR, 2019.

\bibitem[Karimireddy et~al.(2020)Karimireddy, Kale, Mohri, Reddi, Stich, and
  Suresh]{karimireddy2020scaffold}
S.~P. Karimireddy, S.~Kale, M.~Mohri, S.~Reddi, S.~Stich, and A.~T. Suresh.
\newblock Scaffold: Stochastic controlled averaging for federated learning.
\newblock In \emph{International Conference on Machine Learning}, pages
  5132--5143. PMLR, 2020.

\bibitem[Khanduri et~al.(2021)Khanduri, Sharma, Yang, Hong, Liu, Rajawat, and
  Varshney]{khanduri2021stem}
P.~Khanduri, P.~Sharma, H.~Yang, M.~Hong, J.~Liu, K.~Rajawat, and P.~Varshney.
\newblock Stem: A stochastic two-sided momentum algorithm achieving
  near-optimal sample and communication complexities for federated learning.
\newblock \emph{Advances in Neural Information Processing Systems},
  34:\penalty0 6050--6061, 2021.

\bibitem[Kingma and Ba(2014)]{kingma2014adam}
D.~P. Kingma and J.~Ba.
\newblock Adam: A method for stochastic optimization.
\newblock \emph{arXiv preprint arXiv:1412.6980}, 2014.

\bibitem[Li et~al.(2021)Li, Hu, Beirami, and Smith]{li2021ditto}
T.~Li, S.~Hu, A.~Beirami, and V.~Smith.
\newblock Ditto: Fair and robust federated learning through personalization.
\newblock In \emph{International Conference on Machine Learning}, pages
  6357--6368. PMLR, 2021.

\bibitem[Li et~al.(2019)Li, Huang, Yang, Wang, and Zhang]{li2019convergence}
X.~Li, K.~Huang, W.~Yang, S.~Wang, and Z.~Zhang.
\newblock On the convergence of fedavg on non-iid data.
\newblock \emph{arXiv preprint arXiv:1907.02189}, 2019.

\bibitem[Lin et~al.(2018)Lin, Fan, Wang, and Jordan]{lin2018improved}
T.~Lin, C.~Fan, M.~Wang, and M.~I. Jordan.
\newblock Improved sample complexity for stochastic compositional variance
  reduced gradient.
\newblock \emph{arXiv preprint arXiv:1806.00458}, 2018.

\bibitem[Loshchilov and Hutter(2018)]{loshchilov2018decoupled}
I.~Loshchilov and F.~Hutter.
\newblock Decoupled weight decay regularization.
\newblock In \emph{International Conference on Learning Representations}, 2018.

\bibitem[McMahan et~al.(2017)McMahan, Moore, Ramage, Hampson, and
  y~Arcas]{mcmahan2017communication}
B.~McMahan, E.~Moore, D.~Ramage, S.~Hampson, and B.~A. y~Arcas.
\newblock Communication-efficient learning of deep networks from decentralized
  data.
\newblock In \emph{Artificial Intelligence and Statistics}, pages 1273--1282.
  PMLR, 2017.

\bibitem[Mohri et~al.(2019)Mohri, Sivek, and Suresh]{mohri2019agnostic}
M.~Mohri, G.~Sivek, and A.~T. Suresh.
\newblock Agnostic federated learning.
\newblock In \emph{International Conference on Machine Learning}, pages
  4615--4625. PMLR, 2019.

\bibitem[Nguyen et~al.(2017)Nguyen, Liu, Scheinberg, and
  Tak{\'a}{\v{c}}]{nguyen2017sarah}
L.~M. Nguyen, J.~Liu, K.~Scheinberg, and M.~Tak{\'a}{\v{c}}.
\newblock Sarah: A novel method for machine learning problems using stochastic
  recursive gradient.
\newblock In \emph{International Conference on Machine Learning}, pages
  2613--2621. PMLR, 2017.

\bibitem[Reddi et~al.(2020)Reddi, Charles, Zaheer, Garrett, Rush,
  Kone{\v{c}}n{\`y}, Kumar, and McMahan]{reddi2020adaptive}
S.~Reddi, Z.~Charles, M.~Zaheer, Z.~Garrett, K.~Rush, J.~Kone{\v{c}}n{\`y},
  S.~Kumar, and H.~B. McMahan.
\newblock Adaptive federated optimization.
\newblock \emph{arXiv preprint arXiv:2003.00295}, 2020.

\bibitem[Reddi et~al.(2019)Reddi, Kale, and Kumar]{reddi2019convergence}
S.~J. Reddi, S.~Kale, and S.~Kumar.
\newblock On the convergence of adam and beyond.
\newblock \emph{arXiv preprint arXiv:1904.09237}, 2019.

\bibitem[Reisizadeh et~al.(2020)Reisizadeh, Farnia, Pedarsani, and
  Jadbabaie]{reisizadeh2020robust}
A.~Reisizadeh, F.~Farnia, R.~Pedarsani, and A.~Jadbabaie.
\newblock Robust federated learning: The case of affine distribution shifts.
\newblock In \emph{NeurIPS}, 2020.

\bibitem[Stich(2019)]{stich2019local}
S.~U. Stich.
\newblock Local sgd converges fast and communicates little.
\newblock In \emph{International Conference on Learning Representations
  (ICLR)}, 2019.

\bibitem[Tarzanagh et~al.(2022)Tarzanagh, Li, Thrampoulidis, and
  Oymak]{tarzanagh2022fednest}
D.~A. Tarzanagh, M.~Li, C.~Thrampoulidis, and S.~Oymak.
\newblock Fednest: Federated bilevel, minimax, and compositional optimization.
\newblock \emph{arXiv preprint arXiv:2205.02215}, 2022.

\bibitem[Tran-Dinh et~al.(2022)Tran-Dinh, Pham, Phan, and
  Nguyen]{tran2022hybrid}
Q.~Tran-Dinh, N.~H. Pham, D.~T. Phan, and L.~M. Nguyen.
\newblock A hybrid stochastic optimization framework for composite nonconvex
  optimization.
\newblock \emph{Mathematical Programming}, 191\penalty0 (2):\penalty0
  1005--1071, 2022.

\bibitem[Tutunov et~al.(2020)Tutunov, Li, Wang, and
  Bou-Ammar]{tutunov2020compositional}
R.~Tutunov, M.~Li, J.~Wang, and H.~Bou-Ammar.
\newblock Compositional adam: An adaptive compositional solver.
\newblock \emph{arXiv preprint arXiv:2002.03755}, 2020.

\bibitem[Wang et~al.(2021)Wang, Yuan, Ying, and Yang]{wang2021memory}
B.~Wang, Z.~Yuan, Y.~Ying, and T.~Yang.
\newblock Memory-based optimization methods for model-agnostic meta-learning.
\newblock \emph{arXiv preprint arXiv:2106.04911}, 2021.

\bibitem[Wang et~al.(2017{\natexlab{a}})Wang, Fang, and
  Liu]{wang2017stochastic}
M.~Wang, E.~X. Fang, and H.~Liu.
\newblock Stochastic compositional gradient descent: algorithms for minimizing
  compositions of expected-value functions.
\newblock \emph{Mathematical Programming}, 161\penalty0 (1-2):\penalty0
  419--449, 2017{\natexlab{a}}.

\bibitem[Wang et~al.(2017{\natexlab{b}})Wang, Liu, and
  Fang]{wang2017accelerating}
M.~Wang, J.~Liu, and E.~X. Fang.
\newblock Accelerating stochastic composition optimization.
\newblock \emph{The Journal of Machine Learning Research}, 18\penalty0
  (1):\penalty0 3721--3743, 2017{\natexlab{b}}.

\bibitem[Xu et~al.(2021)Xu, Glicksberg, Su, Walker, Bian, and
  Wang]{xu2021federated}
J.~Xu, B.~S. Glicksberg, C.~Su, P.~Walker, J.~Bian, and F.~Wang.
\newblock Federated learning for healthcare informatics.
\newblock \emph{Journal of Healthcare Informatics Research}, 5\penalty0
  (1):\penalty0 1--19, 2021.

\bibitem[Yang et~al.(2021)Yang, Zhang, Hao, Spell, and Carin]{yang2021flop}
Q.~Yang, J.~Zhang, W.~Hao, G.~P. Spell, and L.~Carin.
\newblock Flop: Federated learning on medical datasets using partial networks.
\newblock In \emph{Proceedings of the 27th ACM SIGKDD Conference on Knowledge
  Discovery \& Data Mining}, pages 3845--3853, 2021.

\bibitem[Yuan and Ma(2020)]{yuan2020federated}
H.~Yuan and T.~Ma.
\newblock Federated accelerated stochastic gradient descent.
\newblock \emph{Advances in Neural Information Processing Systems},
  33:\penalty0 5332--5344, 2020.

\bibitem[Yuan et~al.(2022)Yuan, Guo, Chawla, and Yang]{yuan2022compositional}
Z.~Yuan, Z.~Guo, N.~Chawla, and T.~Yang.
\newblock Compositional training for end-to-end deep auc maximization.
\newblock In \emph{International Conference on Learning Representations}, 2022.

\bibitem[Zhang and Xiao(2019)]{zhang2019multi}
J.~Zhang and L.~Xiao.
\newblock Multi-level composite stochastic optimization via nested variance
  reduction.
\newblock \emph{arXiv preprint arXiv:1908.11468}, 2019.

\end{thebibliography}

\newpage

\appendix

\section{Appendix}
In this section, we provide the detailed convergence analysis of our algorithms.

We first introduce some useful notations: $\bar{w}_t = \frac{1}{M} \sum_{m=1}^{M} w^m_t$,
$\bar{x}_t = \frac{1}{M} \sum_{m=1}^{M} x^m_t$,
\begin{align}
& F(x) = \frac{1}{M}\sum_{m=1}^Mf^m(g^m(x)), \quad \nabla F(x) = \frac{1}{M}\sum_{m=1}^{M}\big(\nabla g^m(x)\big)^T \nabla f^m(g^m(x)). \nonumber
\end{align}

Next, we review and provide some useful lemmas.

\begin{lemma} \label{lem:A1}
Given $M$ vectors $\{u^m\}_{m=1}^M$, the following inequalities satisfy: $||u^m + u^j||^2 \leq (1+c)||u^m||^2 + (1+\frac{1}{c})||u^j||^2$ for any $c > 0$, and
$||\sum_{m=1}^M u^m||^2 \le M\sum_{m=1}^{M} ||u^m||^2$.
\end{lemma}

\begin{lemma} \label{lem:A2}
Given a finite sequence $\{u^{m}\}_{m=1}^M$, and $\bar{u} = \frac{1}{M}\sum_{m=1}^M u^{m}$,
the following inequality satisfies $\sum_{m=1}^M \|u^{m} - \bar{u}\|^2 \leq \sum_{m=1}^M \|u^{m}\|^2$.
\end{lemma}

Given a $\rho$-strongly convex function $\varphi(x)$, we define a prox-function (Bregman distance) \cite{censor1981iterative,censor1992proximal}
associated with $\varphi(x)$ as follows:
\begin{align}
 D(z,x) = \varphi(z) - \big[\varphi(x) + \langle\nabla \varphi(x),z-x\rangle\big].
\end{align}
Then we define a generalized projection problem as in \cite{ghadimi2016mini}:
\begin{align} \label{eq:A1}
 x^+ = \arg\min_{z\in \mathcal{X}} \big\{\langle z,w\rangle + \frac{1}{\gamma}D(z,x) + h(z)\big\},
\end{align}
where $\mathcal{X} \subseteq \mathbb{R}^d$, $w \in \mathbb{R}^d$ and $\gamma>0$.
In the paper, we consider $h(x)=0$. Meanwhile, we also define a generalized projected gradient (a.k.a., gradient mapping):
\begin{align} \label{eq:A2}
 \mathcal{G}_{\mathcal{X}}(x,w,\gamma) = \frac{x-x^+}{\gamma}.
\end{align}

\begin{lemma} \label{lem:A3}
(Lemma 1 in \cite{ghadimi2016mini})
Let $x^+$ be given in \eqref{eq:A1}. Then, for any $x\in \mathcal{X}$, $w\in \mathbb{R}^d$ and $\gamma >0$,
we have
\begin{align}
 \langle w,  \mathcal{G}_{\mathcal{X}}(x,w,\gamma)\rangle \geq \rho \|\mathcal{G}_{\mathcal{X}}(x,w,\gamma)\|^2
 + \frac{1}{\gamma}\big[h(x^+)-h(x)\big],
\end{align}
where $\rho>0$ depends on $\rho$-strongly convex function $\varphi(x)$.
\end{lemma}
When $h(x)=0$, in the above lemma \ref{lem:A3}, we have
\begin{align}
 \langle w, \mathcal{G}_{\mathcal{X}}(x,w,\gamma)\rangle \geq \rho \|\mathcal{G}_{\mathcal{X}}(x,w,\gamma)\|^2.
\end{align}

\begin{lemma} \label{lem:A5}
(Restatement of Lemma 1)
Given the above Assumptions \ref{ass:1}-\ref{ass:2},
the function $F(x)$ is $L$-smooth, i.e., for any $x_1,x_2\in \mathbb{R}^d$, we have
\begin{align}
 \|\nabla F(x_1) - \nabla F(x_2)\|^2 \leq L^2\|x_1-x_2\|^2,
\end{align}
where $L = \sqrt{2C^2_fL^2_g + 2C^4_gL^2_f}$.
\end{lemma}

\begin{proof}
Based on Assumptions \ref{ass:1}-\ref{ass:2}, the deterministic functions $f^m(y)=\mathbb{E}\big[f^m(y;\xi^m)\big]$ $g^m(x)=\mathbb{E}\big[f^m(x;\zeta^m)\big]$ and its gradients also satisfy the Lipschitz gradients and bounded gradients.
For example, for any $y_1,y_2 \in \mathbb{R}^n$
\begin{align}
 \big\|\nabla f^m(y_1)-\nabla f^m(y_1)\big\| & = \big\|\mathbb{E}\big[\nabla f^m(y_1;\xi^m)-\nabla f^m(y_1;\xi^m)\big]\big\| \nonumber \\
 & \leq \mathbb{E}\big\|\nabla f^m(y_1;\xi^m)-\nabla f^m(y_1;\xi^m)\big\| \leq L_f\|y_1-y_2\|,
\end{align}
where the first inequality holds by Jensen's inequality, and the last inequality holds by Assumption \ref{ass:1}.

Since $F(x) = \frac{1}{M}\sum_{m=1}^Mf^m(g^m(x))$, we have
\begin{align}
 &\|\nabla F(x_1) - \nabla F(x_2)\|^2 \nonumber \\
 & = \big\| \frac{1}{M}\sum_{m=1}^{M}\big(\nabla g^m(x_1)\big)^T \nabla f^m(g^m(x_1)) - \frac{1}{M}\sum_{m=1}^{M}\big(\nabla g^m(x_2)\big)^T \nabla f^m(g^m(x_2)) \big\|^2 \nonumber \\
 & \leq \frac{1}{M}\sum_{m=1}^{M}\big\| \big(\nabla g^m(x_1)\big)^T \nabla f^m(g^m(x_1)) - \big(\nabla g^m(x_2)\big)^T \nabla f^m(g^m(x_2)) \big\|^2 \nonumber \\
 & = \frac{1}{M}\sum_{m=1}^{M}\big\| \big(\nabla g^m(x_1)\big)^T \nabla f^m(g^m(x_1)) - \big(\nabla g^m(x_2)\big)^T \nabla f^m(g^m(x_1)) + \big(\nabla g^m(x_2)\big)^T \nabla f^m(g^m(x_1)) \nonumber \\
 & \quad - \big(\nabla g^m(x_2)\big)^T \nabla f^m(g^m(x_2)) \big\|^2 \nonumber \\
 & \leq \frac{1}{M}\sum_{m=1}^{M}2C^2_f\big\|\nabla g^m(x_1) - \nabla g^m(x_2) \big\|^2 + \frac{1}{M}\sum_{m=1}^{M}2C^2_g\big\| \nabla f^m(g^m(x_1)) - \nabla f^m(g^m(x_2)) \big\|^2 \nonumber \\
 & \leq 2C^2_fL^2_g \|x_1-x_2\|^2 + 2C^4_gL^2_f\|x_1-x_2\|^2 = \big(2C^2_fL^2_g + 2C^4_gL^2_f\big)\|x_1-x_2\|^2,
\end{align}
where the second last and the last inequalities hold by Assumptions \ref{ass:1}-\ref{ass:2}.

\end{proof}

\begin{lemma} \label{lem:A6}
(Restatement of Lemma 2)
Assume the gradient estimator $\{\bar{w}_t\}_{t=1}^T$ generated from Algorithm \ref{alg:1}, where $w_t = \frac{1}{M}\sum_{m=1}^Mw^m_t$,
we have
\begin{align}
 \|\bar{w}_t - \nabla F(\bar{x}_t)\|^2
 & \leq \frac{1}{M}\sum_{m=1}^{M}\Big( 2C_f^2\|u^m_t - \nabla g^m(\bar{x}_t)\|^2 + 4C_g^2\|v^m_t - \nabla f^m(h^m_t)\|^2 + 4C_g^2L^2_f\|h^m_t - g^m(\bar{x}_t)\|^2 \Big).
\end{align}
\end{lemma}

\begin{proof}
 Since $\bar{w}_t = \frac{1}{M}\sum_{m=1}^M(u^m_t)^Tv^m_t$,
 we have
\begin{align}
& \|\bar{w}_t - \nabla F(\bar{x}_t)\|^2 \nonumber \\
& =\|\frac{1}{M}\sum_{m=1}^M(u^m_t)^Tv^m_t - \frac{1}{M}\sum_{m=1}^M\big(\nabla g^m(\bar{x}_t)\big)^T\nabla f^m(g^m(\bar{x}_t))\|^2 \nonumber \\
& = \|\frac{1}{M}\sum_{m=1}^M(u^m_t)^Tv^m_t -\frac{1}{M}\sum_{m=1}^M\big(\nabla g^m(\bar{x}_t)\big)^Tv^m_t + \frac{1}{M}\sum_{m=1}^M\big(\nabla g^m(\bar{x}_t)\big)^Tv^m_t  - \frac{1}{M}\sum_{m=1}^M\big(\nabla g^m(\bar{x}_t)\big)^T\nabla f^m(g^m(\bar{x}_t))\|^2 \nonumber \\
& \leq \frac{1}{M}\sum_{m=1}^M2C^2_f\|u^m_t - \nabla g^m(\bar{x}_t)\|^2 + \frac{1}{M}\sum_{m=1}^M2C^2_g\|v^m_t-\nabla f^m(g^m(\bar{x}_t))\|^2 \nonumber \\
& = \frac{2C^2_f}{M}\sum_{m=1}^M\|u^m_t - \nabla g^m(\bar{x}_t)\|^2 + \frac{2C^2_g}{M}\sum_{m=1}^M\|v^m_t-\nabla f^m(h^m_t)+\nabla f^m(h^m_t)-\nabla f^m(g^m(\bar{x}_t))\|^2 \nonumber \\
 & \leq \frac{2C^2_f}{M}\sum_{m=1}^M\|u^m_t - \nabla g^m(\bar{x}_t)\|^2 + \frac{4C^2_g}{M}\sum_{m=1}^M\|v^m_t-\nabla f^m(h^m_t)\|^2 + \frac{4C^2_gL^2_f}{M}\sum_{m=1}^M\|h^m_t-g^m(\bar{x}_t)\|^2,
\end{align}
where the first inequality is due to Assumptions \ref{ass:1}-\ref{ass:2} and the above Lemma \ref{lem:A1}.

\end{proof}

\begin{lemma} \label{lem:A7}
 Suppose that the sequence $\big\{\bar{x}_t\big\}_{t=1}^T$ be generated from Algorithm \ref{alg:1}, where $\bar{x}_t = \frac{1}{M}\sum_{m=1}^Mx^m_t$.
 Let $0< \gamma \leq \frac{\rho}{2L\eta_t}$,
 then we have
 \begin{align}
   F(\bar{x}_{t+1}) & \leq F(\bar{x}_t) + \frac{1}{M}\sum_{m=1}^{M}\Big( \frac{2C_f^2\eta_t\gamma}{\rho}\|u^m_t - \nabla g^m(\bar{x}_t)\|^2 + \frac{4C_g^2\eta_t\gamma}{\rho}\|v^m_t - \nabla f^m(h^m_t)\|^2  \nonumber \\
 & \qquad + \frac{4C_g^2L^2_f\eta_t\gamma}{\rho}\|h^m_t - g^m(\bar{x}_t)\|^2  \Big) -\frac{\rho}{2\eta_t\gamma}\|\bar{x}_{t+1}-\bar{x}_t\|^2.
 \end{align}
\end{lemma}

\begin{proof}
According to the above Lemma \ref{lem:A5}, the function $F(x)$ is $L$-smooth. Thus we have
 \begin{align} \label{eq:B1}
  F(\bar{x}_{t+1}) & \leq F(\bar{x}_t) + \langle\nabla F(\bar{x}_t), \bar{x}_{t+1}-\bar{x}_t\rangle + \frac{L}{2}\|\bar{x}_{t+1}-\bar{x}_t\|^2 \\
  & = F(\bar{x}_t) + \underbrace{\langle \bar{w}_t,\bar{x}_{t+1}-\bar{x}_t\rangle}_{=T_1} + \underbrace{\langle \nabla F(\bar{x}_t)-\bar{w}_t,\bar{x}_{t+1}-\bar{x}_t\rangle}_{=T_2} + \frac{L}{2}\|\bar{x}_{t+1}-\bar{x}_t\|^2. \nonumber
 \end{align}

According to Assumption \ref{ass:5}, i.e., $A_t\succ \rho I_d$ for any $t\geq 1$,
the mirror function $\varphi_t(x)=\frac{1}{2}x^TA_t x$ is $\rho$-strongly convex, then we can define a Bregman distance as in \cite{ghadimi2016mini},
\begin{align}
 D_t(x,\bar{x}_t) = \varphi_t(x) - \big[ \varphi_t(\bar{x}_t) + \langle\nabla \varphi_t(\bar{x}_t), x-\bar{x}_t\rangle\big] = \frac{1}{2}(x-\bar{x}_t)^TA_t(x-\bar{x}_t).
\end{align}
When $t=s_t=q\lfloor t/q\rfloor+1$, according to the line 7 of Algorithm \ref{alg:1}, we have
$\bar{x}_{t+1}=\arg\min_{x\in \mathbb{R}^d}\big\{\langle \bar{w}_t, x\rangle + \frac{1}{2\eta_t\gamma}(x-\bar{x}_t)^T A_t (x-\bar{x}_t)\big\}$.
By using Lemma 1 in \cite{ghadimi2016mini} to the problem $\bar{x}_{t+1} = \arg\min_{x \in \mathbb{R}^d} \big\{\langle \bar{w}_t, x\rangle + \frac{1}{2\eta_t\gamma}(x-\bar{x}_t)^T A_t (x-\bar{x}_t)\big\}$, we can obtain
\begin{align}
  \langle \bar{w}_t, \frac{1}{\eta_t\gamma}(\bar{x}_t - \bar{x}_{t+1})\rangle \geq \rho\|\frac{1}{\eta_t\gamma}(\bar{x}_t - \bar{x}_{t+1})\|^2.
\end{align}
When $t\in (s_t,s_t+q)$, according to the line 11 of Algorithm \ref{alg:1}, we have $x^m_{t+1}
=\arg\min_{x\in \mathbb{R}^d}\big\{\langle w^m_t, x\rangle + \frac{1}{2\eta_t\gamma}(x-x^m_t)^T A_t (x-x^m_t)\big\}$.
Similarly, we have
\begin{align}
  \langle w^m_t, \frac{1}{\eta_t\gamma}(x^m_t - x^m_{t+1})\rangle \geq \rho\|\frac{1}{\eta_t\gamma}(x^m_t - x^m_{t+1})\|^2.
\end{align}
Then we have
\begin{align} \label{eq:B2}
  \frac{1}{M}\sum_{m=1}^M\langle w^m_t, \frac{1}{\eta_t\gamma}(x^m_t - x^m_{t+1})\rangle & \geq \rho\frac{1}{M}\sum_{m=1}^M\|\frac{1}{\eta_t\gamma}(x^m_t - x^m_{t+1})\|^2 \nonumber \\
  & \geq \rho\|\frac{1}{\eta_t\gamma}\frac{1}{M}\sum_{m=1}^M(x^m_t - x^m_{t+1})\|^2 = \rho\|\frac{1}{\eta_t\gamma}(\bar{x}_t - \bar{x}_{t+1})\|^2.
\end{align}
Thus we have
\begin{align} \label{eq:B3}
  \langle w^m_t, \frac{1}{\eta_t\gamma}(\bar{x}_t - \bar{x}_{t+1})\rangle \geq \rho\|\frac{1}{\eta_t\gamma}(\bar{x}_t - \bar{x}_{t+1})\|^2.
\end{align}
Since $\bar{w}_t=\frac{1}{M}\sum_{m=1}^Mw^m_t$,
 averaging the above inequality \eqref{eq:B3} from $m=1$ to $M$, we can obtain
\begin{align} \label{eq:B4}
  \langle \bar{w}_t, \frac{1}{\eta_t\gamma}(\bar{x}_t - \bar{x}_{t+1})\rangle = \frac{1}{M}\sum_{m=1}^M\langle w^m_t, \frac{1}{\eta_t\gamma}(\bar{x}_t - \bar{x}_{t+1})\rangle \geq\rho\frac{1}{M}\sum_{m=1}^M\|\frac{1}{\eta_t\gamma}(\bar{x}_t - \bar{x}_{t+1})\|^2=\rho\|\frac{1}{\eta_t\gamma}(\bar{x}_t - \bar{x}_{t+1})\|^2.
\end{align}
Then we have for any $t\in [s_t,s_t+q)$,
\begin{align}
  T_1=\langle \bar{w}_t, \bar{x}_{t+1}-\bar{x}_t\rangle \leq -\frac{\rho }{\eta_t\gamma }\|\bar{x}_{t+1}-\bar{x}_t\|^2.
\end{align}
Since $s_t=q\lfloor t/q\rfloor+1$ and all $t\in [s_t,s_t+q)$, clearly, we have, for all $t\geq 1$
\begin{align} \label{eq:B5}
  T_1=\langle \bar{w}_t, \bar{x}_{t+1}-\bar{x}_t\rangle \leq -\frac{\rho }{\eta_t\gamma }\|\bar{x}_{t+1}-\bar{x}_t\|^2.
\end{align}
Next, consider the bound of the term $T_2$, we have
\begin{align} \label{eq:B6}
  T_2 & = \langle \nabla F(\bar{x}_t)-\bar{w}_t,\bar{x}_{t+1}-\bar{x}_t\rangle \nonumber \\
  & \leq \|\nabla F(\bar{x}_t)-\bar{w}_t\|\cdot\|\bar{x}_{t+1}-\bar{x}_t\| \nonumber \\
  & \leq \frac{\eta_t\gamma}{\rho}\|\nabla F(\bar{x}_t)-\bar{w}_t\|^2+\frac{\rho}{4\eta_t\gamma}\|\bar{x}_{t+1}-\bar{x}_t\|^2,
\end{align}
where the first inequality is due to the Cauchy-Schwarz inequality and the last is due to Young's inequality.
By combining the above inequalities \eqref{eq:B1}, \eqref{eq:B5} with \eqref{eq:B6},
we obtain
\begin{align} \label{eq:B7}
  F(\bar{x}_{t+1}) &\leq F(\bar{x}_t) + \langle \nabla F(\bar{x}_t)-\bar{w}_t,\bar{x}_{t+1}-\bar{x}_t\rangle + \langle \bar{w}_t,\bar{x}_{t+1}-\bar{x}_t\rangle + \frac{L}{2}\|\bar{x}_{t+1}-\bar{x}_t\|^2  \nonumber \\
  & \leq F(\bar{x}_t) + \frac{\eta_t\gamma}{\rho}\|\nabla F(\bar{x}_t)-\bar{w}_t\|^2 + \frac{\rho}{4\eta_t\gamma}\|\bar{x}_{t+1}-\bar{x}_t\|^2  -\frac{\rho}{\eta_t\gamma}\|\bar{x}_{t+1}-\bar{x}_t\|^2 + \frac{L}{2}\|\bar{x}_{t+1}-\bar{x}_t\|^2 \nonumber \\
  & = F(\bar{x}_t) + \frac{\eta_t\gamma}{\rho}\|\nabla F(\bar{x}_t)-\bar{w}_t\|^2 -\frac{\rho}{2\eta_t\gamma}\|\bar{x}_{t+1}-\bar{x}_t\|^2  -\big(\frac{\rho}{4\eta_t\gamma}-\frac{L}{2}\big)\|\bar{x}_{t+1}-\bar{x}_t\|^2 \nonumber \\
  & \leq F(\bar{x}_t) + \frac{\eta_t\gamma}{\rho}\|\nabla F(\bar{x}_t)-\bar{w}_t\|^2 -\frac{\rho}{2\eta_t\gamma}\|\bar{x}_{t+1}-\bar{x}_t\|^2 \nonumber \\
  & \leq F(\bar{x}_t) + \frac{1}{M}\sum_{m=1}^{M}\Big( \frac{2C_f^2\eta_t\gamma}{\rho}\|u^m_t - \nabla g^m(\bar{x}_t)\|^2 + \frac{4C_g^2\eta_t\gamma}{\rho}\|v^m_t - \nabla f^m(h^m_t)\|^2  \nonumber \\
 & \qquad + \frac{4C_g^2L^2_f\eta_t\gamma}{\rho}\|h^m_t - g^m(\bar{x}_t)\|^2  \Big) -\frac{\rho}{2\eta_t\gamma}\|\bar{x}_{t+1}-\bar{x}_t\|^2,
\end{align}
where the second last inequality is due to $0< \gamma \leq \frac{\rho}{2L\eta_t}$, and the last inequality holds by Lemma \ref{lem:A6}.

\end{proof}

\begin{lemma} \label{lem:A7}
Under the above assumptions, and assume the stochastic gradient estimators $\big\{h^m_t,u^m_t,v^m_t\big\}_{t=1}^T$ be generated from Algorithm \ref{alg:1},
we have, for any $m\in [M]$
 \begin{align} \label{eq:A48}
 \mathbb{E}\|h^m_{t+1} - g^m(x^m_{t+1})\|^2
 & \leq (1-\alpha_{t+1})\mathbb{E} \|h^m_t - g^m(x^m_t)\|^2 + 2\alpha_{t+1}^2\sigma^2 \nonumber \\
 & \quad + 2C_g^2\mathbb{E}\|x^m_{t+1}-x^m_t\|^2,
 \end{align}
 \begin{align} \label{eq:A49}
 \mathbb{E}\|u^m_{t+1} - \nabla g^m(x^m_{t+1})\|^2
 & \leq (1-\beta_{t+1})\mathbb{E}\|u^m_t - \nabla g^m(x^m_t)\|^2 + 2\beta_{t+1}^2\sigma^2 \nonumber \\
 & \quad + 2L_g^2\mathbb{E}\|x^m_{t+1}-x^m_t\|^2.
\end{align}
\begin{align} \label{eq:A50}
 \mathbb{E}\|v^m_{t+1} - \nabla f^m(h^m_{t+1})\|^2
 & \leq (1-\varrho_{t+1}) \mathbb{E}\|v^m_t -\nabla f^m(h^m_t)\|^2  + 4L^2_fC^2_g\mathbb{E}\|x^m_{t+1}-x^m_t\|^2 \nonumber \\
 &\quad + 2\varrho^2_{t+1}\sigma^2 + 8\alpha^2_{t+1}L^2_f\mathbb{E}\|h^m_t - g^m(x^m_t)\|^2+ 8L^2_f\alpha^2_{t+1}\sigma^2,
\end{align}
\end{lemma}

\begin{proof}
Without loss of generality, we only prove the above inequality \eqref{eq:A50}, and it is similar to the other inequalities.
Since $v^m_{t+1} = \Pi_{C_f}\Big[ \nabla f^m(h^m_{t+1};\xi^m_{t+1}) + (1-\varrho_{t+1})\big(v^m_t - \nabla f^m(h^m_t;\xi^m_{t+1})\big) \Big]$,
we have
\begin{align} \label{eq:A51}
 &\mathbb{E}\|v^m_{t+1} - \nabla f^m(h^m_{t+1})\|^2  \nonumber \\
 & = \mathbb{E}\big\|\Pi_{C_f}\big[ \nabla f^m(h^m_{t+1};\xi^m_{t+1}) + (1-\varrho_{t+1})\big(v^m_t - \nabla f^m(h^m_t;\xi^m_{t+1})\big) \big] - \Pi_{C_f}\big[ \nabla f^m(h^m_{t+1}) \big]\big\|^2 \nonumber \\
 & \leq \mathbb{E}\| \nabla f^m(h^m_{t+1};\xi^m_{t+1}) + (1-\varrho_{t+1})\big(v^m_t - \nabla f^m(h^m_t;\xi^m_{t+1})\big) - \nabla f^m(h^m_{t+1})\|^2 \nonumber \\
 & = \mathbb{E}\big\| (1-\varrho_{t+1})(v^m_t -\nabla f^m(h^m_t)) - \varrho_{t+1}(\nabla f^m(h^m_{t+1})-\nabla f^m(h^m_{t+1};\xi^m_{t+1})) \nonumber \\
 & \quad +(1-\varrho_{t+1})\big( \nabla f^m(h^m_{t+1};\xi^m_{t+1}) - \nabla f^m(h^m_{t};\xi^m_{t+1}) - \nabla f^m(h^m_{t+1}) + \nabla f^m(h^m_{t})\big) \big\|^2 \nonumber \\
 & =  (1-\varrho_{t+1})^2\mathbb{E}\|v^m_t -\nabla f^m(h^m_t)\|^2 + \mathbb{E}\big\|\varrho_{t+1}(\nabla f^m(h^m_{t+1})-\nabla f^m(h^m_{t+1};\xi^m_{t+1})) \nonumber \\
 & \quad -(1-\varrho_{t+1})\big( \nabla f^m(h^m_{t+1};\xi^m_{t+1}) - \nabla f^m(h^m_{t};\xi^m_{t+1}) - \nabla f^m(h^m_{t+1}) + \nabla f^m(h^m_{t})\big) \big\|^2 \nonumber \\
 & \leq (1-\varrho_{t+1})^2\mathbb{E}\|v^m_t -\nabla f^m(h^m_t)\|^2 + 2\varrho^2_{t+1}\mathbb{E}\big\|\nabla f^m(h^m_{t+1})-\nabla f^m(h^m_{t+1};\xi^m_{t+1})\big\|^2 \nonumber \\
 & \quad +2(1-\varrho_{t+1})^2\big\|\nabla f^m(h^m_{t+1};\xi^m_{t+1}) - \nabla f^m(h^m_{t};\xi^m_{t+1}) - \nabla f^m(h^m_{t+1}) + \nabla f^m(h^m_{t}) \big\|^2 \nonumber \\
 & \leq (1-\varrho_{t+1})^2\mathbb{E}\|v^m_t -\nabla f^m(h^m_t)\|^2 + 2\varrho^2_{t+1}\sigma^2 +2(1-\varrho_{t+1})^2\big\|\nabla f^m(h^m_{t+1};\xi^m_{t+1}) - \nabla f^m(h^m_{t};\xi^m_{t+1}) \big\|^2 \nonumber \\
 & \leq (1-\varrho_{t+1})^2\mathbb{E}\|v^m_t -\nabla f^m(h^m_t)\|^2 + 2\varrho^2_{t+1}\sigma^2 +2(1-\varrho_{t+1})^2L^2_f\mathbb{E}\|h^m_{t+1} - h^m_{t}\|^2,
\end{align}
where the third equality holds by the following fact:
\begin{align}
 &\mathbb{E}_{\xi^m_{t+1}}\big[\varrho_{t+1}(\nabla f^m(h^m_{t+1})-\nabla f^m(h^m_{t+1};\xi^m_{t+1})) - (1-\varrho_{t+1})\big( \nabla f^m(h^m_{t+1};\xi^m_{t+1}) - \nabla f^m(h^m_{t};\xi^m_{t+1}) \nonumber \\
 & \qquad \quad - \nabla f^m(h^m_{t+1}) + \nabla f^m(h^m_{t})\big)\big]=0, \nonumber
\end{align}
and
the second last inequality holds by the inequality $\mathbb{E}\|\zeta-\mathbb{E}[\zeta]\|^2 \leq \mathbb{E}\|\zeta\|^2$ and Assumption~\ref{ass:3};
the last inequality is due to Assumption~\ref{ass:1}.

Since $h^m_{t+1} = g^m(x^m_{t+1};\zeta^m_{t+1}) + (1-\alpha_{t+1})\big(h^m_t - g^m(x^m_t;\zeta^m_{t+1})\big)$,
we have
\begin{align} \label{eq:A52}
  \mathbb{E}\|h^m_{t+1} - h^m_{t}\|^2
 & = \mathbb{E}\|g^m(x^m_{t+1};\zeta^m_{t+1}) - g^m(x^m_t;\zeta^m_{t+1}) -\alpha_{t+1}\big(h^m_t - g^m(x^m_t;\zeta^m_{t+1}) \big)\|^2 \nonumber \\
 & \leq 2\mathbb{E}\|g^m(x^m_{t+1};\zeta^m_{t+1}) - g^m(x^m_t;\zeta^m_{t+1})\|^2 + 2\alpha^2_{t+1}\mathbb{E}\|h^m_t - g^m(x^m_t;\zeta^m_{t+1})\|^2 \nonumber \\
 & \leq 2C^2_g\|x^m_{t+1}-x^m_t\|^2 + 2\alpha^2_{t+1}\mathbb{E}\|h^m_t - g^m(x^m_t;\zeta^m_{t+1})\|^2  \nonumber \\
 & = 2C^2_g\|x^m_{t+1}-x^m_t\|^2 + 2\alpha^2_{t+1}\mathbb{E}\|h^m_t - g^m(x^m_t;\zeta^m_{t+1}) + g^m(x^m_t) - g^m(x^m_t)\|^2 \nonumber \\
 & \leq 2C^2_g\|x^m_{t+1}-x^m_t\|^2 + 4\alpha^2_{t+1}\mathbb{E}\|h^m_t - g^m(x^m_t)\|^2 +  4\alpha^2_{t+1}\mathbb{E}\|g^m(x^m_t;\zeta^m_{t+1}) + g^m(x^m_t)\|^2 \nonumber \\
 & \leq 2C^2_g\|x^m_{t+1}-x^m_t\|^2 + 4\alpha^2_{t+1}\mathbb{E}\|h^m_t - g^m(x^m_t)\|^2 +  4\alpha^2_{t+1}\sigma^2,
\end{align}
where the second inequality holds by Assumption , .

Combining the above inequalities \ref{eq:A51} with \ref{eq:A52}, we have
\begin{align}
 &\mathbb{E}\|v^m_{t+1} - \nabla f^m(h^m_{t+1})\|^2  \nonumber \\
 & \leq (1-\varrho_{t+1})^2\mathbb{E}\|v^m_t -\nabla f^m(h^m_t)\|^2 + 2\varrho^2_{t+1}\sigma^2 +2(1-\varrho_{t+1})^2L^2_f\mathbb{E}\|h^m_{t+1} - h^m_{t}\|^2 \nonumber \\
 & \leq (1-\varrho_{t+1})\mathbb{E}\|v^m_t -\nabla f^m(h^m_t)\|^2 + 2\varrho^2_{t+1}\sigma^2 +4L^2_fC^2_g\|x^m_{t+1}-x^m_t\|^2
  \nonumber \\
 & \quad + 8\alpha^2_{t+1}L^2_f\mathbb{E}\|h^m_t - g^m(x^m_t)\|^2+ 8L^2_f\alpha^2_{t+1}\sigma^2, \nonumber
\end{align}
where the last inequality holds by $0<\varrho_{t+1}\leq 1$.

\end{proof}

\begin{lemma} \label{lem:A10}
Based on the above Assumptions \ref{ass:1}-\ref{ass:2} and \ref{ass:6}, we have
\begin{align}
 & \sum_{m=1}^M\mathbb{E}\big\|\nabla f^m(h^m_t)-\frac{1}{M}\sum_{j=1}^M\nabla f^j(h^j_t)\big\|^2 \leq
 8L_f^2\sum_{m=1}^M\mathbb{E}\|h^m_t-g^m(\bar{x}_t)\|^2 + 4M\delta^2_f + 4ML^2_f\delta^2_g, \nonumber \\
 & \sum_{m=1}^M\mathbb{E}\big\|\nabla g^m(x^m_t)-\frac{1}{M}\sum_{j=1}^M\nabla g^j(x^j_t)\big\|^2 \leq 6L^2_g\sum_{m=1}^M \mathbb{E}\|x^m_t-\bar{x}_t\|^2 + 3M\delta_{g}^2 \nonumber \\
 & \sum_{m=1}^M\mathbb{E}\big\| g^m(x^m_t)-\frac{1}{M}\sum_{j=1}^M g^j(x^j_t)\big\|^2 \leq 6C^2_g\sum_{m=1}^M \mathbb{E}\|x^m_t-\bar{x}_t\|^2 + 3M\delta_{g}^2. \nonumber
\end{align}

\end{lemma}

\begin{proof}
Consider the term $\sum_{m=1}^M\mathbb{E}\big\|\nabla f^m(h^m_t)-\frac{1}{M}\sum_{j=1}^M\nabla f^j(h^j_t)\big\|^2$, we have
\begin{align}
 & \sum_{m=1}^M\mathbb{E}\big\|\nabla f^m(h^m_t)-\frac{1}{M}\sum_{j=1}^M\nabla f^j(h^j_t)\big\|^2   \nonumber \\
 & = \sum_{m=1}^M\mathbb{E}\big\|\nabla f^m(h^m_t)-\nabla f^m(g^m(\bar{x}_t)) + \nabla f^m(g^m(\bar{x}_t))
 - \frac{1}{M}\sum_{j=1}^M\nabla f^j(g^m(\bar{x}_t)) + \frac{1}{M}\sum_{j=1}^M\nabla f^j(g^m(\bar{x}_t)) \nonumber \\
 & \quad - \frac{1}{M}\sum_{j=1}^M\nabla f^j(g^j(\bar{x}_t)) + \frac{1}{M}\sum_{j=1}^M\nabla f^j(g^j(\bar{x}_t))  - \frac{1}{M}\sum_{j=1}^M\nabla f^j(h^j_t)\big\|^2   \nonumber \\
 & \leq \sum_{m=1}^M 4\mathbb{E}\big\|\nabla f^m(h^m_t)-\nabla f^m(g^m(\bar{x}_t))\big\|^2 + \sum_{m=1}^M 4\mathbb{E}\big\|\nabla f^m(g^m(\bar{x}_t))
 - \frac{1}{M}\sum_{j=1}^M\nabla f^j(g^m(\bar{x}_t))\big\|^2 \nonumber \\
 & \quad + \sum_{m=1}^M 4\mathbb{E}\big\|\frac{1}{M}\sum_{j=1}^M\nabla f^j(g^m(\bar{x}_t)) - \frac{1}{M}\sum_{j=1}^M\nabla f^j(g^j(\bar{x}_t)) \big\|^2  +  \sum_{m=1}^M 4\mathbb{E}\big\|\frac{1}{M}\sum_{j=1}^M\nabla f^j(g^j(\bar{x}_t))  - \frac{1}{M}\sum_{j=1}^M\nabla f^j(h^j_t) \big\|^2 \nonumber \\
 & \leq 4L_f^2\sum_{m=1}^M\mathbb{E}\|h^m_t-g^m(\bar{x}_t)\|^2  + 4\sum_{m=1}^M \frac{1}{M}\sum_{j=1}^M \mathbb{E}\|\nabla f^m(g^m(\bar{x}_t)) - \nabla f^j(g^m(\bar{x}_t))\|^2 \nonumber \\
 & \quad + 4L^2_f\sum_{m=1}^M\frac{1}{M}\sum_{j=1}^M\big\|g^m(\bar{x}_t)-g^j(\bar{x}_t)\big\|^2 +
 4L^2_f\sum_{j=1}^M\frac{1}{M}\sum_{m=1}^M\mathbb{E}\big\|g^m(\bar{x}_t)  - h^m_t\big\|^2\nonumber \\
 & \leq 8L_f^2\sum_{m=1}^M\mathbb{E}\|h^m_t-g^m(\bar{x}_t)\|^2 + 4M\delta^2_f + 4ML^2_f\delta^2_g,
\end{align}
where the last inequality holds by Assumption \ref{ass:6}.

Next, we have
\begin{align}
 & \sum_{m=1}^M\mathbb{E}\big\|\nabla g^m(x^m_t)-\frac{1}{M}\sum_{j=1}^M\nabla g^j(x^j_t)\big\|^2   \nonumber \\
 & = \sum_{m=1}^M\mathbb{E}\big\|\nabla g^m(x^m_t)- \nabla g^m(\bar{x}_t) + \nabla g^m(\bar{x}_t) - \frac{1}{M}\sum_{j=1}^M\nabla g^j(\bar{x}_t)
 + \frac{1}{M}\sum_{j=1}^M\nabla g^j(\bar{x}_t)-\frac{1}{M}\sum_{j=1}^M\nabla g^j(x^j_t)\big\|^2   \nonumber \\
 & \leq \sum_{m=1}^M 3\mathbb{E}\big\|\nabla g^m(x^m_t)- \nabla g^m(\bar{x}_t)\big\|^2 + \sum_{m=1}^M 3\mathbb{E}\big\|\nabla g^m(\bar{x}_t) - \frac{1}{M}\sum_{j=1}^M\nabla g^j(\bar{x}_t)\big\|^2 \nonumber \\
 & \quad + \sum_{m=1}^M 3\mathbb{E}\big\|\frac{1}{M}\sum_{j=1}^M\nabla g^j(\bar{x}_t)-\frac{1}{M}\sum_{j=1}^M\nabla g^j(x^j_t)\big\|^2   \nonumber \\
 & \leq 3L^2_g\sum_{m=1}^M \mathbb{E}\|x^m_t-\bar{x}_t\|^2 + 3\sum_{m=1}^M \frac{1}{M}\sum_{j=1}^M \mathbb{E}\|\nabla g^m(\bar{x}_t) - \nabla g^j(\bar{x}_t)\|^2 + 3\sum_{m=1}^M\frac{1}{M}\sum_{j=1}^M\big\|\nabla g^j(\bar{x}_t)-\nabla g^j(x^j_t)\big\|^2 \nonumber \\
 & \leq 6L^2_g\sum_{m=1}^M \mathbb{E}\|x^m_t-\bar{x}_t\|^2 + 3M\delta_{g}^2,
\end{align}
where the last inequality is due to the above Assumption \ref{ass:6}.

Similarly, we can obtain
\begin{align}
   \sum_{m=1}^M\mathbb{E}\big\| g^m(x^m_t)-\frac{1}{M}\sum_{j=1}^M g^j(x^j_t)\big\|^2 \leq 6C^2_g\sum_{m=1}^M \mathbb{E}\|x^m_t-\bar{x}_t\|^2 + 3M\delta_{g}^2.
\end{align}

\end{proof}

\begin{lemma} \label{lem:A11}
Suppose the iterates $\{x^m_t\}_{t=1}^T$, for all $m \in [M]$ generated from Algorithm \ref{alg:1} satisfy:
\begin{align}
& \sum_{m=1}^M \mathbb{E}\|x^m_t- \bar{x}_t \|^2 \leq (q -1)\sum_{l = s_t}^{t-1} \gamma^2\eta_l^2 \sum_{m = 1}^M \mathbb{E}\|d^m_l - \bar{d}_l\|^2,
\end{align}
where $\bar{x}_t=\frac{1}{M}\sum_{m=1}^Mx^m_t$, $d_t^m = \frac{x_t^m - x_{t+1}^m}{\gamma\eta_t}$ and $\bar{d}_t = \frac{\bar{x}_t - \bar{x}_{t+1}}{\eta_t\gamma}$.
\end{lemma}

\begin{proof}
According to the lines 7 and 11 of Algorithm \ref{alg:1}, we have
\begin{align}
 & x_{t+1}^m = x^m_t - \gamma\eta_tA^{-1}_tw^m_t = \arg\min_{x\in \mathbb{R}^d}\big\{ \langle w^m_t, x\rangle + \frac{1}{2\eta_t\gamma}(x-x^m_t)^TA_t(x-x^m_t) \big\}, \nonumber \\
 & \bar{x}_{t+1} = \bar{x}_t - \gamma\eta_tA^{-1}_t\bar{w}_t = \arg\min_{x\in \mathbb{R}^d}\big\{ \langle \bar{w}_t, x\rangle + \frac{1}{2\eta_t\gamma}(x-\bar{x}_t)^TA_t(x-\bar{x}_t) \big\}, \nonumber
\end{align}
and then we define the gradient mappings as in the above \eqref{eq:A2}: $d_t^m = \frac{x_t^m - x_{t+1}^m}{\gamma\eta_t}=A_t^{-1}w^m_t$ and $\bar{d}_t = \frac{\bar{x}_t - \bar{x}_{t+1}}{\eta_t\gamma}=A_t^{-1}\bar{w}_t=\frac{1}{M}\sum_{m=1}^Md_t^m$ for any
$m\in [M]$ and $t\geq 1$.

From the line 7 of Algorithm \ref{alg:1}, when $t = s_t = q\lfloor t/q \rfloor+1$, we have $x^m_t = \bar{x}_t = \frac{1}{M}
\sum_{m=1}^Mx^m_t$ for any $m\in [M]$, so the about inequality in the lemma holds trivially.

When $t\in (s_t,s_t+q)$, we have
\begin{align*}
    x^m_t = x^m_{s_t} - \sum_{l=s_t}^{t-1}\gamma \eta_l d^m_l, \quad \text{and} \quad \bar{x}_{t}  = \bar{x}_{s_t}  - \sum_{l=s_t}^{t-1}\gamma\eta_l \bar{d}_l.
\end{align*}
Thus we have
\begin{align*}
  \sum_{m=1}^M\mathbb{E}\|x^m_t - \bar{x}_t\|^2 & = \sum_{m =1}^M \mathbb{E}\Big\| x^m_{s_t} - \bar{x}_{s_t}
  - \Big(\sum_{l=s_t}^{t-1}\gamma \eta_l d^m_l - \sum_{l=s_t}^{t-1}\gamma\eta_l \bar{d}_l\Big) \Big\|^2 \\
  & = \sum_{m = 1}^M \mathbb{E}\Big\|\Big(\sum_{l=s_t}^{t-1}\gamma \eta_l d^m_l - \sum_{l=s_t}^{t-1}\gamma\eta_l \bar{d}_l\Big)\Big\|^2
  \leq (q -1)\sum_{l = s_t}^{t-1} \gamma^2\eta_l^2 \sum_{m = 1}^M \mathbb{E}\|d^m_l - \bar{d}_l\|^2,
\end{align*}
where the above inequality is due to $t-s_t\leq q-1$.
\end{proof}

\begin{lemma} \label{lem:A12}
Let $C^2_{fg}=\max(C^2_f,C^2_g)$, $L^2_{fg}=L^2_fC^2_g+L^2_g$ and $\eta_t \leq  \frac{\rho}{24\gamma qL_{fg}C_{fg}}$ for all $t\geq 0$. Further let $\alpha_{t+1}=c_1\eta^2_t$, $\beta_{t+1}=c_2\eta^2_t$ and
$\varrho_{t+1}=c_3\eta^2_t$, $c_1,c_2,c_3>0$ and $c^2_1 + c^2_2 \leq \frac{(24)^4q^2\gamma^4L^4_{fg}C^4_{fg}}{9\rho^4}$.
Set $s_t=q\lfloor t/q \rfloor+1$ and $t\in [s_t,s_t+q-1]$, we have
\begin{align} \label{eq:A54}
   & \sum_{t=s_t}^{s_t+q-1}\eta_t\sum_{m=1}^M\mathbb{E}\|d^m_t - \bar{d}_t\|^2 \nonumber \\
 & \leq \frac{6M}{5}\sum_{t=s_t}^{s_t+q-1}\eta_t\mathbb{E}\|\bar{d}_t\|^2 + \frac{\rho^2(c_1^2+c_3^2)}{120q^2\gamma^4C^2_{fg}L^2_{fg}C^2_g}\sum_{t=s_t}^{s_t+q-1}\eta_t\sum_{m=1}^M\mathbb{E}\|h^m_t-g^m(\bar{x}_t)\|^2
   + \frac{3M\hat{\delta}^2}{5\gamma^2 L^2_{fg}}\sum_{t=s_t}^{s_t+q-1}\eta^3_t,
\end{align}
where $\hat{\delta}^2 =  2c_1^2 L^2_f\sigma^2 + c_3^2\sigma^2 + 4c_3^2\delta^2_f + 4c_3^2 L^2_f\delta^2_g + c^2_2\sigma^2 + 3c^2_2\delta_{g}^2$.
\end{lemma}

\begin{proof}
According to the lines 7 and 11 of Algorithm \ref{alg:1}, we have
\begin{align}
 & x_{t+1}^m = \arg\min_{x\in \mathbb{R}^d}\Big\{ \langle w^m_t, x\rangle + \frac{1}{2\eta_t\gamma}(x-x^m_t)^TA_t(x-x^m_t) \Big\}, \nonumber \\
 & \bar{x}_{t+1} = \arg\min_{x\in \mathbb{R}^d}\Big\{ \langle \bar{w}_t, x\rangle + \frac{1}{2\eta_t\gamma}(x-\bar{x}_t)^TA_t(x-\bar{x}_t) \Big\}, \nonumber
\end{align}
and then we define the gradient mappings as in the above \eqref{eq:A2}: $d_t^m = \frac{x_t^m - x_{t+1}^m}{\gamma\eta_t}=A_t^{-1}w^m_t$ and $\bar{d}_t = \frac{\bar{x}_t - \bar{x}_{t+1}}{\eta_t\gamma}=A_t^{-1}\bar{w}_t=\frac{1}{M}\sum_{m=1}^Md_t^m$ for any
$m\in [M]$ and $t\geq 1$.
Then we have
\begin{align} \label{eq:A55}
   \sum_{m=1}^M\mathbb{E}\|d^m_t - \bar{d}_t\|^2 & =
  \sum_{m=1}^M\mathbb{E}\|A_t^{-1}(w^m_t - \bar{w}_t)\|^2 \leq \frac{1}{\rho^2}\sum_{m=1}^M\mathbb{E}\|w^m_t - \bar{w}_t\|^2  \\
  & = \frac{1}{\rho^2}\sum_{m=1}^M\mathbb{E}\|(u^m_t)^T - (u^m_t)^T\bar{v}_t + (u^m_t)^T\bar{v}_t - (\bar{u}_t)^T\bar{v}_t + (\bar{u}_t)^T\bar{v}_t - \frac{1}{M}\sum_{m=1}^M(u^m_t)^Tv^m_t\|^2 \nonumber \\
  & \leq \frac{1}{\rho^2}\sum_{m=1}^M\Big( 3C^2_g\mathbb{E}\|v^m_t - \bar{v}_t\|^2 + 3C^2_f\mathbb{E}\|u^m_t - \bar{u}_t\|^2 + 3\|(\bar{u}_t)^T\bar{v}_t - \frac{1}{M}\sum_{m=1}^M(u^m_t)^Tv^m_t\|^2 \Big), \nonumber
\end{align}
where the first inequality holds by Assumption \ref{ass:5}, i.e., $A_t\succeq \rho I_d$ for all $t\geq 1$,
and the last inequality holds by $\|u^m_t\|^2\leq C^2_g$ and $\|\bar{v}_t\|^2\leq C^2_f$.
Consider the term $\|(\bar{u}_t)^T\bar{v}_t - \frac{1}{M}\sum_{m=1}^M(u^m_t)^Tv^m_t\|^2$, we have
\begin{align} \label{eq:A56}
 \|(\bar{u}_t)^T\bar{v}_t - \frac{1}{M}\sum_{m=1}^M(u^m_t)^Tv^m_t\|^2 & = \|(\bar{u}_t)^T\bar{v}_t - \frac{1}{M}\sum_{m=1}^M(u^m_t)^Tv^m_t\|^2 \nonumber \\
 & \leq \frac{1}{M}\sum_{m=1}^M \|(\bar{u}_t)^T\bar{v}_t - (u^m_t)^Tv^m_t\|^2 \nonumber \\
 & =  \frac{1}{M}\sum_{m=1}^M \|(\bar{u}_t)^T\bar{v}_t - (\bar{u}_t)^Tv^m_t + (\bar{u}_t)^Tv^m_t - (u^m_t)^Tv^m_t\|^2 \nonumber \\
 & \leq \frac{1}{M}\sum_{m=1}^M \Big( 2C^2_g\|v^m_t - \bar{v}_t\|^2 + 2C^2_f\|u^m_t - \bar{u}_t\|^2 \Big).
\end{align}
By combining the above inequalities \eqref{eq:A55} and \eqref{eq:A56}, we have
\begin{align} \label{eq:A57}
  \sum_{m=1}^M\mathbb{E}\|d^m_t - \bar{d}_t\|^2 \leq \frac{9C^2_g}{\rho^2}\sum_{m=1}^M\mathbb{E}\|v^m_t - \bar{v}_t\|^2 + \frac{9C^2_f}{\rho^2}\sum_{m=1}^M\mathbb{E}\|u^m_t - \bar{u}_t\|^2.
\end{align}

Let $t=s_t=q\lfloor t/q\rfloor+1$. When $t=s_t$, we have $v^m_t=\bar{v}_t$ and $u^m_t=\bar{u}_t$ for any $m\in [M]$, so we have $\sum_{m=1}^M\mathbb{E}\|v^m_t - \bar{v}_t\|^2=0$ and $\sum_{m=1}^M\mathbb{E}\|u^m_t - \bar{u}_t\|^2=0$.
According to the above inequality \eqref{eq:A57}, when $t=s_t$, we have $\sum_{m=1}^M\mathbb{E}\|d^m_t - \bar{d}_t\|^2=0$. Clearly, the about inequality \eqref{eq:A54} in the lemma holds trivially.

When $t\in (s_t,s_t+q)$, we first consider the term $\sum_{m=1}^M\mathbb{E}\big\|v^m_t - \bar{v}_t\big\|^2$ as follows:
\begin{align} \label{eq:A58}
  & \sum_{m=1}^M\mathbb{E}\big\|v^m_t - \bar{v}_t\big\|^2  \\
  & = \sum_{m=1}^M\mathbb{E}\big\|v^m_t - \frac{1}{M}\sum_{m=1}^Mv^m_t\big\|^2 \nonumber \\
  & = \sum_{m=1}^M\mathbb{E}\big\|\Pi_{C_f}\big[ \nabla f^m(h^m_t;\xi^m_t) + (1-\varrho_t)\big(v^m_{t-1} - \nabla f^m(h^m_{t-1};\xi^m_t)\big) \big] - \frac{1}{M}\sum_{m=1}^M\Pi_{C_f}\big[ \nabla f^m(h^m_t;\xi^m_t) \nonumber \\
  & \qquad + (1-\varrho_t)\big(v^m_{t-1} - \nabla f^m(h^m_{t-1};\xi^m_t)\big) \big]\big\|^2 \nonumber \\
  & \leq \sum_{m=1}^M\mathbb{E}\big\|\nabla f^m(h^m_t;\xi^m_t) + (1-\varrho_t)\big(v^m_{t-1} - \nabla f^m(h^m_{t-1};\xi^m_t)\big) - \frac{1}{M}\sum_{m=1}^M \Big( \nabla f^m(h^m_t;\xi^m_t) \nonumber \\
  & \qquad + (1-\varrho_t)\big(v^m_{t-1} - \nabla f^m(h^m_{t-1};\xi^m_t)\big)\Big)\big\|^2 \nonumber \\
  & \leq (1+\nu)(1-\varrho_t)^2\sum_{m=1}^M\mathbb{E}\|v^m_{t-1} - \bar{v}_{t-1})\|^2 +
  (1+\frac{1}{\nu})\sum_{m=1}^M\mathbb{E}\big\|\nabla f^m(h^m_t;\xi^m_t)  \nonumber \\
  & \quad - \frac{1}{M}\sum_{m=1}^M\nabla f^m(h^m_t;\xi^m_t) - (1-\varrho_t)\big(\nabla f^m(h^m_{t-1};\xi^m_t)
  - \frac{1}{M}\sum_{m=1}^M \nabla f^m(h^m_{t-1};\xi^m_t) \big)\big\|^2. \nonumber
\end{align}
Then, we consider the last term of \eqref{eq:A58}:
\begin{align} \label{eq:A59}
  & \sum_{m=1}^M \mathbb{E}\big\|\nabla f^m(h^m_t;\xi^m_t)
  - \frac{1}{M}\sum_{m=1}^M\nabla f^m(h^m_t;\xi^m_t) - (1-\varrho_t)\big(\nabla f^m(h^m_{t-1};\xi^m_t)
  - \frac{1}{M}\sum_{m=1}^M \nabla f^m(h^m_{t-1};\xi^m_t) \big)\big\|^2 \nonumber \\
  & = \sum_{m=1}^M \mathbb{E}\big\| \nabla f^m(h^m_t;\xi^m_t) - \nabla f^m(h^m_{t-1};\xi^m_t)
  - \frac{1}{M}\sum_{m=1}^M\big( \nabla f^m(h^m_t;\xi^m_t) - \nabla f^m(h^m_{t-1};\xi^m_t) \big)  \nonumber \\
  & \quad +\varrho_t\big(\nabla f^m(h^m_{t-1};\xi^m_t) - \frac{1}{M}\sum_{m=1}^M \nabla f^m(h^m_{t-1};\xi^m_t) \big)\big\|^2 \nonumber \\
  & \leq 2\sum_{m=1}^M\mathbb{E}\big\|\nabla f^m(h^m_t;\xi^m_t) - \nabla f^m(h^m_{t-1};\xi^m_t)\|^2
   + 2\varrho^2_t\sum_{m=1}^M \mathbb{E}\big\|\nabla f^m(h^m_{t-1};\xi^m_t) - \frac{1}{M}\sum_{m=1}^M \nabla f^m(h^m_{t-1};\xi^m_t) \big\|^2
   \nonumber \\
  & \leq 2L^2_f\sum_{m=1}^M\mathbb{E}\|h^m_t-h^m_{t-1}\|^2 + 2\varrho^2_t\sum_{m=1}^M\mathbb{E}\big\|\nabla f^m(h^m_{t-1};\xi^m_t) - \frac{1}{M}\sum_{m=1}^M \nabla f^m(h^m_{t-1};\xi^m_t) \big\|^2,
\end{align}
where the second last inequality is due to Young inequality and the above Lemma \ref{lem:A2}.

Consider the term $\sum_{m=1}^M \big\|\nabla f^m(h^m_{t-1};\xi^m_t) - \frac{1}{M}\sum_{m=1}^M \nabla f^m(h^m_{t-1};\xi^m_t) \big\|^2$, we have
\begin{align} \label{eq:A60}
 & \sum_{m=1}^M \big\|\nabla f^m(h^m_{t-1};\xi^m_t) - \frac{1}{M}\sum_{m=1}^M \nabla f^m(h^m_{t-1};\xi^m_t) \big\|^2 \nonumber \\
 & = \sum_{m=1}^M \big\|\nabla f^m(h^m_{t-1};\xi^m_t) - \nabla f^m(h^m_{t-1}) - \frac{1}{M}\sum_{m=1}^M\big(\nabla f^m(h^m_{t-1};\xi^m_t) - \nabla f^m(h^m_{t-1}) \big) \nonumber \\
 & \quad \quad + \nabla f^m(h^m_{t-1}) - \frac{1}{M}\sum_{m=1}^M\nabla f^m(h^m_{t-1}) \big\|^2 \nonumber \\
 & \leq 2\sum_{m=1}^M\big\|\nabla f^m(h^m_{t-1};\xi^m_t) - \nabla f^m(h^m_{t-1}) - \frac{1}{M}\sum_{m=1}^M\big(\nabla f^m(h^m_{t-1};\xi^m_t) - \nabla f^m(h^m_{t-1}) \big) \big\| \nonumber \\
 & \quad \quad + 2\sum_{m=1}^M\big\|\nabla f^m(h^m_{t-1}) - \frac{1}{M}\sum_{m=1}^M\nabla f^m(h^m_{t-1}) \big\|^2 \nonumber \\
 & \leq 2\sum_{m=1}^M\big\|\nabla f^m(h^m_{t-1};\xi^m_t) - \nabla f^m(h^m_{t-1})\big\|^2 + 2\sum_{m=1}^M\big\|\nabla f^m(h^m_{t-1}) - \frac{1}{M}\sum_{m=1}^M\nabla f^m(h^m_{t-1}) \big\|^2 \nonumber \\
 & \leq 2M\sigma^2 + 16L_f^2\sum_{m=1}^M\mathbb{E}\|h^m_{t-1}-g^m(\bar{x}_{t-1})\|^2 + 8M\delta^2_f + 8ML^2_f\delta^2_g,
\end{align}
where the last inequality holds by the above Lemma \ref{lem:A10}.

Since $h^m_t = g^m(x^m_t;\zeta^m_t) + (1-\alpha_t)\big(h^m_{t-1} - g^m(x^m_{t-1};\zeta^m_t)\big)$,
we have
\begin{align} \label{eq:A61}
 & \mathbb{E}\|h^m_t - h^m_{t-1}\|^2 \nonumber \\
 & = \mathbb{E}\|g^m(x^m_t;\zeta^m_t) - g^m(x^m_{t-1};\zeta^m_t) -\alpha_t\big(h^m_{t-1} - g^m(x^m_{t-1};\zeta^m_t) \big)\|^2 \nonumber \\
 & \leq 2\mathbb{E}\|g^m(x^m_t;\zeta^m_t) - g^m(x^m_{t-1};\zeta^m_t)\|^2 + 2\alpha^2_t\mathbb{E}\|h^m_{t-1} - g^m(x^m_{t-1};\zeta^m_t)\|^2 \nonumber \\
 & \leq 2C^2_g\|x^m_t-x^m_{t-1}\|^2 + 2\alpha^2_t\mathbb{E}\|h^m_{t-1} - g^m(x^m_{t-1};\zeta^m_t)\|^2  \nonumber \\
 & = 2C^2_g\|x^m_t-x^m_{t-1}\|^2 + 2\alpha^2_t\mathbb{E}\|h^m_{t-1} - g^m(x^m_{t-1};\zeta^m_t) + g^m(x^m_{t-1}) - g^m(x^m_{t-1})\|^2 \nonumber \\
 & \leq 2C^2_g\|x^m_t-x^m_{t-1}\|^2 + 4\alpha^2_t\mathbb{E}\|h^m_{t-1} - g^m(x^m_{t-1})\|^2 +  4\alpha^2_t\mathbb{E}\|g^m(x^m_{t-1};\zeta^m_t) + g^m(x^m_{t-1})\|^2 \nonumber \\
 & \leq 2C^2_g\|x^m_t-x^m_{t-1}\|^2 + 4\alpha^2_t\mathbb{E}\|h^m_{t-1} - g^m(x^m_{t-1})\|^2 +  4\alpha^2_t\sigma^2 \nonumber \\
 & = 2C^2_g\|x^m_t-x^m_{t-1}\|^2 + 4\alpha^2_t\mathbb{E}\|h^m_{t-1} - g^m(\bar{x}_{t-1}) + g^m(\bar{x}_{t-1}) - g^m(x^m_{t-1})\|^2 +  4\alpha^2_t\sigma^2 \nonumber \\
 & \leq 2C^2_g\|x^m_t-x^m_{t-1}\|^2 + 8\alpha^2_t\mathbb{E}\|h^m_{t-1} - g^m(\bar{x}_{t-1})\|^2 + 8\alpha^2_tC^2_g\|\bar{x}_{t-1} - x^m_{t-1}\|^2 +  4\alpha^2_t\sigma^2,
\end{align}
where the second inequality holds by Assumption \ref{ass:3}.

By combining the above inequalities \eqref{eq:A58}, \eqref{eq:A59}, \eqref{eq:A60} and \eqref{eq:A61}, we have
\begin{align} \label{eq:A62}
 & \sum_{m=1}^M\mathbb{E}\|v^m_t - \bar{v}_t\|^2 \\
 & \leq (1+\nu)(1-\varrho_t)^2\sum_{m=1}^M\mathbb{E}\|v^m_{t-1} - \bar{v}_{t-1})\|^2 + (1+\frac{1}{\nu})\sum_{m=1}^M\mathbb{E}\big\|\nabla f^m(h^m_t;\xi^m_t)  \nonumber \\
 & \quad - \frac{1}{M}\sum_{m=1}^M\nabla f^m(h^m_t;\xi^m_t) - (1-\varrho_t)\big(\nabla f^m(h^m_{t-1};\xi^m_t)
  - \frac{1}{M}\sum_{m=1}^M \nabla f^m(h^m_{t-1};\xi^m_t) \big)\big\|^2 \nonumber \\
 & \leq (1+\nu)(1-\varrho_t)^2\sum_{m=1}^M\mathbb{E}\|v^m_{t-1} - \bar{v}_{t-1})\|^2 + (1+\frac{1}{\nu})
 \bigg(4L^2_fC^2_g\sum_{m=1}^M\mathbb{E}\|x^m_t-x^m_{t-1}\|^2 + 16\alpha^2_tL^2_f\sum_{m=1}^M\mathbb{E}\|h^m_{t-1} - g^m(\bar{x}_{t-1})\|^2   \nonumber \\
 & \quad + 16\alpha^2_tL^2_fC^2_g\sum_{m=1}^M\mathbb{E}\|\bar{x}_{t-1} - x^m_{t-1}\|^2 +  8ML^2_f\alpha^2_t\sigma^2 \nonumber \\
 & \quad + 4M\varrho^2_t\sigma^2 + 32\varrho^2_tL_f^2\sum_{m=1}^M\mathbb{E}\|h^m_{t-1}-g^m(\bar{x}_{t-1})\|^2 + 16M\varrho^2_t\delta^2_f + 16ML^2_f\varrho^2_t\delta^2_g \bigg)  \nonumber \\
 & \leq (1+\nu)(1-\varrho_t)^2\sum_{m=1}^M\mathbb{E}\|v^m_{t-1} - \bar{v}_{t-1})\|^2 + (1+\frac{1}{\nu})
 \bigg( 8L^2_fC^2_g\eta^2_{t-1}\gamma^2\sum_{m=1}^M\mathbb{E}\|d^m_{t-1} - \bar{d}_{t-1}\|^2 + 8L^2_fC^2_g\eta^2_{t-1}\gamma^2\sum_{m=1}^M\mathbb{E}\|\bar{d}_{t-1}\|^2    \nonumber \\
 & \quad + 16(q -1)L^2_fC^2_g\alpha^2_t\sum_{l = s_t}^{t-1} \gamma^2\eta_l^2 \sum_{m = 1}^M \mathbb{E}\|d^m_l - \bar{d}_l\|^2  +  8ML^2_f\alpha^2_t\sigma^2 \nonumber \\
 & \quad + 4M\varrho^2_t\sigma^2 + 32(\varrho^2_t+\alpha^2_t)L_f^2\sum_{m=1}^M\mathbb{E}\|h^m_{t-1}-g^m(\bar{x}_{t-1})\|^2 + 16M\varrho^2_t\delta^2_f + 16ML^2_f\varrho^2_t\delta^2_g \bigg) \nonumber \\
 & \leq (1+\nu)(1-\varrho_t)^2\sum_{m=1}^M\mathbb{E}\|v^m_{t-1} - \bar{v}_{t-1})\|^2 + (1+\frac{1}{\nu})
 \Bigg( \frac{72L^2_fC^4_g\eta^2_{t-1}\gamma^2}{\rho^2}\sum_{m=1}^M\mathbb{E}\|v^m_{t-1} - \bar{v}_{t-1}\|^2    \nonumber \\
 & \quad + \frac{72C^2_fL^2_fC^2_g\eta^2_{t-1}\gamma^2}{\rho^2}\sum_{m=1}^M\mathbb{E}\|u^m_{t-1} - \bar{u}_{t-1}\|^2 + 8L^2_fC^2_g\eta^2_{t-1}\gamma^2\sum_{m=1}^M\mathbb{E}\|\bar{d}_{t-1}\|^2 \nonumber \\
 & \quad + 16(q -1)L^2_fC^2_g\alpha^2_t\sum_{l = s_t}^{t-1} \gamma^2\eta_l^2 \Big( \frac{9C^2_g}{\rho^2}\sum_{m=1}^M\mathbb{E}\|v^m_l - \bar{v}_l\|^2 + \frac{9C^2_f}{\rho^2}\sum_{m=1}^M\mathbb{E}\|u^m_l - \bar{u}_l\|^2\Big)   \nonumber \\
 & \quad + 32(\varrho^2_t+\alpha^2_t)L_f^2\sum_{m=1}^M\mathbb{E}\|h^m_{t-1}-g^m(\bar{x}_{t-1})\|^2 + 8ML^2_f\alpha^2_t\sigma^2 + 4M\varrho^2_t\sigma^2 + 16M\varrho^2_t\delta^2_f + 16ML^2_f\varrho^2_t\delta^2_g \Bigg),
\end{align}
where the second last inequality holds by the above Lemma \ref{lem:A11} and the above inequality \eqref{eq:A57}, and the last inequality holds by
$d^m_{t-1} = \frac{x^m_t - x^m_{t-1}}{\eta_t\gamma}$, and the above inequality \eqref{eq:A57}.

Next, we consider the term $\sum_{m=1}^M\mathbb{E}\big\|u^m_t - \bar{u}_t\big\|^2$ as follows:
\begin{align} \label{eq:A63}
  & \sum_{m=1}^M\mathbb{E}\big\|u^m_t - \bar{u}_t\big\|^2  \\
  & = \sum_{m=1}^M\mathbb{E}\big\|u^m_t - \frac{1}{M}\sum_{m=1}^Mu^m_t\big\|^2 \nonumber \\
  & = \sum_{m=1}^M\mathbb{E}\big\|\Pi_{C_g}\big[ \nabla g^m(x^m_t;\zeta^m_t) + (1-\beta_t)\big(u^m_{t-1} - \nabla g^m(x^m_{t-1};\zeta^m_t)\big) \big] - \frac{1}{M}\sum_{m=1}^M\Pi_{C_g}\big[ \nabla g^m(x^m_t;\zeta^m_t) \nonumber \\
  & \qquad + (1-\beta_t)\big(u^m_{t-1} - \nabla g^m(x^m_{t-1};\zeta^m_t)\big) \big]\big\|^2 \nonumber \\
  & \leq \sum_{m=1}^M\mathbb{E}\big\|\nabla g^m(x^m_t;\zeta^m_t) + (1-\beta_t)\big(u^m_{t-1} - \nabla g^m(x^m_{t-1};\zeta^m_t)\big) - \frac{1}{M}\sum_{m=1}^M \Big( \nabla g^m(x^m_t;\zeta^m_t) \nonumber \\
  & \qquad + (1-\beta_t)\big(u^m_{t-1} - \nabla g^m(x^m_{t-1};\zeta^m_t)\big)\Big)\big\|^2 \nonumber \\
  & \leq (1+\nu)(1-\beta_t)^2\sum_{m=1}^M\mathbb{E}\|u^m_{t-1} - \bar{u}_{t-1})\|^2 +
  (1+\frac{1}{\nu})\sum_{m=1}^M\mathbb{E}\big\|\nabla g^m(x^m_t;\zeta^m_t)  \nonumber \\
  & \quad - \frac{1}{M}\sum_{m=1}^M\nabla g^m(x^m_t;\zeta^m_t) - (1-\beta_t)\big(\nabla g^m(x^m_{t-1};\zeta^m_t)
  - \frac{1}{M}\sum_{m=1}^M \nabla g^m(x^m_{t-1};\zeta^m_t) \big)\big\|^2.
\end{align}
Then, we consider the last term of \eqref{eq:A63}:
\begin{align} \label{eq:A64}
  & \sum_{m=1}^M \mathbb{E}\big\|\nabla g^m(x^m_t;\zeta^m_t)
  - \frac{1}{M}\sum_{m=1}^M\nabla g^m(x^m_t;\zeta^m_t) - (1-\beta_t)\big(\nabla g^m(x^m_{t-1};\zeta^m_t)
  - \frac{1}{M}\sum_{m=1}^M \nabla g^m(x^m_{t-1};\zeta^m_t) \big)\big\|^2 \nonumber \\
  & = \sum_{m=1}^M \mathbb{E}\big\| \nabla g^m(x^m_t;\zeta^m_t) - \nabla g^m(x^m_{t-1};\zeta^m_t)
  - \frac{1}{M}\sum_{m=1}^M\big( \nabla g^m(x^m_t;\zeta^m_t) - \nabla g^m(x^m_{t-1};\zeta^m_t) \big)  \nonumber \\
  & \quad +\beta_t\big(\nabla g^m(x^m_{t-1};\zeta^m_t) - \frac{1}{M}\sum_{m=1}^M \nabla g^m(x^m_{t-1};\zeta^m_t) \big)\big\|^2 \nonumber \\
  & \leq 2\sum_{m=1}^M\mathbb{E}\big\|\nabla g^m(x^m_t;\zeta^m_t) - \nabla g^m(x^m_{t-1};\zeta^m_t)\|^2
   + 2\beta^2_t\sum_{m=1}^M \mathbb{E}\big\|\nabla g^m(x^m_{t-1};\zeta^m_t) - \frac{1}{M}\sum_{m=1}^M \nabla g^m(x^m_{t-1};\zeta^m_t) \big\|^2
   \nonumber \\
  & \leq 2L^2_g\sum_{m=1}^M\mathbb{E}\|x^m_t-x^m_{t-1}\|^2 + 2\beta^2_t\sum_{m=1}^M\mathbb{E}\big\|\nabla g^m(x^m_{t-1};\zeta^m_t) - \frac{1}{M}\sum_{m=1}^M \nabla g^m(x^m_{t-1};\zeta^m_t) \big\|^2,
\end{align}
where the second last inequality is due to Young inequality and the above Lemma \ref{lem:A2}.

Consider the term $\sum_{m=1}^M \big\|\nabla g^m(x^m_{t-1};\zeta^m_t) - \frac{1}{M}\sum_{m=1}^M \nabla g^m(x^m_{t-1};\zeta^m_t) \big\|^2$, we have
\begin{align} \label{eq:A65}
 & \sum_{m=1}^M \big\|\nabla g^m(x^m_{t-1};\zeta^m_t) - \frac{1}{M}\sum_{m=1}^M \nabla g^m(x^m_{t-1};\zeta^m_t) \big\|^2 \nonumber \\
 & = \sum_{m=1}^M \big\|\nabla g^m(x^m_{t-1};\zeta^m_t) - \nabla g^m(x^m_{t-1}) - \frac{1}{M}\sum_{m=1}^M\big(\nabla g^m(x^m_{t-1};\zeta^m_t) - \nabla g^m(x^m_{t-1}) \big) \nonumber \\
 & \quad \quad + \nabla g^m(x^m_{t-1}) - \frac{1}{M}\sum_{m=1}^M\nabla g^m(x^m_{t-1}) \big\|^2 \nonumber \\
 & \leq 2\sum_{m=1}^M\big\|\nabla g^m(x^m_{t-1};\zeta^m_t) - \nabla g^m(x^m_{t-1}) - \frac{1}{M}\sum_{m=1}^M\big(\nabla g^m(x^m_{t-1};\zeta^m_t) - \nabla g^m(x^m_{t-1}) \big) \big\| \nonumber \\
 & \quad \quad + 2\sum_{m=1}^M\big\|\nabla g^m(x^m_{t-1}) - \frac{1}{M}\sum_{m=1}^M\nabla g^m(x^m_{t-1}) \big\|^2 \nonumber \\
 & \leq 2\sum_{m=1}^M\big\|\nabla g^m(x^m_{t-1};\zeta^m_t) - \nabla g^m(x^m_{t-1})\big\| + 2\sum_{m=1}^M\big\|\nabla g^m(x^m_{t-1}) - \frac{1}{M}\sum_{m=1}^M\nabla g^m(x^m_{t-1}) \big\|^2 \nonumber \\
 & \leq 2M\sigma^2 + 12L^2_g\sum_{m=1}^M \mathbb{E}\|x^m_{t-1}-\bar{x}_{t-1}\|^2 + 6M\delta_{g}^2,
\end{align}
where the last inequality holds by the above Lemma \ref{lem:A10}.

By combining the above inequalities \eqref{eq:A63}, \eqref{eq:A64} and \eqref{eq:A65}, we have
\begin{align}\label{eq:A66}
 & \sum_{m=1}^M\mathbb{E}\big\|u^m_t - \bar{u}_t\big\|^2  \\
  & \leq (1+\nu)(1-\beta_t)^2\sum_{m=1}^M\mathbb{E}\|u^m_{t-1} - \bar{u}_{t-1})\|^2 +
  (1+\frac{1}{\nu})\sum_{m=1}^M\mathbb{E}\big\|\nabla g^m(x^m_t;\zeta^m_t)  \nonumber \\
  & \quad - \frac{1}{M}\sum_{m=1}^M\nabla g^m(x^m_t;\zeta^m_t) - (1-\beta_t)\big(\nabla g^m(x^m_{t-1};\zeta^m_t)
  - \frac{1}{M}\sum_{m=1}^M \nabla g^m(x^m_{t-1};\zeta^m_t) \big)\big\|^2 \nonumber \\
  & \leq (1+\nu)(1-\beta_t)^2\sum_{m=1}^M\mathbb{E}\|u^m_{t-1} - \bar{u}_{t-1})\|^2 +
  (1+\frac{1}{\nu}) \Big( 2L^2_g\sum_{m=1}^M\mathbb{E}\|x^m_t-x^m_{t-1}\|^2 \nonumber \\
  & \quad + 4M\sigma^2\beta^2_t + 24L^2_g\beta^2_t\sum_{m=1}^M \mathbb{E}\|x^m_{t-1}-\bar{x}_{t-1}\|^2 + 12M\delta_{g}^2\beta^2_t \Big) \nonumber \\
  & \leq (1+\nu)(1-\beta_t)^2\sum_{m=1}^M\mathbb{E}\|u^m_{t-1} - \bar{u}_{t-1})\|^2 +
  (1+\frac{1}{\nu}) \Bigg( 4L^2_g\eta^2_{t-1}\gamma^2\sum_{m=1}^M\mathbb{E}\|d^m_{t-1}-\bar{d}_{t-1}\|^2 + 4L^2_g\eta^2_{t-1}\gamma^2\sum_{m=1}^M\mathbb{E}\|\bar{d}_{t-1}\|^2 \nonumber \\
  & \quad + 4M\sigma^2\beta^2_t + 24(q -1)L^2_g\beta^2_t\sum_{l = s_t}^{t-2} \gamma^2\eta_l^2 \sum_{m = 1}^M \mathbb{E}\|d^m_l - \bar{d}_l\|^2 + 12M\delta_{g}^2\beta^2_t \Bigg) \nonumber \\
  & \leq (1+\nu)(1-\beta_t)^2\sum_{m=1}^M\mathbb{E}\|u^m_{t-1} - \bar{u}_{t-1})\|^2 +
  (1+\frac{1}{\nu}) \Bigg(\frac{36C^2_gL^2_g\eta^2_{t-1}\gamma^2}{\rho^2}\sum_{m=1}^M\mathbb{E}\|v^m_{t-1} - \bar{v}_{t-1}\|^2 \nonumber \\
  & \quad  + \frac{36C^2_fL^2_g\eta^2_{t-1}\gamma^2}{\rho^2}\sum_{m=1}^M\mathbb{E}\|u^m_{t-1} - \bar{u}_{t-1}\|^2 + 4L^2_g\eta^2_{t-1}\gamma^2\sum_{m=1}^M\mathbb{E}\|\bar{d}_{t-1}\|^2 + 4M\sigma^2\beta^2_t + 12M\delta_{g}^2\beta^2_t \nonumber \\
  & \quad + 24(q -1)L^2_g\beta^2_t\sum_{l = s_t}^{t-2} \gamma^2\eta_l^2 \Big( \frac{9C^2_g}{\rho^2}\sum_{m=1}^M\mathbb{E}\|v^m_l - \bar{v}_l\|^2 + \frac{9C^2_f}{\rho^2}\sum_{m=1}^M\mathbb{E}\|u^m_l - \bar{u}_l\|^2\Big) \Bigg),
\end{align}
where the last inequality holds by the above inequality \eqref{eq:A57}.

By summing the above inequalities \eqref{eq:A62} and \eqref{eq:A66}, we have
\begin{align}\label{eq:A67}
 & \sum_{m=1}^M\big(\mathbb{E}\big\|u^m_t - \bar{u}_t \big\|^2 + \mathbb{E}\big\|v^m_t - \bar{v}_t\big\|^2 \big) \\
   & \leq (1+\nu)(1-\beta_t)^2\sum_{m=1}^M\mathbb{E}\|u^m_{t-1} - \bar{u}_{t-1})\|^2 +
  (1+\frac{1}{\nu}) \Bigg(\frac{36C^2_gL^2_g\eta^2_{t-1}\gamma^2}{\rho^2}\sum_{m=1}^M\mathbb{E}\|v^m_{t-1} - \bar{v}_{t-1}\|^2 \nonumber \\
  & \quad  + \frac{36C^2_fL^2_g\eta^2_{t-1}\gamma^2}{\rho^2}\sum_{m=1}^M\mathbb{E}\|u^m_{t-1} - \bar{u}_{t-1}\|^2 + 4L^2_g\eta^2_{t-1}\gamma^2\sum_{m=1}^M\mathbb{E}\|\bar{d}_{t-1}\|^2 + 4M\sigma^2\beta^2_t + 12M\delta_{g}^2\beta^2_t \nonumber \\
  & \quad + 24(q -1)L^2_g\beta^2_t\sum_{l = s_t}^{t-2} \gamma^2\eta_l^2 \Big( \frac{9C^2_g}{\rho^2}\sum_{m=1}^M\mathbb{E}\|v^m_l - \bar{v}_l\|^2 + \frac{9C^2_f}{\rho^2}\sum_{m=1}^M\mathbb{E}\|u^m_l - \bar{u}_l\|^2\Big) \Bigg) \nonumber \\
  & \quad + (1+\nu)(1-\varrho_t)^2\sum_{m=1}^M\mathbb{E}\|v^m_{t-1} - \bar{v}_{t-1})\|^2 + (1+\frac{1}{\nu})
 \Bigg( \frac{72L^2_fC^4_g\eta^2_{t-1}\gamma^2}{\rho^2}\sum_{m=1}^M\mathbb{E}\|v^m_{t-1} - \bar{v}_{t-1}\|^2    \nonumber \\
 & \quad + \frac{72C^2_fL^2_fC^2_g\eta^2_{t-1}\gamma^2}{\rho^2}\sum_{m=1}^M\mathbb{E}\|u^m_{t-1} - \bar{u}_{t-1}\|^2 + 8L^2_fC^2_g\eta^2_{t-1}\gamma^2\sum_{m=1}^M\mathbb{E}\|\bar{d}_{t-1}\|^2 \nonumber \\
 & \quad + 16(q -1)L^2_fC^2_g\alpha^2_t\sum_{l = s_t}^{t-1} \gamma^2\eta_l^2 \Big( \frac{9C^2_g}{\rho^2}\sum_{m=1}^M\mathbb{E}\|v^m_l - \bar{v}_l\|^2 + \frac{9C^2_f}{\rho^2}\sum_{m=1}^M\mathbb{E}\|u^m_l - \bar{u}_l\|^2\Big)   \nonumber \\
 & \quad + 32(\varrho^2_t+\alpha^2_t)L_f^2\sum_{m=1}^M\mathbb{E}\|h^m_{t-1}-g^m(\bar{x}_{t-1})\|^2 + 8ML^2_f\alpha^2_t\sigma^2 + 4M\varrho^2_t\sigma^2 + 16M\varrho^2_t\delta^2_f + 16ML^2_f\varrho^2_t\delta^2_g \Bigg) \nonumber \\
 & \leq \max\Big((1+\nu)(1-\beta_t)^2 + (1+\frac{1}{\nu})\frac{72C^2_f(L^2_fC^2_g+L^2_g)\eta^2_{t-1}\gamma^2}{\rho^2}, (1+\nu)(1-\varrho_t)^2 + (1+\frac{1}{\nu})\frac{72C^2_g(C^2_g L^2_f+L^2_g)\eta^2_{t-1}\gamma^2}{\rho^2} \Big) \nonumber \\
 & \quad \cdot \sum_{m=1}^M\big(\mathbb{E}\big\|u^m_{t-1} - \bar{u}_{t-1} \big\|^2 + \mathbb{E}\big\|v^m_{t-1} - \bar{v}_{t-1}\big\|^2 \big)
 + 8(1+\frac{1}{\nu})(L^2_fC^2_g+L^2_g)\eta^2_{t-1}\gamma^2\sum_{m=1}^M\mathbb{E}\|\bar{d}_{t-1}\|^2 \nonumber \\
 & \quad + 24(1+\frac{1}{\nu})(q -1)\big(L^2_fC^2_g\alpha^2_t + L^2_g\beta^2_t\big) \sum_{l = s_t}^{t-1} \gamma^2\eta_l^2 \Big( \frac{9C^2_g}{\rho^2}\sum_{m=1}^M\mathbb{E}\|v^m_l - \bar{v}_l\|^2 + \frac{9C^2_f}{\rho^2}\sum_{m=1}^M\mathbb{E}\|u^m_l - \bar{u}_l\|^2\Big) \nonumber \\
 & \quad + 32(1+\frac{1}{\nu})(\varrho^2_t+\alpha^2_t)L_f^2\sum_{m=1}^M\mathbb{E}\|h^m_{t-1}-g^m(\bar{x}_{t-1})\|^2 + (1+\frac{1}{\nu})\Big( 8ML^2_f\alpha^2_t\sigma^2 + 4M\varrho^2_t\sigma^2 + 16M\varrho^2_t\delta^2_f \nonumber \\
 & \quad + 16ML^2_f\varrho^2_t\delta^2_g + 4M\sigma^2\beta^2_t + 12M\delta_{g}^2\beta^2_t \Big).
\end{align}

Let $C^2_{fg}=\max(C^2_f,C^2_g)$, $L^2_{fg}=L^2_fC^2_g+L^2_g$, $\nu=\frac{1}{q}$ and $\eta_t \leq  \frac{\rho}{24\gamma qL_{fg}C_{fg}}$ for all $t\geq 0$.
Since $\beta_t \in (0,1)$ for all $t\geq 0$, we have
\begin{align}
 & (1+\nu)(1-\beta_t)^2 + (1+\frac{1}{\nu})\frac{72C^2_f(L^2_fC^2_g+L^2_g)\eta^2_{t-1}\gamma^2}{\rho^2} \nonumber \\
 & \leq 1+ \frac{1}{q} + (1+q)\frac{72C^2_f(L^2_fC^2_g+L^2_g)\gamma^2}{\rho^2}\frac{\rho^2}{576\gamma^2 q^2L^2_{fg}C^2_{fg}} \nonumber \\
 & \leq 1+ \frac{1}{q} + \frac{1+q}{8q^2} \leq 1+ \frac{5}{4q}.
\end{align}
Similarly, since $\varrho_t\in (0,1)$ for all $t\geq 0$, we have $(1+\nu)(1-\varrho_t)^2 + (1+\frac{1}{\nu})\frac{72C^2_g(C^2_g L^2_f+L^2_g)\eta^2_{t-1}\gamma^2}{\rho^2}\leq 1+ \frac{5}{4q}$.
Based on the above inequality \eqref{eq:A67} and the parameters, then we have
\begin{align} \label{eq:A68}
 &  \sum_{m=1}^M\big(\mathbb{E}\big\|u^m_t - \bar{u}_t \big\|^2 + \mathbb{E}\big\|v^m_t - \bar{v}_t\big\|^2 \big) \\
 & \leq \big(1+ \frac{5}{4q} \big)\sum_{m=1}^M\big(\mathbb{E}\big\|u^m_{t-1} - \bar{u}_{t-1} \big\|^2 + \mathbb{E}\big\|v^m_{t-1} - \bar{v}_{t-1}\big\|^2 \big) + 8(q+1)L^2_{fg}\eta^2_{t-1}\gamma^2\sum_{m=1}^M\mathbb{E}\|\bar{d}_{t-1}\|^2 \nonumber \\
 & \quad + 216(q^2 -1)\frac{C^2_{fg}L^2_{fg}\gamma^2}{\rho^2}\big(\alpha^2_t +\beta^2_t\big) \sum_{l = s_t}^{t-1}\eta_l^2 \sum_{m=1}^M\Big(\mathbb{E}\|v^m_l - \bar{v}_l\|^2 + \mathbb{E}\|u^m_l - \bar{u}_l\|^2\Big) \nonumber \\
 & \quad + 32(1+q)(\varrho^2_t+\alpha^2_t)L_f^2\sum_{m=1}^M\mathbb{E}\|h^m_{t-1}-g^m(\bar{x}_{t-1})\|^2 \nonumber \\
 & \quad + 4M(q+1)\Big( 2L^2_f\alpha^2_t\sigma^2 + \varrho^2_t\sigma^2 + 4\varrho^2_t\delta^2_f + 4L^2_f\varrho^2_t\delta^2_g + \sigma^2\beta^2_t + 3\delta_{g}^2\beta^2_t \Big) \nonumber \\
 & \leq \big(1+ \frac{5}{4q} \big)\sum_{m=1}^M\big(\mathbb{E}\big\|u^m_{t-1} - \bar{u}_{t-1} \big\|^2 + \mathbb{E}\big\|v^m_{t-1} - \bar{v}_{t-1}\big\|^2 \big) + \frac{\rho^2}{36qC^2_{fg}}\sum_{m=1}^M\mathbb{E}\|\bar{d}_{t-1}\|^2 \nonumber \\
 & \quad + \frac{3(c^2_1+c^2_2)}{8}\eta^2_{t-1}\sum_{l = s_t}^{t-2}\eta_l^2 \sum_{m=1}^M\Big(\mathbb{E}\|v^m_l - \bar{v}_l\|^2 + \mathbb{E}\|u^m_l - \bar{u}_l\|^2\Big) + \frac{\rho^2(c_1^2+c_3^2)}{9q\gamma^2C^2_{fg}C^2_g}\eta^2_{t-1}\sum_{m=1}^M\mathbb{E}\|h^m_{t-1}-g^m(\bar{x}_{t-1})\|^2\nonumber \\
 & \quad + \frac{M\rho}{3\gamma L_{fg}C_{fg}}\Big( 2c_1^2 L^2_f\sigma^2 + c_3^2\sigma^2 + 4c_3^2\delta^2_f + 4c_3^2 L^2_f\delta^2_g + c^2_2\sigma^2 + 3c^2_2\delta_{g}^2\Big)\eta^3_{t-1},
\end{align}
where the first inequality holds by the above inequality \eqref{eq:A63} and $\nu = \frac{1}{q}$, and
the last inequality holds by $\alpha_t = c_1\eta^2_{t-1}$, $\beta_t = c_2\eta^2_{t-1}$, $\varrho_t = c_3\eta^2_{t-1}$ and
$\eta_t \leq  \frac{\rho}{24\gamma qL_{fg}C_{fg}}$ for all $t\geq 0$, and $\frac{L^2_f}{L^2_{fg}}\leq \frac{1}{C^2_g}$.

According to the above inequality \eqref{eq:A68}, we have
\begin{align} \label{eq:A69}
 &  \sum_{m=1}^M\big(\mathbb{E}\big\|u^m_t - \bar{u}_t \big\|^2 + \mathbb{E}\big\|v^m_t - \bar{v}_t\big\|^2 \big) \nonumber \\
 & \leq  \frac{\rho^2}{36qC^2_{fg}}\sum_{s=s_t}^{t-1}\big(1+ \frac{5}{4q} \big)^{t-1-s}\sum_{m=1}^M\mathbb{E}\|\bar{d}_s\|^2 \nonumber \\
 & \quad + \frac{3(c^2_1+c^2_2)}{8}\sum_{s=s_t}^{t-1}\big(1+ \frac{5}{4q} \big)^{t-1-s}\eta^2_s\sum_{l = s_t}^{s-2}\eta_l^2 \sum_{m=1}^M\Big(\mathbb{E}\|v^m_l - \bar{v}_l\|^2 + \mathbb{E}\|u^m_l - \bar{u}_l\|^2\Big) \nonumber \\
 & \quad + \frac{\rho^2(c_1^2+c_3^2)}{9q\gamma^2C^2_{fg}C^2_g}\sum_{s=s_t}^{t-1}\big(1+ \frac{5}{4q} \big)^{t-1-s}\eta_s^2\sum_{m=1}^M\mathbb{E}\|h^m_{s-1}-g^m(\bar{x}_{s-1})\|^2\nonumber \\
 & \quad + \frac{M\rho}{3\gamma L_{fg}C_{fg}}\Big( 2c_1^2 L^2_f\sigma^2 + c_3^2\sigma^2 + 4c_3^2\delta^2_f + 4c_3^2 L^2_f\delta^2_g + c^2_2\sigma^2 + 3c^2_2\delta_{g}^2\Big)\sum_{s=s_t}^{t-1}\big(1+ \frac{5}{4q} \big)^{t-1-s}\eta^3_s \nonumber \\
 & \leq   \frac{\rho^2}{9qC^2_{fg}}\sum_{s=s_t}^{t-1}\sum_{m=1}^M\mathbb{E}\|\bar{d}_s\|^2
  + \frac{3(c^2_1+c^2_2)}{2}\sum_{s=s_t}^{t-1}\eta^2_s\sum_{l = s_t}^{s-2}\eta_l^2 \sum_{m=1}^M\Big(\mathbb{E}\|v^m_l - \bar{v}_l\|^2 + \mathbb{E}\|u^m_l - \bar{u}_l\|^2\Big) \nonumber \\
 & \quad + \frac{4\rho^2(c_1^2+c_3^2)}{9q\gamma^2C^2_{fg}C^2_g}\sum_{s=s_t}^{t-1}\eta_s^2\sum_{m=1}^M\mathbb{E}\|h^m_{s-1}-g^m(\bar{x}_{s-1})\|^2\nonumber \\
 & \quad + \frac{4M\rho}{3\gamma L_{fg}C_{fg}}\Big( 2c_1^2 L^2_f\sigma^2 + c_3^2\sigma^2 + 4c_3^2\delta^2_f + 4c_3^2 L^2_f\delta^2_g + c^2_2\sigma^2 + 3c^2_2\delta_{g}^2\Big)\sum_{s=s_t}^{t-1}\eta^3_s \nonumber \\
 & \leq \frac{M\rho^2}{9qC^2_{fg}}\sum_{s=s_t}^{t-1}\mathbb{E}\|\bar{d}_s\|^2
  + \frac{\rho^2(c^2_1+c^2_2)}{24*16\gamma^2qL^2_{fg}C^2_{fg}}\sum_{s=s_t}^{t-1}\eta^2_s \sum_{m=1}^M\Big(\mathbb{E}\|v^m_s - \bar{v}_s\|^2 + \mathbb{E}\|u^m_s - \bar{u}_s\|^2\Big) \nonumber \\
 & \quad + \frac{4\rho^2(c_1^2+c_3^2)}{9q\gamma^2C^2_{fg}C^2_g}\sum_{s=s_t}^{t-1}\eta_s^2\sum_{m=1}^M\mathbb{E}\|h^m_{s-1}-g^m(\bar{x}_{s-1})\|^2\nonumber \\
 & \quad + \frac{4M\rho}{3\gamma L_{fg}C_{fg}}\Big( 2c_1^2 L^2_f\sigma^2 + c_3^2\sigma^2 + 4c_3^2\delta^2_f + 4c_3^2 L^2_f\delta^2_g + c^2_2\sigma^2 + 3c^2_2\delta_{g}^2\Big)\sum_{s=s_t}^{t-1}\eta^3_s,
\end{align}
where the second inequality holds by $\big(1+ \frac{5}{4q} \big)^{t-1-s}\leq \big(1+ \frac{5}{4q} \big)^{q} \leq e^{5/4}\leq 4$
and the last inequality holds by $\eta_t \leq  \frac{\rho}{24\gamma qL_{fg}C_{fg}}$ for all $t\geq 0$.

By multiplying both sides of \eqref{eq:A69} by $\eta_t$ and summing over $t=s_t$ to $s_t+q-1$, we have
\begin{align} \label{eq:A70}
 & \sum_{t=s_t}^{s_t+q-1}\eta_t\sum_{m=1}^M\big(\mathbb{E}\big\|u^m_t - \bar{u}_t \big\|^2 + \mathbb{E}\big\|v^m_t - \bar{v}_t\big\|^2 \big) \nonumber \\
 & \leq \frac{M\rho^2}{9C^2_{fg}}\sum_{t=s_t}^{s_t+q-1}\eta_t\mathbb{E}\|\bar{d}_t\|^2 + \frac{\rho^4(c^2_1+c^2_2)}{24^3*16\gamma^4q^2L^4_{fg}C^4_{fg}}\sum_{t=s_t}^{s_t+q-1}\eta_t\sum_{m=1}^M\Big(\mathbb{E}\|v^m_t - \bar{v}_t\|^2 + \mathbb{E}\|u^m_t - \bar{u}_t\|^2\Big) \nonumber \\
 & \quad + \frac{\rho^4(c_1^2+c_3^2)}{24*54q^2\gamma^4C^4_{fg}L^2_{fg}C^2_g}\sum_{t=s_t}^{s_t+q-1}\eta_t\sum_{m=1}^M\mathbb{E}\|h^m_t-g^m(\bar{x}_t)\|^2
  \nonumber \\
 & \quad + \frac{M\rho^2}{18\gamma^2 L^2_{fg}C^2_{fg}}\Big( 2c_1^2 L^2_f\sigma^2 + c_3^2\sigma^2 + 4c_3^2\delta^2_f + 4c_3^2 L^2_f\delta^2_g + c^2_2\sigma^2 + 3c^2_2\delta_{g}^2 \Big)\sum_{t=s_t}^{s_t+q-1}\eta^3_t,
\end{align}

Given $c^2_1 + c^2_2 \leq \frac{(24)^4q^2\gamma^4L^4_{fg}C^4_{fg}}{9\rho^4}$, we have
$\frac{60}{72}\leq 1-\frac{\rho^4(c^2_1+c^2_2)}{24^3*16\gamma^4q^2L^4_{fg}C^4_{fg}}$,
we have
\begin{align}
 & \sum_{t=s_t}^{s_t+q-1}\eta_t\sum_{m=1}^M\big(\mathbb{E}\big\|u^m_t - \bar{u}_t \big\|^2 + \mathbb{E}\big\|v^m_t - \bar{v}_t\big\|^2 \big) \nonumber \\
 & \leq  \frac{2M\rho^2}{15C^2_{fg}}\sum_{t=s_t}^{s_t+q-1}\eta_t\mathbb{E}\|\bar{d}_t\|^2 + \frac{\rho^4(c_1^2+c_3^2)}{1080q^2\gamma^4C^4_{fg}L^2_{fg}C^2_g}\sum_{t=s_t}^{s_t+q-1}\eta_t\sum_{m=1}^M\mathbb{E}\|h^m_t-g^m(\bar{x}_t)\|^2
  \nonumber \\
 & \quad + \frac{M\rho^2}{15\gamma^2 L^2_{fg}C^2_{fg}}\Big( 2c_1^2 L^2_f\sigma^2 + c_3^2\sigma^2 + 4c_3^2\delta^2_f + 4c_3^2 L^2_f\delta^2_g + c^2_2\sigma^2 + 3c^2_2\delta_{g}^2 \Big)\sum_{t=s_t}^{s_t+q-1}\eta^3_t.
\end{align}

According to the above inequality \eqref{eq:A57} and $C^2_{fg}=\max(C^2_f,C^2_g)$, we have
\begin{align}
  \sum_{m=1}^M\mathbb{E}\|d^m_t - \bar{d}_t\|^2 \leq \frac{9C^2_g}{\rho^2}\sum_{m=1}^M\mathbb{E}\|v^m_t - \bar{v}_t\|^2 + \frac{9C^2_f}{\rho^2}\sum_{m=1}^M\mathbb{E}\|u^m_t - \bar{u}_t\|^2 \leq \frac{9C^2_{fg}}{\rho^2}\sum_{m=1}^M\big(\mathbb{E}\big\|u^m_t - \bar{u}_t \big\|^2 + \mathbb{E}\big\|v^m_t - \bar{v}_t\big\|^2 \big).
\end{align}
Thus we have
\begin{align}
 & \sum_{t=s_t}^{s_t+q-1}\eta_t\sum_{m=1}^M\mathbb{E}\|d^m_t - \bar{d}_t\|^2 \nonumber \\
 & \leq \frac{9C^2_{fg}}{\rho^2}\sum_{t=s_t}^{s_t+q-1}\eta_t\sum_{m=1}^M\big(\mathbb{E}\big\|u^m_t - \bar{u}_t \big\|^2 + \mathbb{E}\big\|v^m_t - \bar{v}_t\big\|^2 \big) \nonumber \\
 & \leq \frac{6M}{5}\sum_{t=s_t}^{s_t+q-1}\eta_t\mathbb{E}\|\bar{d}_t\|^2 + \frac{\rho^2(c_1^2+c_3^2)}{120q^2\gamma^4C^2_{fg}L^2_{fg}C^2_g}\sum_{t=s_t}^{s_t+q-1}\eta_t\sum_{m=1}^M\mathbb{E}\|h^m_t-g^m(\bar{x}_t)\|^2
  \nonumber \\
 & \quad + \frac{3M}{5\gamma^2 L^2_{fg}}\Big( 2c_1^2 L^2_f\sigma^2 + c_3^2\sigma^2 + 4c_3^2\delta^2_f + 4c_3^2 L^2_f\delta^2_g + c^2_2\sigma^2 + 3c^2_2\delta_{g}^2 \Big)\sum_{t=s_t}^{s_t+q-1}\eta^3_t.
\end{align}

\end{proof}

\begin{theorem}  \label{th:A1}
(Restatement of Theorem 1)
Assume the sequence $\{\bar{x}_t\}_{t=1}^T$ be generated from \textbf{AdaMFCGD} algorithm.
 Under the above Assumptions, and let $\eta_t=\frac{k}{(n+t)^{1/3}}$ for all $t\geq 0$, $\alpha_{t+1}=c_1\eta_t^2$, $\beta_{t+1}=c_2\eta_t^2$, $\varrho_{t+1}=c_3\eta_t^2$, $n \geq \max\big(2, k^3, (c_1k)^3, (c_2k)^3, (c_3k)^3, \frac{(24k\gamma qL_{fg}C_{fg})^3}{\rho^3}\big)$, $k>0$,
 $c_1 \geq \frac{2}{3k^3} + B$, $c_2 \geq \frac{2}{3k^3} + 5C^2_f$, $c^2_1 + c^2_2 \leq \frac{(24)^4q^2\gamma^4L^4_{fg}C^4_{fg}}{9\rho^4}$, $c_3 \geq \frac{2}{3k^3} + 5C^2_g$, $ \frac{\rho(c^2_1+c^2_3)^{1/4}}{12\sqrt{5q} L_{fg}C_{fg}} \leq \gamma \leq \min\Big(\frac{3\rho qL_{fg}C_{fg}}{4(C^2_g+L^2_g+ 2L^2_fC^2_g)}, \frac{n^{1/3}\rho}{2Lk} \Big)$, $B\geq 20C_g^2L^2_f + \frac{c^2_2C^2_gL^2_f}{216q^3\gamma^3L^3_{fg}C^3_{fg}} + \frac{\Theta\rho^2(c_1^2+c_3^2)}{30q^2\gamma^4C^2_{fg}L^2_{fg}C^2_g}$, $\Theta = \Big(5C^2_fL^2_g + \frac{c^2_2C^2_gL^2_f}{864q^3\gamma^3L^3_{fg}C^3_{fg}} \Big)\frac{\rho^2}{(24)^2L^2_{fg}C^2_{fg}}+\frac{\gamma\rho}{6qL_{fg}C_{fg}}\Big(C_g^2 + L_g^2 + 2L^2_fC^2_g\Big)$ and $\Theta+\frac{BC^2_g\rho^2}{(24)^2L^2_{fg}C^2_{fg}}\leq \frac{5\rho^2}{48}$, we have
\begin{align}
\frac{1}{T}\sum_{t=1}^T\mathbb{E}\|\nabla F(\bar{x}_t)\|  \leq \Big( \frac{\sqrt{2G}n^{1/6}}{T^{1/2}} + \frac{\sqrt{2G}}{T^{1/3}}\Big)\sqrt{\frac{1}{T}\sum_{t=1}^T\mathbb{E}\|A_t\|^2},
\end{align}
where $C^2_{fg}=\max(C^2_f,C^2_g)$, $L^2_{fg}=L^2_fC^2_g+L^2_g$, $G = \frac{4(F(\bar{x}_1) - F^*)}{k\rho\gamma} + \frac{12n^{1/3}\sigma^2}{qk^2\rho^2} + 4k^2\Big(\frac{\hat{\delta}^2}{4\gamma^2 L^2_{fg}} + \frac{\big( c^2_1 + c^2_2 + c^2_3 \big)\sigma^2}{3\rho\gamma qL_{fg}C_{fg}}\Big)\ln(n+T)$  and $\hat{\delta}^2 =  2c_1^2 L^2_f\sigma^2 + c_3^2\sigma^2 + 4c_3^2\delta^2_f + 4c_3^2 L^2_f\delta^2_g + c^2_2\sigma^2 + 3c^2_2\delta_{g}^2$.
\end{theorem}

\begin{proof}
Since $\eta_t=\frac{k}{(n+t)^{1/3}}$ on $t$ is decreasing and $n\geq k^3$, we have $\eta_t \leq \eta_0 = \frac{k}{n^{1/3}} \leq 1$ and $\gamma \leq \frac{n^{1/3}\rho}{2Lk}\leq \frac{\rho}{2L\eta_0} \leq \frac{\rho}{2L\eta_t}$ for any $t\geq 0$. Since $\eta_t \leq \frac{\rho}{24\gamma qL_{fg}C_{fg}}$ for all $t\geq 0$, we have
$\frac{k}{n^{1/3}} =\eta_0\leq \eta_t \leq \frac{\rho}{24\gamma qL_{fg}C_{fg}}$, then we have $n \geq \frac{(24k\gamma qL_{fg}C_{fg})^3}{\rho^3}$.
 Due to $0 < \eta_t \leq 1$ and $n\geq (c_1k)^3$, we have $\alpha_{t+1} = c_1\eta_t^2 \leq c_1\eta_t \leq \frac{c_1k}{n^{1/3}}\leq 1$.
Similarly, due to $n\geq (c_2k)^3$ and $n\geq (c_3k)^3$, we have $\beta_{t+1}\leq 1$ and $\varrho_{t+1}\leq 1$.

 According to Lemma \ref{lem:A7}, for any $m\in [M]$, we have
 \begin{align}
  & \frac{1}{\eta_t}\mathbb{E}\|h^m_{t+1} - g^m(x^m_{t+1})\|^2 - \frac{1}{\eta_{t-1}}\mathbb{E}\|h^m_{t} - g^m(x^m_{t})\|^2  \\
  & \leq \big(\frac{1-\alpha_{t+1}}{\eta_t} - \frac{1}{\eta_{t-1}}\big)\mathbb{E}\|h^m_{t} - g^m(x^m_{t})\|^2 + 2C_g^2\mathbb{E}\|x^m_{t+1}-x^m_t\|^2 + 2\alpha_{t+1}^2\sigma^2 \nonumber \\
  & = \big(\frac{1}{\eta_t} - \frac{1}{\eta_{t-1}} - c_1\eta_t\big)\mathbb{E}\|h^m_{t} - g^m(x^m_{t})\|^2 + 2C_g^2\mathbb{E}\|x^m_{t+1}-x^m_t\|^2 + 2\alpha_{t+1}^2\sigma^2, \nonumber
 \end{align}
 where the second equality is due to $\alpha_{t+1}=c_1\eta^2_t$.
 Similarly, since $\beta_{t+1}=c_2\eta^2_t$, we have
 \begin{align}
 & \frac{1}{\eta_t}\mathbb{E}\|u^m_{t+1} - \nabla g^m(x^m_{t+1})\|^2 - \frac{1}{\eta_{t-1}}\mathbb{E}\|u^m_t - \nabla g^m(x^m_t)\|^2 \\
 & \leq \big(\frac{1-\beta_{t+1}}{\eta_t} -\frac{1}{\eta_{t-1}}\big) \mathbb{E}\|u^m_t - \nabla g^m(x^m_t)\|^2 + 2L_g^2\mathbb{E}\|x^m_{t+1}-x^m_t\|^2 + 2\beta_{t+1}^2\sigma^2 \nonumber \\
 & =  \big(\frac{1}{\eta_t} -\frac{1}{\eta_{t-1}} -c_2\eta_t\big) \mathbb{E}\|u^m_t - \nabla g^m(x^m_t)\|^2  + 2L_g^2\mathbb{E}\|x^m_{t+1}-x^m_t\|^2 + 2\beta_{t+1}^2\sigma^2. \nonumber
 \end{align}
 And we have
  \begin{align}
  & \frac{1}{\eta_t}\mathbb{E}\|v^m_{t+1} - \nabla f^m(h^m_{t+1})\|^2 - \frac{1}{\eta_{t-1}}\mathbb{E}\|v^m_{t} - \nabla f^m(h^m_{t})\|^2 \\
  & \leq \big(\frac{1-\varrho_{t+1}}{\eta_t} - \frac{1}{\eta_{t-1}}\big)\mathbb{E}\|v^m_{t} - \nabla f^m(h^m_{t})\|^2 + 4L^2_fC^2_g\mathbb{E}\|x^m_{t+1}-x^m_t\|^2 + 2\varrho^2_{t+1}\sigma^2 \nonumber \\
  & \quad + 8\alpha^2_{t+1}L^2_f\mathbb{E}\|h^m_t - g^m(x^m_t)\|^2+ 8L^2_f\alpha^2_{t+1}\sigma^2  \nonumber \\
  & = \big(\frac{1}{\eta_t} - \frac{1}{\eta_{t-1}} - c_3\eta_t\big)\mathbb{E}\|v^m_{t} - \nabla f^m(h^m_{t})\|^2 + 4L^2_fC^2_g\mathbb{E}\|x^m_{t+1}-x^m_t\|^2 + 2\varrho^2_{t+1}\sigma^2 \nonumber \\
  & \quad + 8\alpha^2_{t+1}L^2_f\mathbb{E}\|h^m_t - g^m(x^m_t)\|^2+ 8L^2_f\alpha^2_{t+1}\sigma^2, \nonumber
 \end{align}
 where the second equality is due to $\varrho_{t+1}=c_2\eta^2_t$.

By $\eta_t = \frac{k}{(n+t)^{1/3}}$, we have
 \begin{align}
  \frac{1}{\eta_t} - \frac{1}{\eta_{t-1}} &= \frac{1}{k}\big( (n+t)^{\frac{1}{3}} - (n+t-1)^{\frac{1}{3}}\big) \leq \frac{1}{3k(n+t-1)^{2/3}} \leq \frac{1}{3k\big(n/2+t\big)^{2/3}} \nonumber \\
  & \leq \frac{2^{2/3}}{3k(n+t)^{2/3}} = \frac{2^{2/3}}{3k^3}\frac{k^2}{(n+t)^{2/3}} = \frac{2^{2/3}}{3k^3}\eta_t^2 \leq \frac{2}{3k^3}\eta_t,
 \end{align}
 where the first inequality holds by the concavity of function $f(x)=x^{1/3}$, \emph{i.e.}, $(x+y)^{1/3}\leq x^{1/3} + \frac{y}{3x^{2/3}}$; the second inequality is due to $n\geq 2$,  and
 the last inequality is due to $0<\eta_t\leq 1$.

Let $c_1 \geq \frac{2}{3k^3} + B$, for any $m\in [M]$, we have
 \begin{align} \label{eq:W1}
  & \frac{1}{\eta_t}\mathbb{E}\|h^m_{t+1} - g^m(x^m_{t+1})\|^2 - \frac{1}{\eta_{t-1}}\mathbb{E}\|h^m_{t} - g^m(x^m_{t})\|^2   \\
  & \leq -B\eta_t \mathbb{E}\|h^m_{t} - g^m(x^m_{t})\|^2 + 2C_g^2\mathbb{E}\|x^m_{t+1}-x^m_t\|^2 + 2\alpha_{t+1}^2\sigma^2 \nonumber \\
  & = -B\eta_t \mathbb{E}\|h^m_{t} - g^m(x^m_{t})\|^2 + 2C_g^2\eta^2_t\gamma^2\mathbb{E}\|d^m_t-\bar{d}_t+\bar{d}_t\|^2 + 2\alpha_{t+1}^2\sigma^2 \nonumber \\
  & \leq -B\eta_t \mathbb{E}\|h^m_{t} - g^m(x^m_{t})\|^2 + 4C_g^2\eta^2_t\gamma^2\mathbb{E}\|d^m_t-\bar{d}_t\|^2 + 4C_g^2\eta^2_t\gamma^2\mathbb{E}\|\bar{d}_t\|^2 + 2\alpha_{t+1}^2\sigma^2 \nonumber \\
  & \leq -\frac{B}{2}\eta_t\|h^m_t - g^m(\bar{x}_t)\|^2 + B C^2_g\eta_t\|x^m_t-\bar{x}_t\|^2+ 4C_g^2\eta^2_t\gamma^2\mathbb{E}\|d^m_t-\bar{d}_t\|^2 + 4C_g^2\eta^2_t\gamma^2\mathbb{E}\|\bar{d}_t\|^2 + 2\alpha_{t+1}^2\sigma^2 , \nonumber
 \end{align}
 where the last inequality holds by $-\|h^m_t - g^m(x^m_t)\|^2 \leq -\frac{1}{2}\|h^m_t - g^m(\bar{x}_t)\|^2 + \|g^m(x^m_t) -g^m(\bar{x}_t)\|^2 \leq -\frac{1}{2}\|h^m_t - g^m(\bar{x}_t)\|^2 + C^2_g\|x^m_t-\bar{x}_t\|^2$.

Let $c_2 \geq \frac{2}{3k^3} + 5C^2_f$, for any $m\in [M]$, we have
 \begin{align} \label{eq:W2}
  & \frac{1}{\eta_t}\mathbb{E}\|u^m_{t+1} - \nabla g^m(x^m_{t+1})\|^2 - \frac{1}{\eta_{t-1}}\mathbb{E}\|u^m_t - \nabla g^m(x^m_t)\|^2  \\
  & \leq -5C^2_f\eta_t \mathbb{E}\|u^m_t - \nabla g^m(x^m_t)\|^2 +
   2L_g^2\mathbb{E}\|x^m_{t+1}-x^m_t\|^2 + 2\beta_{t+1}^2\sigma^2 \nonumber \\
  & = -5C^2_f\eta_t \mathbb{E}\|u^m_t - \nabla g^m(x^m_t)\|^2 +
   2L_g^2\eta^2_t\gamma^2\mathbb{E}\|d^m_t - \bar{d}_t + \bar{d}_t\|^2 + 2\beta_{t+1}^2\sigma^2 \nonumber \\
  & \leq -5C^2_f\eta_t \mathbb{E}\|u^m_t - \nabla g^m(x^m_t)\|^2 +
   4L_g^2\eta^2_t\gamma^2\mathbb{E}\|d^m_t - \bar{d}_t\|^2 + 4L_g^2\eta^2_t\gamma^2\mathbb{E}\|\bar{d}_t\|^2 + 2\beta_{t+1}^2\sigma^2 \nonumber \\
  & \leq -\frac{5C^2_f\eta_t}{2} \mathbb{E}\|u^m_t - \nabla g^m(\bar{x}_t)\|^2 + 5C^2_fL^2_g\eta_t\|x^m_t-\bar{x}_t\|^2 +
   4L_g^2\eta^2_t\gamma^2\mathbb{E}\|d^m_t - \bar{d}_t\|^2 + 4L_g^2\eta^2_t\gamma^2\mathbb{E}\|\bar{d}_t\|^2 + 2\beta_{t+1}^2\sigma^2, \nonumber
 \end{align}
 where the last inequality holds by $-\|u^m_t - \nabla g^m(x^m_t)\|^2 \leq -\frac{1}{2}\|u^m_t - \nabla g^m(\bar{x}_t)\|^2 + \|\nabla g(x^m_t) - \nabla g^m(\bar{x}_t)\|^2 \leq -\frac{1}{2}\|u^m_t - \nabla g^m(\bar{x}_t)\|^2 + L^2_g\|x^m_t-\bar{x}_t\|^2$.

Let $c_3 \geq \frac{2}{3k^3} + 5C^2_g$, for any $m\in [M]$, we have
 \begin{align} \label{eq:W3}
  & \frac{1}{\eta_t}\mathbb{E}\|v^m_{t+1} - \nabla f^m(h^m_{t+1})\|^2 - \frac{1}{\eta_{t-1}}\mathbb{E}\|v^m_{t} - \nabla f^m(h^m_{t})\|^2   \\
  & \leq -5C^2_g\eta_t \mathbb{E}\|v^m_{t} - \nabla f^m(h^m_{t})\|^2 + 4L^2_fC^2_g\mathbb{E}\|x^m_{t+1}-x^m_t\|^2 + 2\varrho^2_{t+1}\sigma^2 + 8\alpha^2_{t+1}L^2_f\mathbb{E}\|h^m_t - g^m(x^m_t)\|^2+ 8L^2_f\alpha^2_{t+1}\sigma^2 \nonumber \\
  & = -5C^2_g\eta_t \mathbb{E}\|v^m_{t} - \nabla f^m(h^m_{t})\|^2 + 4L^2_fC^2_g\eta^2_t\gamma^2\mathbb{E}\|d^m_t - \bar{d}_t + \bar{d}_t\|^2 + 2\varrho^2_{t+1}\sigma^2 + 8\alpha^2_{t+1}L^2_f\mathbb{E}\|h^m_t - g^m(x^m_t)\|^2+ 8L^2_f\alpha^2_{t+1}\sigma^2 \nonumber \\
  & \leq -5C^2_g\eta_t \mathbb{E}\|v^m_{t} - \nabla f^m(h^m_{t})\|^2 + 8L^2_fC^2_g\eta^2_t\gamma^2\mathbb{E}\|d^m_t - \bar{d}_t\|^2 + 8L^2_fC^2_g\eta^2_t\gamma^2\mathbb{E}\|\bar{d}_t\|^2 + 2\varrho^2_{t+1}\sigma^2 \nonumber \\
  & \quad + 8\alpha^2_{t+1}L^2_f\mathbb{E}\|h^m_t - g^m(x^m_t)\|^2+ 8L^2_f\alpha^2_{t+1}\sigma^2 \nonumber \\
  & \leq -5C^2_g\eta_t \mathbb{E}\|v^m_{t} - \nabla f^m(h^m_{t})\|^2 + 8L^2_fC^2_g\eta^2_t\gamma^2\mathbb{E}\|d^m_t - \bar{d}_t\|^2 + 8L^2_fC^2_g\eta^2_t\gamma^2\mathbb{E}\|\bar{d}_t\|^2 + 2\varrho^2_{t+1}\sigma^2 \nonumber \\
  & \quad + \frac{c^2_2L^2_f}{864q^3\gamma^3L^3_{fg}C^3_{fg}}\eta_t\mathbb{E}\|h^m_t - g^m(\bar{x}_t)\|^2 + \frac{c^2_2C^2_gL^2_f}{864q^3\gamma^3L^3_{fg}C^3_{fg}}\eta_t\mathbb{E}\|x^m_t - \bar{x}_t\|^2 + 8L^2_f\alpha^2_{t+1}\sigma^2, \nonumber
 \end{align}
where the last inequality holds by Assumption , $\alpha_{t+1}=c_2\eta^2_t$ and $\eta_t \leq \frac{\rho}{24q\gamma L_{fg}C_{fg}}$ for all $t\geq 0$.

According to Lemma \ref{lem:A5}, we have
\begin{align} \label{eq:W4}
 F(\bar{x}_{t+1})-F(\bar{x}_t) & \leq  \frac{1}{M}\sum_{m=1}^{M}\Big( \frac{2C_f^2\eta_t\gamma}{\rho}\|u^m_t - \nabla g^m(\bar{x}_t)\|^2 + \frac{4C_g^2\eta_t\gamma}{\rho}\|v^m_t - \nabla f^m(h^m_t)\|^2  \nonumber \\
 & \qquad + \frac{4C_g^2L^2_f\eta_t\gamma}{\rho}\|h^m_t - g^m(\bar{x}_t)\|^2  \Big) -\frac{\rho}{2\eta_t\gamma}\|\bar{x}_{t+1}-\bar{x}_t\|^2 \nonumber \\
 & = \frac{1}{M}\sum_{m=1}^{M}\Big( \frac{2C_f^2\eta_t\gamma}{\rho}\|u^m_t - \nabla g^m(\bar{x}_t)\|^2 + \frac{4C_g^2\eta_t\gamma}{\rho}\|v^m_t - \nabla f^m(h^m_t)\|^2  \nonumber \\
 & \qquad + \frac{4C_g^2L^2_f\eta_t\gamma}{\rho}\|h^m_t - g^m(\bar{x}_t)\|^2  \Big) -\frac{\rho\eta_t\gamma}{2}\|\bar{d}_t\|^2.
\end{align}

\begin{align}
   & \sum_{t=s_t}^{s_t+q-1}\eta_t\sum_{m=1}^M\mathbb{E}\|d^m_t - \bar{d}_t\|^2 \nonumber \\
 & \leq \frac{6M}{5}\sum_{t=s_t}^{s_t+q-1}\eta_t\mathbb{E}\|\bar{d}_t\|^2 + \frac{\rho^2(c_1^2+c_3^2)}{120q^2\gamma^4C^2_{fg}L^2_{fg}C^2_g}\sum_{t=s_t}^{s_t+q-1}\eta_t\sum_{m=1}^M\mathbb{E}\|h^m_t-g^m(\bar{x}_t)\|^2
   + \frac{3M\hat{\delta}^2}{5\gamma^2 L^2_{fg}}\sum_{t=s_t}^{s_t+q-1}\eta^3_t,
\end{align}

Next, we define a \emph{potential} function, for any $t\geq 1$
\begin{align}
 \Omega_t & = \mathbb{E}\Big [F(\bar{x}_t) + \frac{\gamma}{\rho\eta_{t-1}} \frac{1}{M}\sum_{m=1}^M \Big( \|h^m_{t} - g^m(x^m_{t})\|^2
  + \|u^m_t - \nabla g^m(x^m_t)\|^2 + \|v^m_{t} - \nabla f^m(h^m_{t})\|^2 \Big)\Big]. \nonumber
\end{align}
Then we have
 \begin{align} \label{eq:W5}
 & \Omega_{t+1} - \Omega_t \nonumber \\
 & = F(\bar{x}_{t+1}) - F(\bar{x}_t)
 + \frac{\gamma}{M\rho}\sum_{m=1}^M \Bigg( \frac{1}{\eta_t}\mathbb{E}\|h^m_{t+1} - g^m(x^m_{t+1})\|^2 - \frac{1}{\eta_{t-1}}\mathbb{E}\|h^m_{t} - g^m(x^m_{t})\|^2 + \frac{1}{\eta_t}\|u^m_{t+1} - \nabla g^m(x^m_{t+1})\|^2\nonumber \\
 & \quad -\frac{1}{\eta_{t-1}}\|u^m_t - \nabla g^m(x^m_t)\|^2 + \frac{1}{\eta_t}\|v^m_{t+1} - \nabla f^m(h^m_{t+1})\|^2
 - \frac{1}{\eta_{t-1}}\|v^m_{t} - \nabla f^m(h^m_{t})\|^2 \Bigg) \nonumber \\
 & \leq \frac{1}{M}\sum_{m=1}^{M}\Big( \frac{2C_f^2\eta_t\gamma}{\rho}\|u^m_t - \nabla g^m(\bar{x}_t)\|^2 + \frac{4C_g^2\eta_t\gamma}{\rho}\|v^m_t - \nabla f^m(h^m_t)\|^2 + \frac{4C_g^2L^2_f\eta_t\gamma}{\rho}\|h^m_t - g^m(\bar{x}_t)\|^2  \Big) -\frac{\rho\eta_t\gamma}{2}\|\bar{d}_t\|^2 \nonumber \\
 & \quad + \frac{\gamma}{M\rho}\sum_{m=1}^M \bigg( -\frac{B}{2}\eta_t\|h^m_t - g^m(\bar{x}_t)\|^2 + B C^2_g\eta_t\|x^m_t-\bar{x}_t\|^2+ 4C_g^2\eta^2_t\gamma^2\mathbb{E}\|d^m_t-\bar{d}_t\|^2 + 4C_g^2\eta^2_t\gamma^2\mathbb{E}\|\bar{d}_t\|^2 + 2\alpha_{t+1}^2\sigma^2  \nonumber \\
 & \quad -\frac{5C^2_f\eta_t}{2} \mathbb{E}\|u^m_t - \nabla g^m(\bar{x}_t)\|^2 + 5C^2_fL^2_g\eta_t\|x^m_t-\bar{x}_t\|^2 +
   4L_g^2\eta^2_t\gamma^2\mathbb{E}\|d^m_t - \bar{d}_t\|^2 + 4L_g^2\eta^2_t\gamma^2\mathbb{E}\|\bar{d}_t\|^2 + 2\beta_{t+1}^2\sigma^2 \nonumber \\
 & \quad -5C^2_g\eta_t \mathbb{E}\|v^m_{t} - \nabla f^m(h^m_{t})\|^2 + 8L^2_fC^2_g\eta^2_t\gamma^2\mathbb{E}\|d^m_t - \bar{d}_t\|^2 + 8L^2_fC^2_g\eta^2_t\gamma^2\mathbb{E}\|\bar{d}_t\|^2 + 2\varrho^2_{t+1}\sigma^2 \nonumber \\
 & \quad + \frac{c^2_2L^2_f}{864q^3\gamma^3L^3_{fg}C^3_{fg}}\eta_t\mathbb{E}\|h^m_t - g^m(\bar{x}_t)\|^2 + \frac{c^2_2C^2_gL^2_f}{864q^3\gamma^3L^3_{fg}C^3_{fg}}\eta_t\mathbb{E}\|x^m_t - \bar{x}_t\|^2 + 8L^2_f\alpha^2_{t+1}\sigma^2 \bigg) \nonumber \\
 & \leq \frac{1}{M}\sum_{m=1}^{M}\Bigg( -\frac{C_f^2\gamma}{2\rho}\eta_t\|u^m_t - \nabla g^m(\bar{x}_t)\|^2 - \frac{C_g^2\gamma}{\rho}\eta_t\|v^m_t - \nabla f^m(h^m_t)\|^2 \nonumber \\
 & \quad - \frac{\gamma}{\rho}\big(\frac{B}{2} - 4C_g^2L^2_f -  \frac{c^2_2C^2_gL^2_f}{864q^3\gamma^3L^3_{fg}C^3_{fg}}\big)\eta_t\|h^m_t - g^m(\bar{x}_t)\|^2  \Bigg) -\Big(\frac{\rho\gamma\eta_t}{2} - \frac{4C_g^2\eta^2_t\gamma^3}{\rho} - \frac{4L_g^2\eta^2_t\gamma^3}{\rho} - \frac{8L^2_fC^2_g\eta^2_t\gamma^3}{\rho} \Big)\|\bar{d}_t\|^2 \nonumber \\
 & \quad + \frac{\gamma}{M\rho}\Big( BC^2_g + 5C^2_fL^2_g + \frac{c^2_2C^2_gL^2_f}{864q^3\gamma^3L^3_{fg}C^3_{fg}} \Big)\eta_t(q -1)\sum_{l = s_t}^{t-1} \gamma^2\eta_l^2 \sum_{m = 1}^M \mathbb{E}\|d^m_l - \bar{d}_l\|^2 \nonumber \\
 & \quad + \frac{\gamma}{M\rho}\Big(4C_g^2\eta^2_t\gamma^2 + 4L_g^2\eta^2_t\gamma^2 + 8L^2_fC^2_g\eta^2_t\gamma^2\Big)\sum_{m = 1}^M\mathbb{E}\|d^m_t - \bar{d}_t\|^2 + \frac{2\sigma^2\gamma}{\rho}\big( \alpha^2_{t+1} + \beta^2_{t+1} + \varrho^2_{t+1}\big),
 \end{align}
where the first inequality holds by the above inequalities \eqref{eq:W1}, \eqref{eq:W2}, \eqref{eq:W3} and \eqref{eq:W4},
and the last inequality is due to Lemma \ref{lem:A11}.

Let $s_t=q\lfloor t/q\rfloor+1$, summing the above inequality \eqref{eq:W5} over $t=s_t$ to $s_t+q-1$,
we have
 \begin{align} \label{eq:W6}
 & \sum_{t=s_t}^{s_t+q-1}\big( \Omega_{t+1} - \Omega_t \big) \nonumber \\
 & \leq \sum_{t=s_t}^{s_t+q-1}\frac{1}{M}\sum_{m=1}^{M}\Bigg( -\frac{C_f^2\gamma}{2\rho}\eta_t\|u^m_t - \nabla g^m(\bar{x}_t)\|^2 - \frac{C_g^2\gamma}{\rho}\eta_t\|v^m_t - \nabla f^m(h^m_t)\|^2 \nonumber \\
 & \quad - \frac{\gamma}{\rho}\big(\frac{B}{2} - 4C_g^2L^2_f -  \frac{c^2_2C^2_gL^2_f}{864q^3\gamma^3L^3_{fg}C^3_{fg}}\big)\eta_t\|h^m_t - g^m(\bar{x}_t)\|^2  \Bigg) \nonumber \\
 & \quad -\sum_{t=s_t}^{s_t+q-1}\Big(\frac{\rho\gamma\eta_t}{2} - \frac{4C_g^2\eta^2_t\gamma^3}{\rho} - \frac{4L_g^2\eta^2_t\gamma^3}{\rho} - \frac{8L^2_fC^2_g\eta^2_t\gamma^3}{\rho} \Big)\|\bar{d}_t\|^2 \nonumber \\
 & \quad + \frac{\gamma}{M\rho}\Big( BC^2_g + 5C^2_fL^2_g + \frac{c^2_2C^2_gL^2_f}{864q^3\gamma^3L^3_{fg}C^3_{fg}} \Big)\sum_{t=s_t}^{s_t+q-1}\eta_t(q -1)\sum_{l = s_t}^{t-1} \gamma^2\eta_l^2 \sum_{m = 1}^M \mathbb{E}\|d^m_l - \bar{d}_l\|^2 \nonumber \\
 & \quad + \sum_{t=s_t}^{s_t+q-1}\frac{\gamma}{M\rho}\Big(4C_g^2\eta^2_t\gamma^2 + 4L_g^2\eta^2_t\gamma^2 + 8L^2_fC^2_g\eta^2_t\gamma^2\Big)\sum_{m = 1}^M\mathbb{E}\|d^m_t - \bar{d}_t\|^2 + \sum_{t=s_t}^{s_t+q-1}\frac{2\sigma^2\gamma}{\rho}\big( \alpha^2_{t+1} + \beta^2_{t+1} + \varrho^2_{t+1}\big) \nonumber \\
 & \leq  \sum_{t=s_t}^{s_t+q-1}\frac{1}{M}\sum_{m=1}^{M}\Bigg( -\frac{C_f^2\gamma}{2\rho}\eta_t\|u^m_t - \nabla g^m(\bar{x}_t)\|^2 - \frac{C_g^2\gamma}{\rho}\eta_t\|v^m_t - \nabla f^m(h^m_t)\|^2 \nonumber \\
 & \quad - \frac{\gamma}{\rho}\big(\frac{B}{2} - 4C_g^2L^2_f -  \frac{c^2_2C^2_gL^2_f}{864q^3\gamma^3L^3_{fg}C^3_{fg}}\big)\eta_t\|h^m_t - g^m(\bar{x}_t)\|^2  \Bigg) \nonumber \\
 & \quad -\sum_{t=s_t}^{s_t+q-1}\Big(\frac{\rho\gamma\eta_t}{2} - \frac{4C_g^2\eta^2_t\gamma^3}{\rho} - \frac{4L_g^2\eta^2_t\gamma^3}{\rho} - \frac{8L^2_fC^2_g\eta^2_t\gamma^3}{\rho} \Big)\|\bar{d}_t\|^2 \nonumber \\
 & \quad + \frac{\gamma}{M\rho}\Big( BC^2_g + 5C^2_fL^2_g + \frac{c^2_2C^2_gL^2_f}{864q^3\gamma^3L^3_{fg}C^3_{fg}} \Big)\frac{\rho^2}{(24)^2L^2_{fg}C^2_{fg}}\sum_{t=s_t}^{s_t+q-1}\eta_t\sum_{m = 1}^M \mathbb{E}\|d^m_t - \bar{d}_t\|^2 \nonumber \\
 & \quad + \frac{\gamma}{M\rho}\frac{\gamma\rho}{6qL_{fg}C_{fg}}\Big(C_g^2 + L_g^2 + 2L^2_fC^2_g\Big)\sum_{t=s_t}^{s_t+q-1}\eta_t\sum_{m = 1}^M\mathbb{E}\|d^m_t - \bar{d}_t\|^2 \nonumber \\
 & \quad  +\frac{\sigma^2}{12qL_{fg}C_{fg}}\big( c^2_1 + c^2_2 + c^2_3 \big)\sum_{t=s_t}^{s_t+q-1}\eta^3_t,
 \end{align}
where the second inequality is due to $\eta_t \leq \frac{\rho}{24 q\gamma L_{fg}C_{fg}}$ for all $t\geq 0$.

Let $\gamma^2\geq\frac{\rho^2\sqrt{c^2_1+c^2_3}}{24\sqrt{30}q L^2_{fg}C^2_{fg}}$, we have
\begin{align} \label{eq:W7}
\frac{\rho^2C^2_g}{(24)^2L^2_{fg}C^2_{fg}} \frac{\rho^2(c_1^2+c_3^2)}{120q^2\gamma^4C^2_{fg}L^2_{fg}C^2_g} \leq \frac{1}{4}.
\end{align}
Set $\Theta = \Big(5C^2_fL^2_g + \frac{c^2_2C^2_gL^2_f}{864q^3\gamma^3L^3_{fg}C^3_{fg}} \Big)\frac{\rho^2}{(24)^2L^2_{fg}C^2_{fg}}+\frac{\gamma\rho}{6qL_{fg}C_{fg}}\Big(C_g^2 + L_g^2 + 2L^2_fC^2_g\Big)$.
Based on the above Lemma \ref{lem:A12}, then we have
\begin{align} \label{eq:W8}
 & \sum_{t=s_t}^{s_t+q-1}\big( \Omega_{t+1} - \Omega_t \big) \nonumber \\
 & \leq  \sum_{t=s_t}^{s_t+q-1}\frac{1}{M}\sum_{m=1}^{M}\Bigg( -\frac{C_f^2\gamma}{2\rho}\eta_t\mathbb{E}\|u^m_t - \nabla g^m(\bar{x}_t)\|^2 - \frac{C_g^2\gamma}{\rho}\eta_t\mathbb{E}\|v^m_t - \nabla f^m(h^m_t)\|^2 \nonumber \\
 & \quad - \frac{\gamma}{\rho}\big(\frac{B}{2} - 4C_g^2L^2_f -  \frac{c^2_2C^2_gL^2_f}{864q^3\gamma^3L^3_{fg}C^3_{fg}}\big)\eta_t\mathbb{E}\|h^m_t - g^m(\bar{x}_t)\|^2  \Bigg) \nonumber \\
 & \quad -\sum_{t=s_t}^{s_t+q-1}\Big(\frac{\rho\gamma\eta_t}{2} - \frac{4C_g^2\eta^2_t\gamma^3}{\rho} - \frac{4L_g^2\eta^2_t\gamma^3}{\rho} - \frac{8L^2_fC^2_g\eta^2_t\gamma^3}{\rho} \Big)\mathbb{E}\|\bar{d}_t\|^2 \nonumber \\
 & \quad + \frac{\gamma}{M\rho}\Big( BC^2_g + 5C^2_fL^2_g + \frac{c^2_2C^2_gL^2_f}{864q^3\gamma^3L^3_{fg}C^3_{fg}} \Big)\frac{\rho^2}{(24)^2L^2_{fg}C^2_{fg}} \nonumber \\
 & \qquad \cdot\Bigg( \frac{6M}{5}\sum_{t=s_t}^{s_t+q-1}\eta_t\mathbb{E}\|\bar{d}_t\|^2+ \frac{\rho^2(c_1^2+c_3^2)}{120q^2\gamma^4C^2_{fg}L^2_{fg}C^2_g}\sum_{t=s_t}^{s_t+q-1}\eta_t\sum_{m=1}^M\mathbb{E}\|h^m_t-g^m(\bar{x}_t)\|^2
   + \frac{3M\hat{\delta}^2}{5\gamma^2 L^2_{fg}}\sum_{t=s_t}^{s_t+q-1}\eta^3_t \Bigg) \nonumber \\
 & \quad + \frac{\gamma}{M\rho}\frac{\gamma\rho}{6qL_{fg}C_{fg}}\Big(C_g^2 + L_g^2 + 2L^2_fC^2_g\Big) \nonumber \\
 & \qquad \cdot \Bigg( \frac{6M}{5}\sum_{t=s_t}^{s_t+q-1}\eta_t\mathbb{E}\|\bar{d}_t\|^2 + \frac{\rho^2(c_1^2+c_3^2)}{120q^2\gamma^4C^2_{fg}L^2_{fg}C^2_g}\sum_{t=s_t}^{s_t+q-1}\eta_t\sum_{m=1}^M\mathbb{E}\|h^m_t-g^m(\bar{x}_t)\|^2+ \frac{3M\hat{\delta}^2}{5\gamma^2 L^2_{fg}}\sum_{t=s_t}^{s_t+q-1}\eta^3_t\Bigg)  \nonumber \\
 & \quad +\frac{\sigma^2}{12qL_{fg}C_{fg}}\big( c^2_1 + c^2_2 + c^2_3 \big)\sum_{t=s_t}^{s_t+q-1}\eta^3_t \nonumber \\
 & \leq \sum_{t=s_t}^{s_t+q-1}\frac{1}{M}\sum_{m=1}^{M}\Bigg( -\frac{C_f^2\gamma}{2\rho}\eta_t\mathbb{E}\|u^m_t - \nabla g^m(\bar{x}_t)\|^2 - \frac{C_g^2\gamma}{\rho}\eta_t\mathbb{E}\|v^m_t - \nabla f^m(h^m_t)\|^2 \nonumber \\
 & \quad - \frac{\gamma}{\rho}\Big(\frac{B}{4} - 4C_g^2L^2_f -  \frac{c^2_2C^2_gL^2_f}{864q^3\gamma^3L^3_{fg}C^3_{fg}} - \frac{\Theta\rho^2(c_1^2+c_3^2)}{120q^2\gamma^4C^2_{fg}L^2_{fg}C^2_g}\Big)\eta_t\mathbb{E}\|h^m_t - g^m(\bar{x}_t)\|^2  \Bigg) \nonumber \\
 & \quad -\sum_{t=s_t}^{s_t+q-1}\Big(\frac{\rho\gamma\eta_t}{2} - \frac{4C_g^2\eta^2_t\gamma^3}{\rho} - \frac{4L_g^2\eta^2_t\gamma^3}{\rho} - \frac{8L^2_fC^2_g\eta^2_t\gamma^3}{\rho} - \frac{6\gamma}{5\rho}\big(\Theta+\frac{BC^2_g\rho^2}{(24)^2L^2_{fg}C^2_{fg}}\big)\eta_t \Big)\mathbb{E}\|\bar{d}_t\|^2 \nonumber \\
 & \quad + \Big(\Theta+\frac{BC^2_g\rho^2}{(24)^2L^2_{fg}C^2_{fg}}\Big)\frac{3\hat{\delta}^2}{5\rho\gamma L^2_{fg}}\sum_{t=s_t}^{s_t+q-1}\eta^3_t + \frac{\sigma^2}{12qL_{fg}C_{fg}}\big( c^2_1 + c^2_2 + c^2_3 \big)\sum_{t=s_t}^{s_t+q-1}\eta^3_t \nonumber \\
 & \leq \sum_{t=s_t}^{s_t+q-1}\frac{1}{M}\sum_{m=1}^{M}\Bigg( -\frac{C_f^2\gamma}{2\rho}\eta_t\mathbb{E}\|u^m_t - \nabla g^m(\bar{x}_t)\|^2 - \frac{C_g^2\gamma}{\rho}\eta_t\mathbb{E}\|v^m_t - \nabla f^m(h^m_t)\|^2 - \frac{\gamma C^2_gL^2_f}{\rho}\eta_t\mathbb{E}\|h^m_t - g^m(\bar{x}_t)\|^2  \Bigg) \nonumber \\
 & \quad -\sum_{t=s_t}^{s_t+q-1}\frac{\rho\gamma\eta_t}{4} \mathbb{E}\|\bar{d}_t\|^2  + \frac{\rho\hat{\delta}^2}{16\gamma L^2_{fg}}\sum_{t=s_t}^{s_t+q-1}\eta^3_t + \frac{\sigma^2}{12qL_{fg}C_{fg}}\big( c^2_1 + c^2_2 + c^2_3 \big)\sum_{t=s_t}^{s_t+q-1}\eta^3_t,
 \end{align}
where the second inequality holds by the above inequality \eqref{eq:W7}, and the last inequality holds by
$B\geq 20C_g^2L^2_f + \frac{c^2_2C^2_gL^2_f}{216q^3\gamma^3L^3_{fg}C^3_{fg}} + \frac{\Theta\rho^2(c_1^2+c_3^2)}{30q^2\gamma^4C^2_{fg}L^2_{fg}C^2_g}$,
$\gamma \leq \frac{3\rho qL_{fg}C_{fg}}{4(C^2_g+L^2_g+ 2L^2_fC^2_g)}$ (i.e., the following inequality \eqref{eq:W9}) and $\Theta+\frac{BC^2_g\rho^2}{(24)^2L^2_{fg}C^2_{fg}}\leq \frac{5\rho^2}{48}$.

Since $\eta_t \leq \frac{\rho}{24q\gamma L_{fg}C_{fg}}$ and $\gamma \leq \frac{3\rho qL_{fg}C_{fg}}{4(C^2_g+L^2_g+ 2L^2_fC^2_g)}$, we have
\begin{align} \label{eq:W9}
 \gamma^2 \leq \frac{\rho^2}{32(C^2_g+L^2_g+ 2L^2_fC^2_g)}\frac{24q\gamma L_{fg}C_{fg}}{\rho} \leq \frac{\rho^2}{32\eta_t(C^2_g+L^2_g+ 2L^2_fC^2_g)}.
\end{align}

Summing the above inequality \ref{eq:W8} from $t=1$ to $T$, then we have
\begin{align} \label{eq:W12}
 & \sum_{t=1}^{T}\big( \Omega_{t+1} - \Omega_t \big) \nonumber \\
 & \leq \sum_{t=1}^{T}\frac{1}{M}\sum_{m=1}^{M}\Bigg( -\frac{C_f^2\gamma}{2\rho}\eta_t\mathbb{E}\|u^m_t - \nabla g^m(\bar{x}_t)\|^2 - \frac{C_g^2\gamma}{\rho}\eta_t\mathbb{E}\|v^m_t - \nabla f^m(h^m_t)\|^2 - \frac{\gamma C^2_gL^2_f}{\rho}\eta_t\mathbb{E}\|h^m_t - g^m(\bar{x}_t)\|^2  \Bigg) \nonumber \\
 & \quad -\sum_{t=1}^{T}\frac{\rho\gamma\eta_t}{4} \mathbb{E}\|\bar{d}_t\|^2  + \frac{\rho\hat{\delta}^2}{16\gamma L^2_{fg}}\sum_{t=1}^{T}\eta^3_t + \frac{\sigma^2}{12qL_{fg}C_{fg}}\big( c^2_1 + c^2_2 + c^2_3 \big)\sum_{t=1}^{T}\eta^3_t.
 \end{align}

Since $h^m_1 = \frac{1}{q}\sum_{j=1}^q g^m(x^m_1;\zeta^m_{1,j})$, $u^m_1 = \frac{1}{q}\sum_{j=1}^q \nabla g^m(x^m_1;\zeta^m_{1,j})$ and $v^m_1 = \frac{1}{q}\sum_{j=1}^q \nabla f(h^m_1;\xi^m_{1,j})$ for all $m\in [M]$, we have
\begin{align} \label{eq:W13}
 \Omega_1 &= \mathbb{E}\Big [F(\bar{x}_1) + \frac{\gamma}{\rho\eta_0} \frac{1}{M}\sum_{m=1}^M \Big( \|h^m_1 - g^m(x^m_1)\|^2
  + \|u^m_1 - \nabla g^m(x^m_1)\|^2 + \|v^m_1 - \nabla f^m(h^m_1)\|^2 \Big)\Big] \nonumber \\
 & \leq F(\bar{x}_1) + \frac{3\gamma\sigma^2}{q\rho\eta_0},
\end{align}
where the last inequality holds by Assumption \ref{ass:2}.

Since $\eta_t=\frac{k}{(n+t)^{1/3}}$ is decreasing, i.e., $\eta_T^{-1} \geq \eta_t^{-1}$ for any $0\leq t\leq T$, we have
 \begin{align}  \label{eq:W14}
 & \frac{1}{T} \sum_{t=1}^{T}\mathbb{E}\Big[\frac{1}{M}\sum_{m=1}^{M}\frac{1}{\rho^2}\Big( 2C_f^2\|u^m_t - \nabla g^m(\bar{x}_t)\|^2 + 4C_g^2\|v^m_t - \nabla f^m(h^m_t)\|^2 + 4C^2_gL^2_f\|h^m_t - g^m(\bar{x}_t)\|^2 \Big) +\|\bar{d}_t\|^2 \Big] \nonumber \\
 & \leq  \frac{4}{T\rho\gamma\eta_T} \sum_{t=1}^T\big(\Omega_t - \Omega_{t+1}\big) + \frac{\hat{\delta}^2}{4T\gamma^2\eta_T L^2_{fg}}\sum_{t=1}^{T}\eta^3_t + \frac{\big( c^2_1 + c^2_2 + c^2_3 \big)\sigma^2}{3T\rho\gamma\eta_TqL_{fg}C_{fg}}\sum_{t=1}^{T}\eta^3_t \nonumber \\
 & \leq \frac{4}{T\rho\gamma\eta_T} \big( F(\bar{x}_1) + \frac{3\gamma\sigma^2}{\rho\eta_0} - F^* \big) + \Big(\frac{\hat{\delta}^2}{4T\gamma^2\eta_T L^2_{fg}} + \frac{\big( c^2_1 + c^2_2 + c^2_3 \big)\sigma^2}{3T\rho\gamma\eta_TqL_{fg}C_{fg}}\Big)\sum_{t=1}^T\eta^3_t \nonumber \\
 & \leq \frac{4}{T\rho\gamma\eta_T} \big( F(\bar{x}_1) + \frac{3\gamma\sigma^2}{q\rho\eta_0} - F^* \big) + \Big(\frac{\hat{\delta}^2}{4T\gamma^2\eta_T L^2_{fg}} + \frac{\big( c^2_1 + c^2_2 + c^2_3 \big)\sigma^2}{3T\rho\gamma\eta_TqL_{fg}C_{fg}}\Big)\int^T_1\frac{k^3}{n+t} dt \nonumber \\
 & \leq \frac{4}{T\rho\gamma\eta_T} \big( F(\bar{x}_1) + \frac{3\gamma\sigma^2}{q\rho\eta_0} - F^* \big) + \frac{1}{T\eta_T}\Big(\frac{\hat{\delta}^2}{4\gamma^2 L^2_{fg}} + \frac{\big( c^2_1 + c^2_2 + c^2_3 \big)\sigma^2}{3\rho\gamma qL_{fg}C_{fg}}\Big)\ln(n+T) \nonumber \\
 & = \bigg( \frac{4(F(\bar{x}_1) - F^*)}{k\rho\gamma} + \frac{12n^{1/3}\sigma^2}{qk^2\rho^2} + 4k^2\Big(\frac{\hat{\delta}^2}{4\gamma^2 L^2_{fg}} + \frac{\big( c^2_1 + c^2_2 + c^2_3 \big)\sigma^2}{3\rho\gamma qL_{fg}C_{fg}}\Big)\ln(n+T) \bigg)\frac{(n+T)^{1/3}}{T},
\end{align}
where the second inequality holds by the above inequality \eqref{eq:W13}.
Let $G = \frac{4(F(\bar{x}_1) - F^*)}{k\rho\gamma} + \frac{12n^{1/3}\sigma^2}{qk^2\rho^2} + 4k^2\Big(\frac{\hat{\delta}^2}{4\rho\gamma^2 L^2_{fg}} + \frac{\big( c^2_1 + c^2_2 + c^2_3 \big)\sigma^2}{3\rho\gamma qL_{fg}C_{fg}}\Big)\ln(n+T)$.
According to the above Lemma \ref{lem:A6}, then
we have
\begin{align} \label{eq:W15}
  & \frac{1}{T} \sum_{t=1}^{T}\mathbb{E}\Big[\frac{1}{\rho^2}\|\bar{w}_t - \nabla F(\bar{x}_t)\|^2 +\|\bar{d}_t\|^2 \Big]  \nonumber \\
  & \leq \frac{1}{T} \sum_{t=1}^{T}\mathbb{E}\Big[\frac{1}{M}\sum_{m=1}^{M}\frac{1}{\rho^2}\Big( 2C_f^2\|u^m_t - \nabla g^m(\bar{x}_t)\|^2 + 4C_g^2\|v^m_t - \nabla f^m(h^m_t)\|^2 + 4C^2_gL^2_f\|h^m_t - g^m(\bar{x}_t)\|^2 \Big) +\|\bar{d}_t\|^2 \Big] \nonumber \\
  &  \leq \frac{G}{T}(n+T)^{1/3},
\end{align}
where the first inequality holds by Lemma \ref{lem:A6}, and the last inequality holds by \eqref{eq:W14}.

Since $\bar{d}_t = \frac{\bar{x}_t - \bar{x}_{t+1}}{\eta_t\gamma} = \frac{\eta_t\gamma A_t^{-1}\bar{w}_t}{\eta_t\gamma} = A_t^{-1}\bar{w}_t$,
and let $\mathcal{G}_t = \frac{1}{\rho}\|\bar{w}_t - \nabla F(\bar{x}_t)\| +\|\bar{d}_t\|$, we have
\begin{align}
\mathcal{G}_t & = \frac{1}{\rho}\|\bar{w}_t - \nabla F(\bar{x}_t)\| +\|\bar{d}_t\| = \frac{1}{\rho}\|\bar{w}_t - \nabla F(\bar{x}_t)\| +\|A_t^{-1}\bar{w}_t\| \nonumber \\
& \geq \|A^{-1}_t\bar{w}_t\|
+ \frac{1}{\rho}\|\bar{w}_t-\nabla F(\bar{x}_t)\| \nonumber \\
& = \frac{1}{\|A_t\|}\|A_t\|\|A_t^{-1}\bar{w}_t\|
+ \frac{1}{\rho}\|\bar{w}_t - \nabla F(\bar{x}_t)\| \nonumber \\
& \geq \frac{1}{\|A_t\|}\|\bar{w}_t\|
+ \frac{1}{\rho}\|\bar{w}_t - \nabla F(\bar{x}_t)\| \nonumber \\
& \mathop{\geq}^{(i)} \frac{1}{\|A_t\|}\|\bar{w}_t\|
+ \frac{1}{\|A_t\|}\|\nabla F(\bar{x}_t)-\bar{w}_t\| \nonumber \\
& \geq \frac{1}{\|A_t\|}\|\nabla F(\bar{x}_t)\|,
\end{align}
where the inequality $(i)$ holds by $\|A_t\| \geq \rho$ for all $t\geq1$ due to Assumption \ref{ass:6}. Then we have
\begin{align}
 \|\nabla F(\bar{x}_t)\| \leq \|A_t\|\mathcal{G}_t.
\end{align}
According to Cauchy-Schwarz inequality, we have
\begin{align} \label{eq:W16}
\frac{1}{T}\sum_{t=1}^T\mathbb{E}\|\nabla F(\bar{x}_t)\| \leq \frac{1}{T}\sum_{t=1}^T\mathbb{E}\big[\mathcal{G}_t \|A_t\|\big] \leq \sqrt{\frac{1}{T}\sum_{t=1}^T\mathbb{E}[\mathcal{G}_t^2]} \sqrt{\frac{1}{T}\sum_{t=1}^T\mathbb{E}\|A_t\|^2}.
\end{align}
According to the above inequality \eqref{eq:W15}, we have
\begin{align} \label{eq:W17}
 \frac{1}{T} \sum_{t=1}^T \mathbb{E}[\mathcal{G}_t^2] & \leq \frac{1}{T} \sum_{t=1}^{T}\mathbb{E}\Big[\frac{2}{\rho^2}\|\bar{w}_t - \nabla F(\bar{x}_t)\|^2 + 2 \|\bar{d}_t\|^2 \Big] \nonumber \\
 & \leq \frac{2G}{T}(n+T)^{1/3}.
\end{align}

Combining the above inequalities \eqref{eq:W16} with \eqref{eq:W17}, we have
\begin{align}
\frac{1}{T}\sum_{t=1}^T\mathbb{E}\|\nabla F(\bar{x}_t)\| & \leq  \sqrt{\frac{1}{T}\sum_{t=1}^T\mathbb{E}[\mathcal{G}_t^2]} \sqrt{\frac{1}{T}\sum_{t=1}^T\mathbb{E}\|A_t\|^2} \nonumber \\
& \leq \Big( \frac{\sqrt{2G}n^{1/6}}{T^{1/2}} + \frac{\sqrt{2G}}{T^{1/3}}\Big)\sqrt{\frac{1}{T}\sum_{t=1}^T\mathbb{E}\|A_t\|^2}.
\end{align}
\end{proof}

\end{document}